\documentclass{article}

\usepackage[release]{tfc_report}

\usepackage{nicefrac}
\usepackage{graphicx}
\usetikzlibrary{arrows.meta,calc,decorations.pathreplacing,patterns}
\usepgfplotslibrary{groupplots}
\usepackage{xspace}
\usepackage{placeins}

\newif\iftodos
\ifdefined\externalbuild\todosfalse\else\todostrue\fi

\colorlet{charlietzeroalpha}{tfcgreen}
\colorlet{charlietzerobeta}{tfcengineering}
\colorlet{charlietzerorc}{tfcengineering}
\colorlet{charliechronostwo}{tfcproduct}
\colorlet{charlietimesfmtwopfive}{tfcevent}
\colorlet{charlietirextwo}{tfcresearch}
\definecolor{charlietimesfmthree}{HTML}{B0407A}
\colorlet{charlienaivertdtwo}{tfcmuted}
\usepackage{pgfplotstable}
\usetikzlibrary{calc,positioning}
\usepgfplotslibrary{groupplots}
\usepgfplotslibrary{fillbetween}
\pgfplotsset{compat=1.18}

\providecolor{charlienaivertdtwo}{HTML}{C9521C}
\providecolor{charlietzeroalpha}{HTML}{5B3FB0}
\providecolor{charlietzerobeta}{HTML}{0F7A52}
\providecolor{charliechronostwo}{HTML}{2A78D6}
\providecolor{charlietimesfmtwopfive}{HTML}{9A7100}
\providecolor{charlietimesfmthree}{HTML}{BD4A7C}
\providecolor{charlietirextwo}{HTML}{8C4A00}
\providecolor{charliecontrol}{HTML}{5B6470}
\providecolor{charlierule}{HTML}{99A0A8}
\providecolor{charlieperiodone}{HTML}{2456A4}
\providecolor{charlieperiodtwo}{HTML}{B26A00}
\providecolor{charlieperiodthree}{HTML}{1D7A46}
\providecolor{charlietzerorc}{HTML}{0F7A52}

\pgfplotsset{charlieaxis/.style={
    axis lines=left,
    axis line style={draw=black!55},
    tick align=outside,
    tick style={draw=black!55},
    ymajorgrids=true,
    grid style={draw=black!12},
    every axis label/.append style={font=\small},
    tick label style={font=\scriptsize},
    legend style={font=\scriptsize, draw=none, fill=none},
    legend cell align=left,
}}

\pgfplotsset{charliefan/.style={charlieaxis, ymajorgrids=false, xmin=0, xmax=23,
    xtick={0,6,12,18}, width=\charliepanelwidth, height=\charliepanelheight}}
\providecommand{\charliepanelwidth}{4.2cm}
\providecommand{\charliepanelheight}{3.4cm}

\definecolor{figblue}{HTML}{2F6FB2}
\definecolor{figorange}{HTML}{C05B28}
\definecolor{figgreen}{HTML}{4A8F5D}
\definecolor{figgray}{HTML}{8A8F98}
\newcommand{\tzero}{$t_0$\xspace}
\newcommand{\tzeroalpha}{\texttt{t0-alpha}\xspace}
\newcommand{\tzerobeta}{\texttt{t0-beta}\xspace}
\newcommand{\tzeroalphahl}{\mbox{\textcolor{tfcgreen}{\texttt{t0-alpha}}}\xspace}
\newcommand{\tzerobetahl}{\mbox{\textcolor{tfcengineering}{\texttt{t0-beta}}}\xspace}

\newcommand{\weightsref}{\href{https://huggingface.co/theforecastingcompany/t0-alpha}%
  {\texttt{huggingface.co/theforecastingcompany/t0-alpha}}}

\newcommand{\weightsbetaref}{\href{https://huggingface.co/theforecastingcompany/t0-beta}%
  {\textcolor{tfcengineering}{\texttt{huggingface.co/theforecastingcompany/t0-beta}}}}

\title{\tzero: A Time-Series Foundation Model\\for Forecasting with Context}

\newcommand{\eqcontrib}{\textsuperscript{*}}
\newcommand{\extaffil}{\textsuperscript{\S}}

\author{%
  Lucas~Meyer\eqcontrib{} \quad Claudio~Sole\eqcontrib{} \quad
  Huikan~Xiang\eqcontrib{} \quad Nicolas~Li \quad Lucas~Franceschino \quad
  Arnau~Quera-Bofarull\extaffil{} \quad Maarten~P.~Scholl\extaffil{} \quad
  Joachim~Fainberg\eqcontrib{} \quad Geoffrey~N\'egiar\eqcontrib{}%
}

\authornote{\eqcontrib{}Equal contribution. \quad
  \extaffil{}Macrocosm, author of the independent evaluation in
  Section~\ref{sec:usecase-ercot}.}
\authornote{Correspondence: \href{mailto:research@theforecastingcompany.com}%
  {\texttt{research@theforecastingcompany.com}}}

\reportmeta{Date}{\today}
\reportmeta{Weights (\tzeroalpha)}{\weightsref{}}
\reportmeta{Weights (\tzerobeta)}{\weightsbetaref{}}
\reportmeta{Code}{\url{https://github.com/theforecastingcompany/tfc-t0}}
\reportmeta{Python Package}{\code{tfc-t0}}

\begin{document}

\begin{abstract}

We present \tzero, a family of open-weights foundation models for forecasting
with multivariate context. We release its first two members: \tzeroalphahl
and \tzerobetahl, respectively 102M and 256M parameters. Both condition their
forecasts on target history, past covariates, and known-future covariates,
without task-specific retraining. Their transformer layers alternate attention
along time and across variates. They produce probabilistic forecasts
through quantile predictions. Pretraining combines curated public data with
synthetic generator families constructed to contain covariate-to-target
dependencies. On GIFT-Eval, \tzeroalphahl reaches an aggregate CRPS of 0.4941,
and \tzerobetahl a CRPS of 0.4738 and a MASE of 0.6865, third on both and
within 4.0\% of the best zero-shot TSFM. On fev-bench they score 42.2 and 46.7
in skill, the latter third again and 2.0 points behind the leader. We analyze
\tzeroalphahl in depth. Known-future covariates raise its skill by 6.3
percentage points across 30 tasks. The report also examines its calibration,
its rollout strategy on long horizons, and its robustness to missing data. On
the Victoria electricity-demand benchmark, \tzerobetahl is among the most
accurate models with a context of nearly a year. In an independent Macrocosm
evaluation of hourly ERCOT prices over 29 months, both cut the MAE of the lagged-price
baseline by 38\%.

\end{abstract}

\maketitle

\enlargethispage{2\baselineskip}
\tfctoc

\section{Introduction}

Organizations plan against the future they expect. How much energy a region
will consume next week, how many data centers to build over five years,
whether demand for winter clothing rises next season: energy management,
capacity provisioning, inventory management, and staffing all rest on a
forecast.

Forecasting methods have moved from simple baselines to statistical models,
then to deep learning, and most recently from task-specialized models to
general time series foundation models (TSFMs)\label{gl:tsfm}. The
\emph{naive} forecast repeats the last observed value across the whole
horizon, and a seasonal naive\label{gl:seasonal-naive} forecast repeats the
value observed one seasonal period earlier. Above them sit models fitted to
data. ARIMA \citep{box1970timeseries} and gradient-boosted trees such as
XGBoost \citep{chen2016xgboost}, CatBoost \citep{prokhorenkova2018catboost},
and LightGBM \citep{ke2017lightgbm} took the top places in the 2020 M5 forecasting competition
\citep{makridakis2022m5} and were until recently the \emph{de facto} choice.
Every model of that generation is fitted anew for each dataset, and refitted
as the distribution shifts, with manual tuning and feature engineering at
each turn.

\begin{figure}[tb]
  \centering
  \newcommand{\demandyscale}{0.82}
\newcommand{\demandrowheight}{0.42cm}

\pgfplotsset{
  variaterow/.style={
    width=5.3cm, height=\demandrowheight, scale only axis,
    axis y line=none, axis x line=none,
    xmin=0.5, xmax=12.5,
    enlarge y limits={value=0.12},
    clip=false,
  },
  ctx/.style={restrict x to domain=0.5:9},   
  fut/.style={restrict x to domain=9:12.5},  
}

\newcommand{\rowlab}[4]{%
  \node[anchor=east, font=\tiny, color=#3, inner sep=1.5pt] at (#1,#2) {#4};%
}

\newcommand{\cityhead}[3]{%
  \node[anchor=south, font=\scriptsize, color=tfcink, inner sep=1pt]
    at ({#1+2.65},#2) {#3};%
}

\newcommand{\targetrow}[4]{%
  \begin{axis}[variaterow, at={(#1cm,#2cm)}]
    \addplot[fut, draw=none, name path=lo] table
      [col sep=comma, x=month, y=#4_lo] {figures/seasonal-demand.csv};
    \addplot[fut, draw=none, name path=hi] table
      [col sep=comma, x=month, y=#4_hi] {figures/seasonal-demand.csv};
    \addplot[#3, fill opacity=0.16, draw=none] fill between[of=lo and hi];
    \addplot[fut, draw=#3, dashed, line width=0.6pt] table
      [col sep=comma, x=month, y=#4_fc] {figures/seasonal-demand.csv};
    \addplot[ctx, draw=#3, line width=0.7pt] table
      [col sep=comma, x=month, y=#4] {figures/seasonal-demand.csv};
  \end{axis}%
}

\newcommand{\covrow}[4]{%
  \begin{axis}[variaterow, at={(#1cm,#2cm)}]
    \addplot[#3] table [col sep=comma, x=month, y=#4] {figures/seasonal-demand.csv};
  \end{axis}%
}

\newcommand{\months}[2]{%
  \foreach \m/\L in {1/J,2/F,3/M,4/A,5/M,6/J,7/J,8/A,9/S,10/O,11/N,12/D}{%
    \node[font=\tiny, color=tfcmuted, anchor=north, inner sep=1pt]
      at ({#1 + (\m-0.5)/12*5.3}, #2) {\L};%
  }%
  \draw[tfchair, line width=0.4pt] (#1,#2) -- ({#1+5.3},#2);%
}

\newcommand{\descr}[3]{%
  \node[anchor=west, font=\tiny\itshape, color=tfcmuted,
        fill=tfcsurface, inner xsep=3pt, inner ysep=1.5pt,
        text width=5.15cm] at (#1,#2) {#3};%
}

\newcommand{\cutoffrule}[1]{%
  \draw[tfcgreenstroke, line width=0.5pt]
    ({#1+3.754},-4.24) -- ({#1+3.754},0.92);
  \node[font=\tiny, color=tfcgreen, anchor=south, align=center, inner sep=1pt]
    at ({#1+3.754},0.92) {information\\cutoff};%
}

\newcommand{\kline}[1]{%
  \tikz[baseline=-0.55ex]{\draw[#1] (0,0) -- (0.40,0);}}
\newcommand{\kband}[1]{%
  \tikz[baseline=-0.55ex]{%
    \fill[#1, fill opacity=0.16] (0,-0.05) rectangle (0.40,0.05);%
    \draw[#1, dashed, line width=0.6pt] (0,0) -- (0.40,0);}}

\begin{tikzpicture}[yscale=\demandyscale]

  \node[anchor=west, font=\scriptsize\bfseries, color=tfcink]
    at (0,1.62) {France};
  \cutoffrule{1.35}

  \cityhead{1.35}{0.98}{Paris}
  \rowlab{1.25}{0.70}{tfcbody}{gloves}
  \rowlab{1.25}{0.12}{tfcbody}{flip-flops}
  \rowlab{1.25}{-0.46}{tfcmuted}{temperature}
  \rowlab{1.25}{-1.04}{tfcmuted}{promotion}
  \targetrow{1.35}{0.50}{tfccapuni}{paris_gl}
  \targetrow{1.35}{-0.08}{tfccapmulti}{paris_ff}
  \covrow{1.35}{-0.66}{ctx, draw=tfcmuted, dashed, line width=0.6pt}{paris_temp}
  \covrow{1.35}{-1.24}{draw=tfccapcov, line width=0.6pt,
                       const plot mark right}{paris_promo}

  \cityhead{1.35}{-1.98}{Marseille}
  \rowlab{1.25}{-2.26}{tfcbody}{gloves}
  \rowlab{1.25}{-2.84}{tfcbody}{flip-flops}
  \rowlab{1.25}{-3.42}{tfcmuted}{temperature}
  \rowlab{1.25}{-4.00}{tfcmuted}{promotion}
  \targetrow{1.35}{-2.46}{tfccapuni}{marseille_gl}
  \targetrow{1.35}{-3.04}{tfccapmulti}{marseille_ff}
  \covrow{1.35}{-3.62}{ctx, draw=tfcmuted, dashed,
                       line width=0.6pt}{marseille_temp}
  \covrow{1.35}{-4.20}{draw=tfccapcov, line width=0.6pt,
                       const plot mark right}{marseille_promo}
  \months{1.35}{-4.24}

  \rowlab{1.25}{-4.88}{tfcmuted}{gloves descr.}
  \rowlab{1.25}{-5.46}{tfcmuted}{flip-flops descr.}
  \descr{1.35}{-4.88}{gant en cuir doubl\'e laine, adulte, noir}
  \descr{1.35}{-5.46}{tong en caoutchouc, mixte, EU 36--46}

  \node[anchor=west, font=\scriptsize\bfseries, color=tfcink]
    at (6.80,1.62) {United States};
  \cutoffrule{8.15}

  \cityhead{8.15}{0.98}{New York}
  \rowlab{8.05}{0.70}{tfcbody}{gloves}
  \rowlab{8.05}{0.12}{tfcbody}{flip-flops}
  \rowlab{8.05}{-0.46}{tfcmuted}{temperature}
  \rowlab{8.05}{-1.04}{tfcmuted}{promotion}
  \targetrow{8.15}{0.50}{tfccapuni}{nyc_gl}
  \targetrow{8.15}{-0.08}{tfccapmulti}{nyc_ff}
  \covrow{8.15}{-0.66}{ctx, draw=tfcmuted, dashed, line width=0.6pt}{nyc_temp}
  \covrow{8.15}{-1.24}{draw=tfccapcov, line width=0.6pt,
                       const plot mark right}{nyc_promo}

  \cityhead{8.15}{-1.98}{San Francisco}
  \rowlab{8.05}{-2.26}{tfcbody}{gloves}
  \rowlab{8.05}{-2.84}{tfcbody}{flip-flops}
  \rowlab{8.05}{-3.42}{tfcmuted}{temperature}
  \rowlab{8.05}{-4.00}{tfcmuted}{promotion}
  \targetrow{8.15}{-2.46}{tfccapuni}{sf_gl}
  \targetrow{8.15}{-3.04}{tfccapmulti}{sf_ff}
  \covrow{8.15}{-3.62}{ctx, draw=tfcmuted, dashed, line width=0.6pt}{sf_temp}
  \covrow{8.15}{-4.20}{draw=tfccapcov, line width=0.6pt,
                       const plot mark right}{sf_promo}
  \months{8.15}{-4.24}

  \rowlab{8.05}{-4.88}{tfcmuted}{gloves descr.}
  \rowlab{8.05}{-5.46}{tfcmuted}{flip-flops descr.}
  \descr{8.15}{-4.88}{wool-lined leather glove, adult, tan}
  \descr{8.15}{-5.46}{rubber thong sandal, unisex, US 5--13}

  \node[anchor=north west, font=\tiny, color=tfcbody, inner sep=0pt]
    at (1.35,-5.95) {%
      \kline{tfccapuni, line width=0.7pt}\,target, observed\quad
      \kband{tfccapuni}\,forecast\quad
      \kline{tfcmuted, dashed, line width=0.6pt}\,temperature: past
        covariate\quad
      \kline{tfccapcov, line width=0.6pt}\,promotion: known future\quad
      {\itshape\color{tfcmuted}description: static}};

\end{tikzpicture}
  \caption{Illustrative example of a time series problem involving
  multiple covariates. Demand must be forecast for two retail items in
  stores across different cities and countries, using historical
  temperatures, known-future promotions, and the product
  description.}
  \label{fig:multivariate-example}
\end{figure}
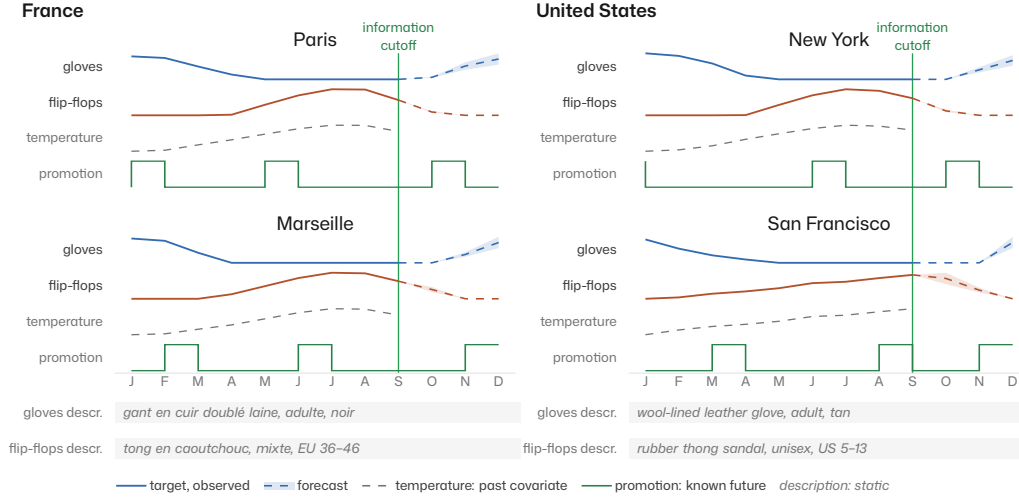

Time series foundation models such as Chronos \citep{ansari2024chronos},
Moirai \citep{woo2024moirai}, Toto \citep{cohen2024toto}, and \tzeroalpha, the
subject of this report, can forecast a dataset unseen at training time, thus achieving so-called
zero-shot\label{gl:zero-shot} forecasting. Pretrained once on a large and
diverse corpus, they generalize at inference time as large language models do,
and the best of them now beat tuned task-specific baselines on public
benchmarks \citep{aksu2024gifteval}. Most of them, however, read one
univariate series and nothing else. Forecasting problems rarely come that
bare. Demand responds to promotions, holidays, the weather and local events,
and the items of a catalog move together
(Figure~\ref{fig:multivariate-example}). We built \tzeroalpha to forecast
conditioned on that context.

This report also introduces \tzerobeta, its open-weights
256M-parameter successor, which places third on
GIFT-Eval and third on fev-bench among the zero-shot models we compare, and
carries its largest margins on the tasks that come with covariates.

Our contributions are as follows.
\begin{enumerate}
  \item \textbf{An open-weights artifact.} A 102M-parameter TSFM released on
    Hugging Face, with an inference Python package (\texttt{tfc-t0}) and a
    notebook that reproduces our GIFT-Eval results.
  \item \textbf{An architecture that leverages covariates through
    alternating attention.} Besides relying on attention along the time axis, the
    model attends across variates, which carry the past and
    known-future covariates alongside the target.
  \item \textbf{A thorough evaluation of covariate lift and deployment
    properties.} On the fev-bench public benchmark, passing covariates to the
    same model checkpoint raises skill by 6.3 points on the tasks with known-future
    covariates and by 2.7 points on those with past covariates only. The
    gains concentrate where a covariate plausibly drives the target. Beyond
    accuracy on the public benchmarks, we evaluate and report the properties
    a TSFM needs for deployment on real applications. We validate the
    calibration of \tzeroalpha and the behavior of its tails, measure its resilience to
    gaps in the context history, study how its accuracy holds as the context
    shortens, check that its rank is stable across the eleven metrics the
    leaderboard records, and benchmark it on an electricity-demand case study.
\end{enumerate}

\section{Problem Formulation}
\label{sec:problem}

A \emph{time series} is a collection of related variates indexed by the same
time axis, where a \emph{variate} is a quantity observed at successive time steps,
$y_{1:T} = (y_1, \dots, y_T)$ with $y_t \in \mathbb{R}$ for every $t$. When a
time series has a single variate it is said to be univariate. The time
series is then synonymous with its variate. Real-world applications typically
involve multivariate time series. For instance, a retailer might forecast the
demand for seasonal items based on temperature and promotions across different
cities
(Figure~\ref{fig:multivariate-example}).

The variates of a time series differ in the role they play, and in when their
values become available. A time series holds $V \ge 1$ \emph{target} variates
$\mathbf{y} = (y^{(v)}_{1:T})_{v=1}^{V}$, the quantities to forecast. It may
also hold $P \ge 0$ \emph{past covariates}\label{gl:past-cov}
$\mathbf{x} = (x^{(p)}_{1:T})_{p=1}^{P}$, observed over the context
$1{:}T$ only, and $F \ge 0$ \emph{known-future covariates}\label{gl:future-cov}
$\mathbf{z} = (z^{(f)}_{1:T+H})_{f=1}^{F}$, observed over the
context and over a horizon of $H$ steps. Covariates differ from target variates
in that they carry information the forecast can use but are not themselves
predicted for the task at hand. Figure~\ref{fig:multivariate-example}
illustrates
this distinction as the task is to predict item demands with the help of weather information,
and not predicting the weather itself. A time series may include \emph{static
covariates}\label{gl:static-cov} as well, which do not vary in
time, such as the category of a product or the format of a store. We leave static
covariates to future work.

Forecasting starts from an \emph{information cutoff datetime}, the time
step $T$ that bounds the context and after which nothing is observed, except
for future covariates. Given a prediction length of $H$ steps, the task is to
predict the targets over $T{+}1, \dots, T{+}H$ from the context, that is
everything observed up to $T$, together with any known-future covariates
over the prediction length itself.

A model can answer that task with a single value per time step, but this is rarely enough. Most applications require predictions with associated probabilities. For instance, staffing and energy-trading decisions hinge on how costly the worst plausible outcomes are, not on the expected outcome alone. Models are thus expected to predict multiple values
for each time step, corresponding either to different samples or to
quantile levels directly.

Given a set of $Q = |\mathcal{Q}|$ quantile levels $\mathcal{Q} \subset (0,1)$,
a TSFM produces a forecast $\hat{y}^{(v),q}_{T+h} \in \mathbb{R}^{V \times H
\times Q}$, an estimate of the $q$-quantile of the marginal distribution of
$y^{(v)}_{T+h}$ conditional on everything observed (\eqnref{eq:forecast}).

\begin{equation}
  \label{eq:forecast}
  \hat{y}^{(v),q}_{T+h}
  = f_\theta\!\left(\mathbf{y}_{1:T}, \mathbf{x}_{1:T},
    \mathbf{z}_{1:T+H}\right)^{(v),q}_{h},
  \qquad v \in [V],\; q \in \mathcal{Q},\; h \in [H].
\end{equation}

The goal of a TSFM, $f_\theta$, is to predict $\hat{y}$ for \emph{any} time series, regardless of $V$, $P$, $F$, $T$, and $H$ - that is, regardless of how many targets and covariates
it carries, how much context it provides, and how long a horizon is required.

\section[Related Work]{Related Work\protect\footnote{An interactive version
  of this section is available as a blog post:
  \href{https://www.theforecastingcompany.com/blog/from-arima-to-foundation-models/}{From ARIMA to Foundation Models}.}}
\label{sec:related}

\paragraph{Traditional statistical methods}

Forecasting vastly predates deep learning. The roots of modern forecasting may be found in antiquity, when Babylonian scholars turned centuries of recorded
observations into omen-based inference and, by the first millennium BCE, eclipse
prediction \citep{rochberg2004heavenly,steele2000eclipse}. As such, forecasting methods have evolved
throughout the years \citep{hyndman2021fpp}. Three families account for most of
the practice that came before foundation models: the autoregressive integrated
moving average (ARIMA) family \citep{box1970timeseries}, exponential smoothing,
and gradient-boosted decision trees (GBTs). Each of them fits its parameters to
one series or one dataset at a time.

As its name suggests, ARIMA combines three mechanisms. It is
\emph{autoregressive}: the next value is regressed on the values that precede
it. It takes a \emph{moving average}: the error term of that regression is
itself a linear combination of past errors. And it is \emph{integrated}: both
apply to the differences between consecutive time steps rather than to the raw
series, a transformation that removes trend. Each mechanism carries an order,
which dictates the number of associated parameters that are fitted at training
time. Variants extend the same core with seasonal terms (SARIMA
\citep{box2015tsa5}), exogenous regressors (ARIMAX \citep{pankratz1991dynreg}),
and a vector form for several series at once (VARIMA \citep{lutkepohl2005mts}).

Exponential smoothing methods produce forecasts by averaging the history of a
time series, modulating each observation by a factor that shrinks exponentially
with time. They are described by the ETS framework, whose name comes from the
three components it considers: error, trend, and seasonality
\citep{hyndman2008ets}. Every component carries parameters of its own.

The third family of forecasting methods, the GBTs, originates from tabular
machine learning rather than from the time series domain
\citep{friedman2001gbm}. Nonetheless, they have been used successfully in most
of the top-ranked entries of the M5 forecasting competition
\citep{makridakis2022m5}. Forecasting with them requires flattening the series
into a table of lagged values, calendar features, and covariates, which is how
they absorb promotions or weather with no change of machinery. The most common
implementations, namely XGBoost \citep{chen2016xgboost}, LightGBM
\citep{ke2017lightgbm}, and CatBoost \citep{prokhorenkova2018catboost}, differ
in how each tree is fitted and how categorical features are handled.

\paragraph{Deep learning models}

While ARIMA and ETS methods are limited to a few parameters fitted per time series, deep learning methods fit many more
parameters on datasets of time series, following a trend initiated by GBTs. They also relax the assumptions made about the structure of the
forecast. Nothing prescribes a decomposition into level, trend, and
seasonality. Which components matter and how they combine is now left to the model to learn from the data.

The models that followed diverge on three axes: how a series enters the
network, the architecture that processes it, and the form of the output. On the
first, DeepAR \citep{salinas2020deepar} and MQ-RNN \citep{wen2017mqrnn} read
the series one step at a time through a recurrence. N-BEATS takes the whole
lookback window at once, as a flat vector \citep{oreshkin2020nbeats}. PatchTST
cuts it into patches of consecutive steps and treats each patch as a token,
shortening the sequence attention must cover \citep{nie2023patchtst}. On the
second axis, the backbone is a recurrent network for DeepAR and MQ-RNN, a deep
stack of fully connected layers tied by backward and forward residual links for
N-BEATS, and attention for PatchTST and MQTransformer
\citep{eisenach2020mqtransformer}. They differ last in the form of the output.
N-BEATS and PatchTST are point-forecasters. They produce one value per horizon
step, without any information about its uncertainty. DeepAR instead emits the
parameters of a distribution at every step, and forecasts are drawn from it as
sample paths. The MQ family drops the sampling altogether and decodes every
horizon step and every quantile level in a single pass.

For all their differences, the models of this generation share the same
constraints: each is fitted to a single dataset and generalizes poorly beyond
it. Each is also tied to a single input schema. N-BEATS and PatchTST only
handle univariate time series. The MQ family declares a fixed number of past,
future, and static covariates prior to training. A new dataset, or a different
number of covariates, requires retraining.

\paragraph{Time series foundation models}

Foundation models supersede the previous generations through their versatility.
Where prior models had to be fitted anew for every dataset, a single TSFM
forecasts series from datasets it has never seen. No retraining is required.

Their architectures vary, but along recurring axes. The first is about how
a model ingests a time series. Most TSFMs break the series into patches.
Consecutive steps are grouped together, and the model produces an embedding for each
group. A patch is the closest analogue to the token of a language model, with one
major difference: language tokens are discrete, whereas time
series typically hold continuous values. Moirai varies the patch size with the
frequency of the series \citep{woo2024moirai}. Chronos-2 and Toto keep a single
patch size, which is now the common practice \citep{ansari2025chronos2,
cohen2024toto}. Other TSFMs, like TS-ICL and Falcon-X, abandon the patch grid
completely to avoid cutting the series artificially: the former casts
forecasting as timestamp-aligned regression \citep{lenaour2026tsicl}, the
latter maps each variate into a shared latent space \citep{liu2026falconx}.
Patches reappear on the output side, where one decoded patch covers
several horizon steps. Toto-2 produces multiple patches from the same context
before relying on rollout to generate forecasts beyond a given horizon
\citep{ansari2025chronos2}. TimesFM-3.0 decodes horizon patches longer than its
stride and blends the overlap \citep{timesfm3}, while forking sequences decode
a forecast from every position of the series in a single pass, so that the
overlapping horizons average part of the volatility away \citep{wen2017mqrnn,
potosnak2025forking}.

Regardless of how a model embeds the series, the result is always a sequence
of tokens. The second axis is how these tokens exchange information. Most TSFMs
are transformers whose attention runs along the time axis
\citep{ansari2025chronos2, cohen2024toto, das2024timesfm}. The TiRex line is
the exception, mixing along time with xLSTM blocks instead
\citep{auer2025tirex, podest2026tirex2, beck2024xlstm}.

Most TSFMs are univariate\label{gl:univariate}. They take one
variate per series and run attention along the time axis only. We distinguish
multi-target\label{gl:multi-target} models, which forecast
several targets of one series in a single pass, from
multivariate\label{gl:multivariate} models, which also condition
on past or known-future covariates. TSFMs with covariate support add a second
attention along the variate axis, decoupled from the first. Chronos-2, TiRex-2,
and TimesFM-3.0 all alternate the two \citep{ansari2025chronos2,
podest2026tirex2, timesfm3}. A complementary line of work adds no native
covariate support, and instead adapts frozen univariate TSFMs to covariates
\emph{post hoc} \citep{arango2025chronosx, qin2025cora, auer2025cosmic,
das2024icf}.

TSFMs also differ in the form of their output, though here they appear to be
converging. One group predicts the parameters of a distribution and draws
sample paths from it. For instance, Toto took this route \citep{cohen2024toto}.
Most models now emit quantiles at fixed levels instead. The Chronos family,
TiRex-2, and Toto-2, the successor of Toto, all belong to this group
\citep{ansari2024chronos, ansari2025chronos2, podest2026tirex2,
khwaja2026toto2}. Quantiles are simpler and cheaper, since sampling adds a step
at inference, but they are marginal, one per horizon step. Predicting sample
paths is still advantageous where the joint structure across steps matters.

How a model is trained follows from these architectural choices. Encoder-only
models such as Chronos-2 borrow the architecture of masked-language
models, with a slightly different objective. Rather than masking patches at random and
reconstructing them, they split each training example into a context and a
future window. They then compute the quantile loss on the future window alone
\citep{ansari2025chronos2}. Decoder-only architectures such as Toto-2
rely on teacher forcing and next-patch prediction with causal attention along
the time axis \citep{khwaja2026toto2}. The two types of models blur under the contiguous patch
masking (CPM) introduced by TiRex, where spans of patches are blanked at random and the model
must fill them in or continue past them \citep{auer2025tirex}. Toto-2 and
TimesFM-3.0 both reuse it \citep{khwaja2026toto2, timesfm3}, as does the model introduced in this report.

In contrast to the model architecture choices, pretraining corpora are more
uniform across models. Most models mix public datasets, an internal pool for
the teams that hold one, and synthetic series. LOTSA \citep{woo2024moirai} and
GIFT-Eval Pretrain \citep{aksu2024gifteval} are the most common public
datasets. Toto also trains on internal observability telemetry
\citep{cohen2024toto}. The synthetic component follows the same handful of
recipes almost everywhere: Gaussian-process kernel composition
\citep{ansari2024chronos}, causal kernels \citep{xie2025cauker}, structural
priors, and random transforms. They are then combined in varying mixes by
TiRex, TS-ICL, and Chronos-2 \citep{auer2025tirex, lenaour2026tsicl,
ansari2025chronos2}.

\begin{figure}[tb]
  \centering
  \input{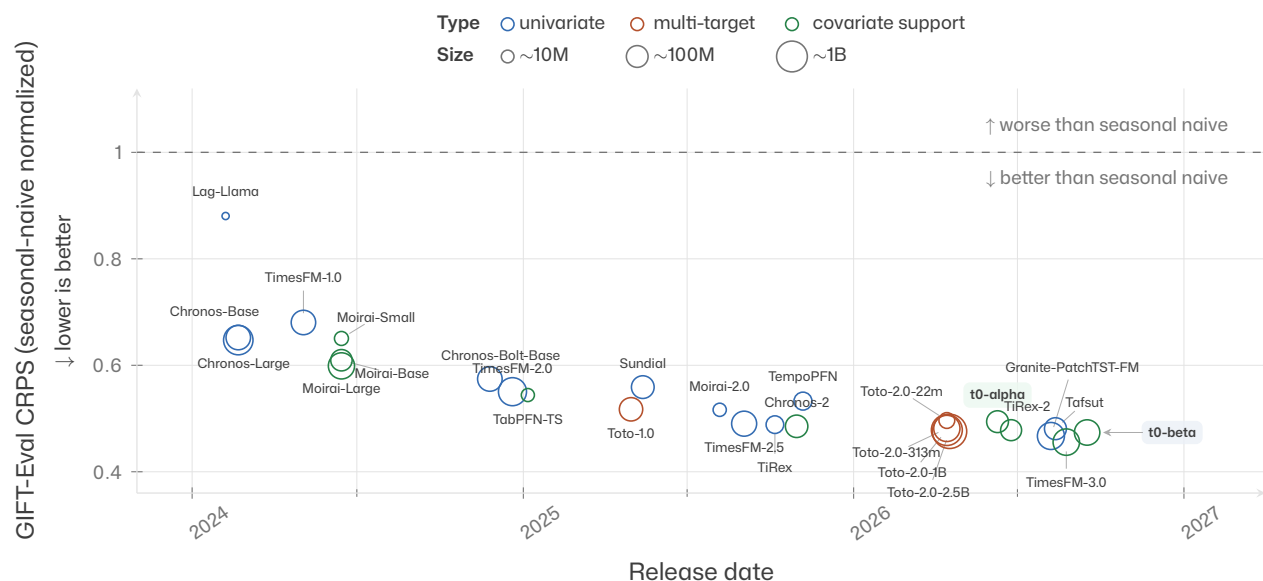}
  \caption{The TSFM landscape on the GIFT-Eval benchmark
  \citep{aksu2024gifteval}. Each model sits at its release date against its
  aggregate CRPS, normalized by the seasonal-naive baseline. Ring radius
  represents the parameter count on a log scale, from the 2.4M of Lag-Llama to
  the 2.45B of Toto-2.0-2.5B. Color informs the variate capability of the model:
  \capuni, one variate per series; \capmulti, several variates of one series
  forecast jointly in a single pass; or \capcov, covariates leveraged to
  forecast the target. \tzeroalpha and \tzerobeta are highlighted.
  }
  \label{fig:size-vs-crps}
\end{figure}

Finally, progress on TSFMs, as elsewhere in machine learning, is paced by benchmarks. The most widely adopted are GIFT-Eval \citep{aksu2024gifteval} and fev-bench
\citep{shchur2025fevbench}. The series of GIFT-Eval are mostly univariate and leave
covariates little if any role. The more recent fev-bench does include covariate
tasks, yet univariate models such as TiRex still sit near the top of its leaderboard.
This admits two readings: univariate accuracy comes first, a model
that reads covariates but forecasts the target poorly helps no one, and the benchmark does not isolate covariate skill as sharply as it could.
The field still has room for benchmarks that target specific capabilities, covariate use among them. Figure~\ref{fig:size-vs-crps} places the models of this
section on GIFT-Eval, evaluated with the continuous ranked probability
score\label{gl:crps} (CRPS), which compares a full
predictive distribution against the single observed value.

\section{The \tzero Model}
\label{sec:model}

We introduce \tzero, a model family and its architecture. This report describes
\tzeroalpha, the first released model of this family.

\subsection{Architecture}
\label{sec:architecture}

\begin{figure}[!tb]
  \centering
  \resizebox{\linewidth}{!}{\input{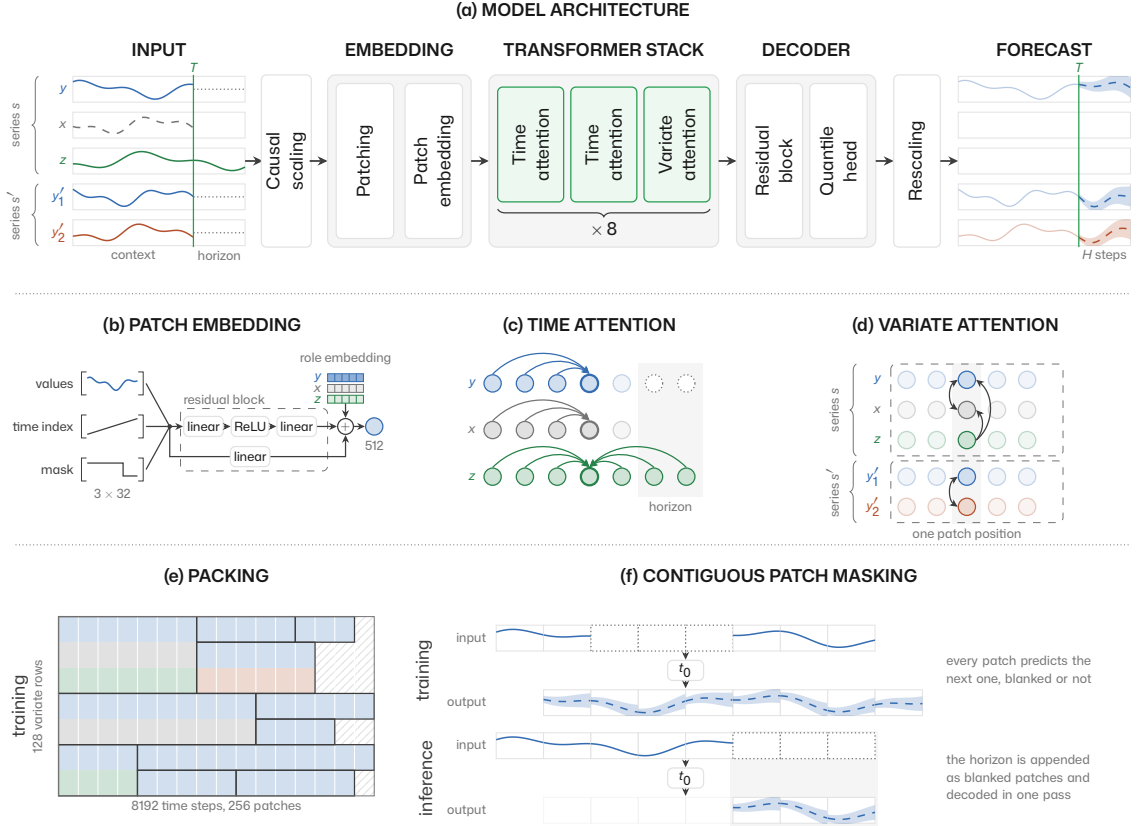}}
  \caption{The \tzero pipeline. \textbf{(a)} Time series are processed in
  parallel in one mini-batch. They are normalized with causal statistics, cut into
  patches, embedded (b), read by a stack alternating time (c) and variate (d)
  attention blocks, and decoded into quantiles, rescaled at inference into the
  forecast of the requested targets. \textbf{(b)} Embedding of one patch.
  \textbf{(c)} Time self-attention, causal for targets and historical
  covariates, bidirectional for known-future ones. \textbf{(d)} Variate
  self-attention, restricted to the variates of one time series. It is
  bidirectional between targets and past covariates, asymmetric for
  known-future ones, to avoid a two-hop
  leak of future information. \textbf{(e)} Packing of time series of
  different lengths into one mini-batch. The variate dimension is the batch
  dimension. Series boundaries are marked for attention and loss.
  \textbf{(f)} Contiguous patch masking, in training and at inference,
  following TiRex \citep{auer2025tirex}. The figure layout is inspired by Figure~1 of
  Chronos-2 \citep{ansari2025chronos2}.}
  \label{fig:architecture}
\end{figure}

\tzero is a decoder-style patch transformer with decoupled time and variate
attention. Figure~\ref{fig:architecture} gives an overview of the architecture.

\paragraph{Input preparation}
\label{sec:model_input}

Values are normalized one variate at a time, then cut into patches of 32
steps. Targets $\mathbf{y}$ and past covariates $\mathbf{x}$ are standardized
with \emph{causal} running statistics. The mean $\mu_t$ and standard deviation
$\sigma_t$ at step $t$ are computed from $y_{1:t}$ alone. An
$\operatorname{arcsinh}$ transform follows, which compresses heavy tails while
staying linear near zero. Chronos-2 popularized it for time series
\citep{ansari2025chronos2}. Known-future covariates $\mathbf{z}$ are observed
over the whole span $1{:}T{+}H$ by definition, so they are standardized with
statistics over that entire length.

In training, input and target are not scaled with the same statistics. For a
forecast issued at $T$, dropping the variate index $v$, the model reads the
left-hand side and predicts the right-hand side of
\begin{equation} \label{eq:scaling}
  \tilde{y}_t = \operatorname{arcsinh}\!\left(\frac{y_t -
  \mu_t}{\sigma_t}\right) \;\;\text{for } t \le T, \qquad \tilde{y}_{T+h} =
  \operatorname{arcsinh}\!\left(\frac{y_{T+h} - \mu_T}{\sigma_T}\right)
\;\;\text{for } h \in [H]. \end{equation} Each input step is scaled with the
statistics available at that step, its own value included, whereas every target
step is scaled with the statistics frozen at $T$, which never include the
target itself, since nothing later is available when the forecast is issued.
The model must therefore predict any change of level or spread over the horizon itself.
At inference, the prediction is brought back to the data space by inverting the same transform,
$\hat{y}^{q}_{T+h} = \mu_T + \sigma_T
\sinh\bigl(\hat{\tilde{y}}^{q}_{T+h}\bigr)$. No information past $T$ reaches
the model through scaling.

\paragraph{Missing values}

Time series often contain missing values. We track them with a binary mask alongside the values.
Every non-observed step is excluded from the normalization statistics and the loss.

\paragraph{Embedding}

Each patch is embedded from three channels of 32 entries: the standardized
values, a within-patch time index, and the validity mask of every step. The
time index ranges linearly from $0$ to $31/32$ over the patch. The three are
concatenated and projected by a residual block. The output is added to a
learned embedding of the role of the variate (either target $\mathbf{y}$, past
covariate $\mathbf{x}$, or known-future covariate $\mathbf{z}$). The encoder is
adapted from Chronos-2 \citep{ansari2025chronos2}. No timestamp or
frequency feature is required by the encoder.

\paragraph{Attention layers}

Covariate support hinges on the layer stack at the heart of \tzero, which alternates time and variate attention.
Both operate on the embedded patches rather than on the original time steps.

\textbf{Time attention} is causal self-attention along the patch sequence of
each variate. It uses rotary position embeddings \citep{su2024roformer} indexed
by patch position. There is one rotary unit per patch. The offset of the time step
within a patch is only considered by the model through the embedding. The
rotary embedding scales queries and keys by reciprocal position-dependent
factors \citep{sun2023xpos}, so that each attention score decays exponentially
with the distance between the two patches. Far-away patches thus weigh less,
which keeps attention stable on contexts longer than any seen in training. The
time attention is causal for target $\mathbf{y}$ and past covariates
$\mathbf{x}$, while known-future $\mathbf{z}$ attend bidirectionally over
$1{:}T{+}H$. It is how accessible information placed over the horizon is read.

\textbf{Variate attention} is self-attention across variates at each patch
position, with no positional encoding to remain permutation equivariant since
variates carry no natural order. The attention is asymmetric. While target
$\mathbf{y}$ and $\mathbf{x}$ attend to one another, known-future $\mathbf{z}$
are read by the two other roles without attending back. This avoids a two-hop
future information leakage that the alternating attention would otherwise open.
Indeed, a known-future $\mathbf{z}$ attends over its
whole span, so had it absorbed target content at some step $s$, a later time
layer would carry that content to every earlier step $t < s$, and a later
variate layer would hand it to the target at $t$.

Attention layers are arranged in a repeating pattern of two time-attention
layers followed by one variate-attention layer. \tzeroalpha has 24 attention
layers in total: 16 for time, 8 for variates. Each attention block uses
pre-norm with RMSNorm \citep{zhang2019rmsnorm} and is followed by a SwiGLU
residual feed-forward layer \citep{shazeer2020glu}. They also normalize queries
and keys per head before the attention product to prevent the softmax from
saturating, following QK-norm \citep{henry2020qknorm, dehghani2023vit22b}. A
final RMSNorm closes the stack.

\paragraph{Decoder}

The decoder turns the representation of a patch into a forecast. It is a
residual block, a two-layer perceptron with a linear skip connection, applied
to every patch in parallel. Under causal attention, the representation of the
patch ending at step $T$ summarizes everything observed up to $T$. The decoder
reads it and predicts the next patch, the 32 steps $T{+}1, \dots, T{+}32$, at
the five native quantile levels $\mathcal{Q}_0 = \{0.1, 0.25,
0.5, 0.75, 0.9\}$. Every patch boundary therefore sits at information cutoff
date. The output $\hat{\tilde{y}}^{(v),q}_{T+h}$ is rescaled to the data
space as described in Section~\ref{sec:model_input}.

Quantile crossing is a standard defect of independently estimated quantiles,
usually repaired after the fact by rearrangement
\citep{chernozhukov2010rearrangement}. The decoder instead makes the five
quantiles monotone by construction. It predicts the lowest level directly and
every other level as a positive increment on the level below. For consecutive
levels $q_k < q_{k+1}$ of $\mathcal{Q}_0$, $\hat{\tilde{y}}^{\,q_{k+1}} =
\hat{\tilde{y}}^{\,q_k} + \operatorname{softplus}(r_{k+1})$, where $r$ is the
raw decoder output and $\operatorname{softplus}(u) = \log(1 + e^{u})$ is
strictly positive.

\subsection{Training}
\label{sec:model_training}

\paragraph{Training data}
\label{sec:data}

\tzeroalpha is pretrained on five corpora (Table~\ref{tab:corpora}), one
real and four synthetic. While pretraining on synthetic data alone is an established alternative
\citep{dooley2023forecastpfn, moroshan2025tempopfn, hollmann2025tabpfnv2,
xie2025cauker}, like most forecasting models, \tzeroalpha mixes synthetic and real
data.

The real corpus is the GIFT-Eval pretraining split \citep{aksu2024gifteval}.
Its 89 datasets span domains like energy, cloud operations, transport, sales,
climate, and healthcare. We rebalance the corpus. The ERA5 and CMIP6 climate
collections were excluded for their size and homogeneity, and one
cryptocurrency series for its extreme variability.
Most of its series are univariate.
Past covariates account for about a sixth of its values, and fewer than one
series in ten carries more than one target.

The four synthetic corpora supply most of the multivariate series, each from
one generator family. \emph{Kernel-composition} generators sample Gaussian
processes with randomly composed kernels, in the spirit of KernelSynth
\citep{ansari2024chronos}. \emph{Causal-graph} generators sample structural causal models in which nodes are time series and edges define causal dependencies among them. Nodes are then randomly assigned to targets, past covariates, or known-future covariates. This allows for the simulation of complex yet realistic data-generating processes, where each time series results from multiple interacting causes (see, e.g., \citep{lenaour2026tsicl, xie2025cauker, apollopfn2026time} for more details).
\emph{Covariate-effect} generators build multivariate samples from
hand-designed effect families with known ground truth: level shifts and
closures, spikes, interactions between covariates, lag and lead responses,
and decoy covariates carrying no signal, with both past and known-future
covariates. A \emph{mixing} augmentation combines real series into convex
mixtures, in the spirit of TSMixup \citep{ansari2024chronos}.

\begin{table}[htb]
  \centering
  \small
  \caption{Training corpora of \tzeroalpha and probability of each data pool over three different phases of the training.
  Series and target steps are computed as the total for each corpus.}
  \label{tab:corpora}
  \setlength{\tabcolsep}{4.5pt}
  \begin{tabular}{llrrrrr}
    \toprule
    & & & & \multicolumn{3}{c}{Draw probability} \\
    \cmidrule(l){5-7}
    Corpus & Covariates & Series & Target steps & 0--40k & 40--80k & 80--150k \\
    \midrule
    GIFT-Eval pretrain, rebalanced & past, in part & 2.97M & 24.9B & 12\% & 10\% & 8\% \\
    Mixing & none & 10.0M & 21.2B & 39\% & 35\% & 26\% \\
    Kernel composition & none & 10.0M & 81.9B & 39\% & 35\% & 26\% \\
    Causal graph & past and future & 1.00M & 22.5B & 9\% & 18\% & 36\% \\
    Covariate effects & past and future & 0.10M & 0.11B & 1\% & 2\% & 4\% \\
    \bottomrule
  \end{tabular}
\end{table}

The mix of the five corpora changes over training, in the manner of
curriculum learning \citep{bengio2009curriculum}. Series are drawn from two
pools. A multivariate pool is made of the causal-graph and covariate-effect
corpora, and a pool of the other three, which is mostly univariate. A draw
first picks a pool, then a series uniformly within it, so a corpus weighs in
proportion to its series count within its pool. The multivariate pool
receives 10\% of the draws over the first 40k steps, 20\% over the next
40k, and 40\% from step 80k on (Table~\ref{tab:corpora}). The model is thus
first exposed to mostly univariate series, and the share of multivariate
ones grows from a tenth to two fifths. The schedule rests on the assumption
that leveraging multivariate information is the harder skill, and easier to
learn once the behavior of a single series is captured.

Each series undergoes a few transformations on the fly. A window of 8192 steps
is cut at a random position, or the whole series when shorter. Its edges are
jittered by up to one patch, so during training the model sees series
padded at their start. Patch boundaries also fall on varying steps of
the same time series whenever it is sampled multiple times. Missing values are
flagged in the mask and their slot filled with zero.

\paragraph{Loss objective}

Since the model predicts quantiles directly, it is trained with the quantile
loss that underlies the CRPS, as are Chronos-2 and the TiRex models
\citep{ansari2025chronos2, auer2025tirex, podest2026tirex2}. Every patch
boundary of a target variate is an information cutoff datetime $T$ at which
the decoder predicts the next 32 steps at the five native levels. Each predicted
step is scored against its observation in the normalized space of
\eqnref{eq:scaling}, both scaled with the statistics frozen at $T$. The CRPS of
a quantile forecast is an integral of pinball losses over quantile levels
\citep{laio2007verification, gneiting2011comparing}, which we discretize over
$\mathcal{Q}_0$ by the centered rectangle rule. Dropping the variate index as
in \eqnref{eq:scaling}, the loss of step $h$ is \begin{equation}
  \label{eq:loss} \ell_{T,h} = \sum_{q \in \mathcal{Q}_0} 2\,
  w_q\, \rho_q\!\left(\tilde{y}_{T+h} - \hat{\tilde{y}}^{\,q}_{T+h}\right),
\qquad \rho_q(u) = u\,\bigl(q - \mathbf{1}\{u < 0\}\bigr), \end{equation} where
$\rho_q$ is the pinball loss \citep{koenker1978quantiles}. The weight $w_q$
of a level is the width of the cell between the midpoints to its neighbors,
the two outer cells extended to $0$ and $1$. $w = (0.175,
0.2, 0.25, 0.2, 0.175)$, summing to one. The pinball losses at $0.1$ and $0.9$
thus stand in for the whole tails, and nothing in training asks the model to
place a level beyond them.

The model is trained for next-patch prediction. Every patch of a series, but
the first, is considered as the target predicted from the context of the
previous patches. The loss is computed on the time steps of the target patch.
Each of the 32 steps of a predicted patch is scored against its observed value
with \eqnref{eq:loss}. Only target variates are scored. The model also predicts
the past covariates, but their outputs are masked out of the loss, so a
covariate influences a forecast through attention alone. Missing values and
padded steps are discarded from the loss. The loss is then averaged per
variate row of the mini-batch, then across rows. With $\mathcal{S}_r$ the
scored positions $(T, h)$ of row $r$ and $\mathcal{R}$ the rows holding at
least one, \begin{equation} \label{eq:loss_reduction}
  \mathcal{L} = \frac{1}{|\mathcal{R}|} \sum_{r \in \mathcal{R}}
  \frac{1}{|\mathcal{S}_r|} \sum_{(T,h) \in \mathcal{S}_r} \ell_{T,h}.
\end{equation}

\paragraph{Packing}

As the length of time series varies across samples and domains, the training
pipeline must cope with this variability. A simple strategy is to pad every
series to a maximum length, which wastes compute on padding. Instead, following
large language model training practice \citep{raffel2020t5, krell2021packing,
ding2024fewer}, time series are packed during training, as in
Figure~\ref{fig:architecture}(e).
Language model packing fills one axis, the sequence of tokens, because text is
one-dimensional. \tzero packs in two dimensions. A time series is a 2D block
whose dimensions are variate and time. A series with $n$ variates and $\ell$
steps occupies a rectangle of $n$ rows by $\ell$ columns. During training, time
series are tiled in a pack of 128 variate rows by 8192 time steps. Several
series land side by side along the time axis and several are stacked along the
variate axis. Building a mini-batch this way is a two-dimensional bin-packing
problem \citep{lodi2002twodim}. For \tzeroalpha, the packer draws a buffer of
candidate series, fills the grid with those that fit the space left, and
discards the remainder.
Figure~\ref{fig:architecture}(e)
shows one such grid. A multivariate series occupies a block of several rows, a
univariate one a single row, and the cells no series fits are left unused.

The packing idea for time series originates from Toto, which stacks independent
multi-target time series along the variate axis \citep{cohen2024toto}. To avoid
the spill of information from one time series to another, the packing adds a
mask identifying the time series each time step of a mini-batch belongs to,
similarly to Chronos-2 \citep{ansari2025chronos2}. This mask is used both by
the attention layers and the loss computation. Additionally, each time series
is padded to a multiple of the patch size, so every patch includes information
about a single time series. To our knowledge, \tzero is the first model
family to rely on this two-dimensional time series packing for training.

\paragraph{Contiguous patch masking}

The training employs the contiguous patch masking strategy introduced by TiRex
\citep{auer2025tirex} and adopted since by TiRex-2, Toto-2, and TimesFM-3.0
\citep{podest2026tirex2, khwaja2026toto2, timesfm3}. Contiguous spans of target
patches are blanked from the input, as in Figure~\ref{fig:architecture}(f),
so the model has to reconstruct them from the surrounding context. Spans of one to sixteen patches are blanked at random
on each target row independently. The training objective, however, remains the
next-patch prediction loss. It is computed at every patch boundary, whether or
not the patches read by the model were blanked, which keeps a high proportion
of patches contributing to the loss. Figure~\ref{fig:architecture}(f) shows
one blanked span and the prediction every patch still emits for the next.

\paragraph{Training run}

\tzeroalpha was trained from scratch on a single node with four NVIDIA H100 80\,GB
GPUs, using data parallelism and mixed-precision \texttt{bfloat16}. The
released checkpoint was saved at step 150k after 28 hours.
Each GPU
processes one pack of 128 variate rows by 8192 steps per optimization step.
The model has been trained on $5.4 \times 10^{11}$ observed time steps,
counting every variate of every series, targets and covariates alike. The
optimizer is AdamW \citep{loshchilov2019adamw} with a peak learning rate of $3
\times 10^{-4}$, a weight decay of $0.01$. The learning rate follows a linear warmup over the first 5k steps,
then a cosine decay down to $5 \times 10^{-5}$ at step 150k. Gradient clipping was set at a global
norm of $1$.

\subsection{Inference}
\label{sec:inference}

At inference, the model must produce the forecast $\hat{y}^{(v),q}_{T+h}$ over
a prediction length of $H$ steps (\eqnref{eq:forecast}). Because it was trained with
contiguous patch masking, two regimes are distinguished by this length. When
$H$ is at most 1024 steps, the horizon is appended to the context as blanked
patches and a single forward pass yields the quantiles at every step $h \in
[H]$ and every native level $q \in \mathcal{Q}_0$, the inference half of
Figure~\ref{fig:architecture}(f). This already extrapolates
the training regime, where at most 512 target steps were blanked at once. When $H$ is larger,
the model performs a rollout, following Toto-2 \citep{khwaja2026toto2}.
Forecasts are produced by chunks of 1024 steps, each chunk being folded back
into the context, until the full horizon is covered.

In a rollout, the context must be extended with the previous forecast. Rather
than feed a single point path back, we replicate the context into
$|\mathcal{Q}_0| = 5$ paths, one per native level, and the path of level $q$ is
extended with its own quantile predictions $\hat{y}^{(v),q}$. So the
uncertainty of one chunk is carried into the next. Each path then yields five
quantiles. These values are pooled, weighted by the probability mass of their
path and of their level, and reduced back to the five native levels as weighted
quantiles. This reduced prediction is merged into the previous context and fed
back to the model for the next iteration. The rollout thus follows several
paths and reduces them at every chunk. This is the strategy of Chronos-2
\citep{ansari2025chronos2}, whereas Toto-2 feeds only the median back
\citep{khwaja2026toto2}. Reducing to the median is simpler, but it can
oversmooth the forecast. Following one path per quantile level is expected to
limit this smoothing, at the cost of a more expensive inference, since the
batch is inflated by the number of paths.

The full stack has approximately 102M parameters.
Table~\ref{tab:arch} summarizes the configuration of \tzeroalpha.

\begin{table}[htb]
  \centering
  \caption{\tzeroalpha architecture summary.}
  \label{tab:arch}
  \begin{tabular}{ll}
    \toprule\tfcstripes
    Parameters & $\approx$102M \\
    Layers & 24 (16 time-attention, 8 variate-attention) \\
    Layer pattern & 2 time layers, then 1 variate layer \\
    Embedding dim & 512 \\
    Feed-forward dim & 2048 (SwiGLU) \\
    Attention heads & 8 \\
    Patch size & 32 \\
    Norm & pre-norm RMSNorm, QK-norm, final RMSNorm \\
    Quantile head & monotone by construction (cumsum-of-softplus) \\
    Positional encoding & RoPE by patch position (time axis only) \\
    Native quantile levels & 0.1, 0.25, 0.5, 0.75, 0.9 \\
    Single-pass horizon & up to 1024 steps \\
    Trained blanked span & up to 512 steps \\
    \bottomrule
  \end{tabular}
\end{table}

\addtocontents{toc}{\protect\columnbreak}
\section{Evaluation}
\label{sec:evaluation}

\subsection{Public benchmarks}
\label{sec:eval-protocol}

We evaluate \tzeroalpha on three public benchmarks: GIFT-Eval for univariate
accuracy, fev-bench for accuracy with covariates, and TIME for accuracy on
datasets assembled to lie outside public pretraining corpora.

The comparison is against other TSFMs. Every rank we report is to be understood
within the list of released zero-shot checkpoints. The public leaderboards are
wider. They include traditional statistical systems that are systematically outperformed, but also agentic systems, ensembles and fine-tuned entries.
These systems generally route, ensemble or search over one or more pretrained
forecasters rather than provide forecasts directly themselves. \tzeroalpha is a candidate
component of such a pipeline rather than an alternative to one, so we leave
those entries out of the ranking.

The results of this section mostly concern \tzeroalpha. We also report
results on the public benchmarks for \tzerobeta, its successor. When no
model is named, a result concerns \tzeroalpha.
Among the released zero-shot models of the three tables, \tzeroalpha ranks 11th
of 17 on GIFT-Eval by CRPS, 12th of 20 on fev-bench by skill score and 8th of
12 on TIME by CRPS. \tzerobeta would enter the same tables third, third and
fifth, and it holds third place on GIFT-Eval under MASE as well as CRPS.

A rank alone hides how close these models are. TimesFM-3.0 leads GIFT-Eval at
0.4557 CRPS, and \tzerobeta passes Toto-2.0-2.5B by 0.0021 of CRPS, 0.45\% in
relative terms, and under MASE as well, with a tenth of its
parameter count. \tzerobeta is 4.0\% of CRPS short of the top of GIFT-Eval,
2.0 skill points short of the top of fev-bench and 3.5\% of CRPS short of
the top of TIME, and \tzeroalpha is 5.8\% of CRPS short of
second place on GIFT-Eval.

The model runs the same way on the three benchmarks. It reads a context of at most 8192
steps. Longer time series are trimmed to their last 8192 steps. Shorter ones
are left-padded to the next multiple of the 32-step patch. The model returns
the five native quantiles that are used for evaluation. Whenever a benchmark
asks for another level, it is obtained by interpolation between the two
neighboring native levels. Point metrics such as MASE are computed on the
median.

One and only checkpoint serves the three benchmarks, as it does for most of the
models we compare with. TiRex-2 is however an exception. It enters GIFT-Eval
and fev-bench with a different checkpoint for each. It is the same architecture
but trained on different data, so its rows in the two tables are less
comparable than the others. Additionally, some models like Granite-PatchTST-FM
or TiRex-2 do not appear in all the three benchmarks. When it is the case, we
report model results only for the benchmark they have been publicly run on.

Several studies below run on a 12-pair subset of GIFT-Eval spanning all
seven domains and frequencies from 15-minute to monthly, chosen so that
peer checkpoints could be run under identical conditions within our compute
budget. Table~\ref{tab:gifteval-subset} in
Appendix~\ref{app:gifteval-subset} lists the twelve datasets.

\subsubsection{GIFT-Eval}
\label{sec:gifteval}

GIFT-Eval \citep{aksu2024gifteval} has become the standard benchmark for
pretrained forecasting models. It comprises 97 forecasting tasks built from 23
datasets across seven domains, with frequencies from seconds to years and
univariate as well as multivariate series. A task pairs a dataset, sampled at
one frequency, with a forecast term that sets the horizon length to predict.
Each task provides a context from which to predict and a target to compare the
forecast against.

We compute two metrics, CRPS and MASE, which measure the accuracy of the
  probabilistic forecast and of the median point forecast, respectively. MASE is the mean absolute error of the forecasted median over the target
  windows, scaled per series by the in-sample mean absolute error of the seasonal
  naive forecast. This scaling makes errors comparable across series of different
  magnitude. The CRPS is approximated using the weighted quantile loss and  normalized by the target magnitude: for each of nine
  quantile levels, $q \in \{0.1, 0.2, \ldots, 0.9\}$, twice the pinball loss is
  summed over every series and horizon step of a dataset and divided by the
  summed absolute value of the target; the nine values are then averaged. Each metric is then
  normalized by the value the seasonal naive forecast obtains on the same dataset,
  frequency and horizon configuration. Consequently, an error of 1 indicates the
  method matches the baseline, and lower is otherwise better. The overall score
  is the geometric mean of these normalized values over the 97 configurations.

GIFT-Eval carries covariates on few of its series, but we use none. Every
variate is forecast on its own, from its own past. We therefore use GIFT-Eval
to measure univariate zero-shot accuracy. Consequently, the model could even
get better results by using the covariates that are present. On this protocol
\tzeroalpha reaches an aggregate CRPS of 0.4941 and a MASE of
0.7240. Table~\ref{tab:gifteval}
  places
  these scores next to those of the other zero-shot models, sorted by CRPS,
  under which \tzeroalpha is 11th of the 17 leaderboard models. Under MASE it
  would be 12th. \tzerobeta reaches a
  CRPS of 0.4738 and a MASE of 0.6865 on the same 97 pairs. It enters
  Table~\ref{tab:gifteval} third by CRPS and third by MASE, behind only
  TimesFM-3.0 and Granite-PatchTST-FM. It passes Toto-2.0-2.5B by 0.45\% of
  CRPS, and under MASE as well, while carrying a tenth of its
  parameters. Against \tzeroalpha it lowers CRPS by 4.1\% and MASE by 5.2\%.

\begin{table}[!ht]
  \centering
  \caption{GIFT-Eval aggregated CRPS and MASE results (lower is better).
    Each row corresponds to a zero-shot model. Rows are sorted by CRPS.
    The best value of each column is in bold. \tzeroalpha is shaded green
    and \tzerobeta, its successor, blue.}
  \label{tab:gifteval}
\begin{tabular}{lrrr}
\toprule
\tfcheadrow{Model} & \tfcheadrow{Params} & \tfcheadrow{CRPS $\downarrow$} & \tfcheadrow{MASE $\downarrow$} \\
\midrule
TimesFM-3.0 & 331M & \textbf{0.4557} & \textbf{0.6668} \\
Granite-PatchTST-FM & 385M & 0.4672 & 0.6846 \\
\rowcolor{tfcbandblue} \tzerobeta & 256M & 0.4738 & 0.6865 \\
Toto-2.0-2.5B & 2.45B & 0.4759 & 0.6956 \\
TiRex-2 & 82.5M & 0.4781 & 0.6973 \\
Toto-2.0-1B & 1.04B & 0.4784 & 0.6992 \\
Tafsut & 105M & 0.4809 & 0.6926 \\
Toto-2.0-313m & 313M & 0.4814 & 0.7028 \\
Chronos-2 & 119M & 0.4854 & 0.6978 \\
TiRex & 35M & 0.4885 & 0.7158 \\
TimesFM-2.5 & 231M & 0.4903 & 0.7050 \\
\rowcolor{tfcband} \tzeroalpha & 102M & 0.4941 & 0.7240 \\
Toto-2.0-22m & 21.9M & 0.4963 & 0.7188 \\
Moirai-2.0 & 11.4M & 0.5164 & 0.7281 \\
Toto-1.0 & 151M & 0.5173 & 0.7501 \\
TempoPFN & 38M & 0.5327 & 0.7875 \\
TabPFN-TS & 11M & 0.5441 & 0.7709 \\
Sundial & 128M & 0.5590 & 0.7502 \\
Seasonal naive & --- & 1.0000 & 1.0000 \\
\bottomrule
\end{tabular}

\end{table}

\begin{figure}[!ht]
  \centering
%
\pgfplotsset{colormap={tfcwinloss}{rgb255(0cm)=(180,81,45) rgb255(1cm)=(240,239,236) rgb255(2cm)=(38,130,72)}}
\begin{tikzpicture}
  \begin{groupplot}[
      group style={group size=2 by 1, horizontal sep=2.8cm},
      tfcaxis, axis line style={draw=none}, tick style={draw=none},
      axis x line=top, axis y line=left,
      scale only axis, width=4.10cm, height=3.50cm,
      xmin=-0.5, xmax=4.5, ymin=-0.5, ymax=6.5,
      xtick={0,1,2,3,4}, ytick={0,1,2,3,4,5,6},
      xticklabels={TiRex-2, Toto-2.0, Chronos-2, TiRex, TimesFM-2.5},
      x tick label style={rotate=45, anchor=west, font=\scriptsize, color=tfcbody},
      y tick label style={font=\scriptsize, color=tfcbody},
      grid=minor, minor tick num=1, minor grid style={draw=white, line width=0.9pt},
      axis on top,
      colormap name=tfcwinloss, point meta min=0, point meta max=1,
      clip=false,
    ]
    \nextgroupplot[ylabel={Domain}, yticklabels={Sales (4), Healthcare (5), Econ/Fin (6), Transport (15), Nature (15), Web/CloudOps (20), Energy (32)}]
      \addplot[matrix plot*, point meta=explicit, mesh/cols=5]
        table[col sep=comma,x=x, y=y, meta=domain_win_rate]{figures/win-loss-map.csv};
      \addplot[only marks, mark=none, point meta=explicit,
               nodes near coords={\pgfmathprintnumber[fixed, fixed zerofill, precision=2]{\pgfplotspointmeta}},
               every node near coord/.append style={anchor=center, font=\tiny, text=tfcink},
               restrict expr to domain={\thisrow{domain_white_text}}{0:0}]
        table[col sep=comma,x=x, y=y, meta=domain_win_rate]{figures/win-loss-map.csv};
      \addplot[only marks, mark=none, point meta=explicit,
               nodes near coords={\pgfmathprintnumber[fixed, fixed zerofill, precision=2]{\pgfplotspointmeta}},
               every node near coord/.append style={anchor=center, font=\tiny, text=white},
               restrict expr to domain={\thisrow{domain_white_text}}{1:1}]
        table[col sep=comma,x=x, y=y, meta=domain_win_rate]{figures/win-loss-map.csv};
    \nextgroupplot[ylabel={Frequency}, yticklabels={Yearly (1), Quarterly (1), Monthly (5), Weekly (8), Daily (15), Hourly (31), Sub-hourly (36)}]
      \addplot[matrix plot*, point meta=explicit, mesh/cols=5]
        table[col sep=comma,x=x, y=y, meta=frequency_win_rate]{figures/win-loss-map.csv};
      \addplot[only marks, mark=none, point meta=explicit,
               nodes near coords={\pgfmathprintnumber[fixed, fixed zerofill, precision=2]{\pgfplotspointmeta}},
               every node near coord/.append style={anchor=center, font=\tiny, text=tfcink},
               restrict expr to domain={\thisrow{frequency_white_text}}{0:0}]
        table[col sep=comma,x=x, y=y, meta=frequency_win_rate]{figures/win-loss-map.csv};
      \addplot[only marks, mark=none, point meta=explicit,
               nodes near coords={\pgfmathprintnumber[fixed, fixed zerofill, precision=2]{\pgfplotspointmeta}},
               every node near coord/.append style={anchor=center, font=\tiny, text=white},
               restrict expr to domain={\thisrow{frequency_white_text}}{1:1}]
        table[col sep=comma,x=x, y=y, meta=frequency_win_rate]{figures/win-loss-map.csv};
  \end{groupplot}
  \begin{axis}[
      at={($(group c1r1.south east)!0.5!(group c2r1.south west)$)}, anchor=north, yshift=-6pt,
      hide axis, scale only axis, width=0pt, height=0pt, xmin=0, xmax=1, ymin=0, ymax=1,
      colormap name=tfcwinloss, point meta min=0, point meta max=1, colorbar/width=0.2cm,
      colorbar horizontal,
      colorbar style={
        at={(0.5,0)}, anchor=north, width=6.0cm, xtick={0,0.25,0.5,0.75,1},
        xticklabel style={/pgf/number format/fixed, /pgf/number format/precision=2},
        every tick label/.append style={font=\scriptsize, color=tfcmuted},
        tick style={draw=none}, axis line style={draw=tfchair, line width=0.4pt},
        xlabel={\tzeroalpha{} win rate against the column model \textcolor{tfcmuted}{(0.5 is parity)}},
        xlabel style={font=\scriptsize, color=tfcbody},
      },
    ]
    \addplot[draw=none] coordinates {(0,0)};
  \end{axis}
\end{tikzpicture}
  \caption{\tzeroalpha{} per-pair win rate against the five models
    ranked above it, by domain and by frequency class on GIFT-Eval. Green: \tzeroalpha{} wins the
    majority. Brown: it loses the majority. 0.5 is parity. Pair counts
    per group in parentheses.}
  \label{fig:win-loss}
\end{figure}
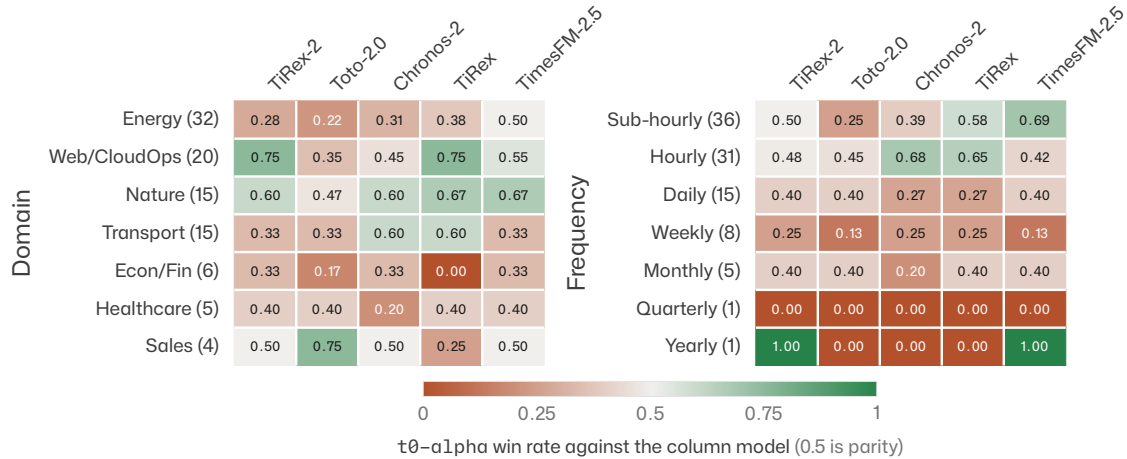

GIFT-Eval aggregates over tasks to produce a final benchmark value. However,
investigating the individual tasks underlying this value can provide additional
insights. We therefore split the benchmark by domain and frequency and compare
\tzeroalpha with five models ranked above it. For each model, we compute the
win rate of \tzeroalpha, defined as the fraction of the 97 dataset/term pairs
in which its CRPS is strictly lower than that of the compared model, broken
down by domain and frequency class
(Figure~\ref{fig:win-loss}).
Over all pairs, \tzeroalpha performs on par with TiRex (win rate 0.51, CI
0.40--0.61) and TimesFM-2.5 (0.49, CI 0.39--0.60), and behind TiRex-2 (0.45, CI
0.35--0.56), Chronos-2 (0.43, CI 0.33--0.54) and Toto-2.0 (0.33, CI
0.24--0.42). Only the Toto-2.0 deficit is statistically unambiguous, its CI
lying entirely below parity. By domain, \tzeroalpha beats both TiRex
generations on web and cloud-operations telemetry and on nature data, and
Toto-2.0 on sales. By frequency, it holds its own at hourly resolution, where
it beats Chronos-2 and TiRex, but loses ground at sub-hourly resolution against
Toto-2.0 and Chronos-2, and at daily, weekly and monthly frequencies against
every model. However, it is not possible to draw more general conclusions for
quarterly and yearly frequencies, which contain only one dataset/term pair
each. Stratifying the ranking of Table~\ref{tab:gifteval}
by
horizon term adds one more fact: among its 17 leaderboard models, \tzeroalpha
ranks 12th on the 55 short-term pairs, 9th on the 21 medium-term pairs and 6th
on the 21 long-term pairs. The model climbs the table as horizons lengthen,
consistent with single-pass decoding of long horizons
(Section~\ref{sec:inference}), which suggests that
its long forecasts are more stable than those of its competitors.

\subsubsection{fev-bench}
\label{sec:fevbench}

fev-bench \citep{shchur2025fevbench} complements GIFT-Eval for multivariate
evaluation. Whereas GIFT-Eval scores every variate on its own, 68 of the 100
fev-bench tasks carry covariates or several targets. 42 tasks declare past or
known-future covariates and 35 several targets, with some overlap between the
two. Regarding data, fev-bench also extends GIFT-Eval. It reuses some datasets but
also brings new time series to evaluate on. Each task is scored by the scaled
quantile loss, which corresponds to the quantile loss over the same nine
levels, divided per series by the in-sample mean absolute seasonal difference.
It has the same denominator as the MASE for GIFT-Eval. The benchmark reports two
aggregates. The skill score of a model and the win rate. The former is one
minus the geometric mean over tasks of the ratio between its loss and that of
the seasonal naive baseline, each ratio clipped to $[0.01, 100]$ before
aggregation. It is reported as a percentage, 0 matches the baseline and higher
is better. The latter is the share of task and opponent pairs in which the model
attains a lower loss than the opponent, a tie counting one half. It depends on
the models on the board and moves as entries are added, while the skill score
remains fixed. Two imputation rules apply to every entry: a task a model fails
is scored as seasonal naive, and a task whose dataset lies in the declared
pretraining corpus of the model is scored as Chronos-Bolt. \tzeroalpha declares
eight such tasks, drawn from the GEFCom load, M5, Favorita, KDD Cup 2022 and
FRED-MD datasets, so its skill score under the rule of the benchmark, 42.2, is
below the 42.9 its own forecasts obtain.

\begin{figure}[!ht]
  \centering
  \includegraphics[width=\linewidth]{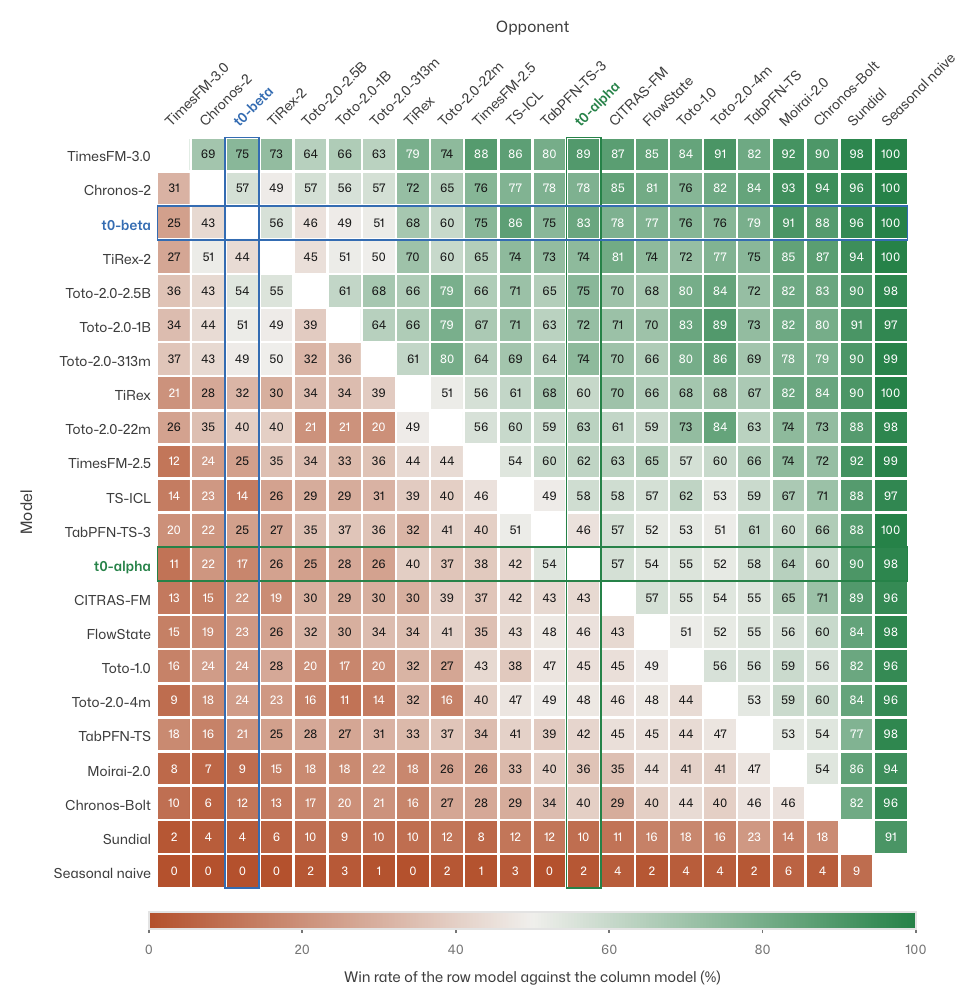}
  \caption{Pairwise win rates on fev-bench, all 100 tasks. It represents the share of
    tasks on which the row model attains a lower scaled quantile loss than
    the column model, a tie counting one half. The
    \tzeroalpha row and column are outlined in green, those of \tzerobeta
    in blue.}
  \label{fig:fevbench-pairwise-winrate}
\end{figure}

We use fev-bench to test how the model leverages covariates, following the
design of the covariate ablation of Chronos-2 \citep{ansari2025chronos2}. We run the
same checkpoint twice on every task, once with the time series covariates
passed and once with them dropped, so the difference isolates the covariate
pathway. On the 58 tasks without covariates the two runs perform the exact
same pass, which we verified exactly, so the effect rests on the 42
covariate-informed tasks. With covariates, \tzeroalpha attains a skill score of
42.2\% and a win rate of 59.6\% on all 100 tasks, and 39.7\% and 62.8\%, respectively, on the
42 covariate-informed tasks (Table~\ref{tab:fevbench}). \tzerobeta, whose pretraining corpora were built to exclude every dataset
declared by the benchmark, is zero-shot on all 100 tasks. Across all tasks,
it attains 46.7\% in skill score and 78.8\% in win rate, and on the
covariate-informed tasks, 46.2\% and 84.2\%, respectively. It ranks third by
skill score on both slices, behind only TimesFM-3.0 and Chronos-2, and third
by win rate on the covariate-informed tasks. The win rates count both of our
models as entries on the leaderboard, so each is an opponent of the other:
\tzerobeta beats \tzeroalpha on 83 of the 100 tasks. We refer to
Figure~\ref{fig:fevbench-pairwise-winrate} for a detailed breakdown of the
win rate by opponent.

\begin{table}[!ht]
  \centering
  \caption{fev-bench skill score and win rate, in percent, on all 100 tasks
    and on the 42 covariate-informed tasks (higher is better).
    Win rates are computed over the whole fev-bench field, with
    \tzeroalpha and \tzerobeta added to it so that each is an opponent of the
    other. Each row corresponds to a zero-shot model. Rows are sorted by skill
    score on all tasks. The best value of each column is in bold.
    \tzeroalpha is shaded green and \tzerobeta, its successor, blue.
    }
  \label{tab:fevbench}
\begin{tabular}{lrrrr}
\toprule
 & \multicolumn{2}{c}{\tfcheadrow{All tasks (100)}} & \multicolumn{2}{c}{\tfcheadrow{Covariate-informed (42)}} \\
\cmidrule(lr){2-3} \cmidrule(lr){4-5}
\tfcheadrow{Model} & \tfcheadrow{Skill (\%) $\uparrow$} & \tfcheadrow{Win rate (\%) $\uparrow$} & \tfcheadrow{Skill (\%) $\uparrow$} & \tfcheadrow{Win rate (\%) $\uparrow$} \\
\midrule
TimesFM-3.0 & \textbf{48.7} & \textbf{86.8} & \textbf{48.2} & \textbf{88.1} \\
Chronos-2 & 47.3 & 81.2 & 47.0 & 85.0 \\
\rowcolor{tfcbandblue} \tzerobeta & 46.7 & 78.8 & 46.2 & 84.2 \\
TiRex-2 & 45.5 & 77.7 & 44.8 & 81.1 \\
Toto-2.0-1B & 44.4 & 76.6 & 38.9 & 68.4 \\
Toto-2.0-2.5B & 44.4 & 77.5 & 38.8 & 69.0 \\
Toto-2.0-313m & 44.2 & 75.0 & 38.7 & 66.7 \\
TabPFN-TS-3 & 43.1 & 61.7 & 43.3 & 67.4 \\
TS-ICL & 43.1 & 62.3 & 41.9 & 65.6 \\
Toto-2.0-22m & 42.9 & 68.1 & 37.6 & 61.2 \\
TiRex & 42.6 & 69.8 & 38.7 & 68.4 \\
TimesFM-2.5 & 42.2 & 65.9 & 37.4 & 62.5 \\
\rowcolor{tfcband} \tzeroalpha & 42.2 & 59.6 & 39.7 & 62.8 \\
TabPFN-TS & 41.5 & 56.1 & 42.5 & 65.8 \\
CITRAS-FM & 41.2 & 59.3 & 39.0 & 63.9 \\
Toto-1.0 & 40.8 & 57.2 & 35.2 & 51.1 \\
Toto-2.0-4m & 40.7 & 56.5 & 35.4 & 50.7 \\
FlowState & 39.9 & 59.3 & 36.7 & 61.1 \\
Moirai-2.0 & 39.3 & 51.6 & 36.6 & 54.3 \\
Chronos-Bolt & 38.9 & 51.0 & 35.9 & 52.4 \\
Sundial & 33.4 & 33.5 & 28.0 & 34.3 \\
Seasonal naive & 0.0 & 12.9 & 0.0 & 12.8 \\
\bottomrule
\end{tabular}

\end{table}

\begin{figure}[!ht]
  \centering

\colorlet{tfccovwith}{tfcgreen}
\colorlet{tfccovwithout}{tfcproduct}
\begin{tikzpicture}
  \begin{axis}[
      name=covlift0,
      tfcaxis, ymajorgrids=false,
      scale only axis, width=9.5cm, height=8.37cm,
      xmin=-0.586, xmax=0.384, xtick={-0.5229,-0.3010,0.0000,0.3010}, xticklabels={0.3,0.5,1,2},
      ymin=-0.6, ymax=30.6, ytick={0,1,2,3,4,5,6,7,8,9,10,11,12,13,14,15,16,17,18,19,20,21,22,23,24,25,26,27,28,29,30},
      yticklabels={favorita\_stores\_1W, favorita\_stores\_1M, rohlik\_sales\_1W, rohlik\_orders\_1W, uci\_air\_quality\_1D, epf\_de, rohlik\_sales\_1D, epf\_np, rohlik\_orders\_1D, favorita\_transactions\_1D, m5\_1M, favorita\_stores\_1D, proenfo\_gfc17, m5\_1W, solar\_with\_weather\_1H, proenfo\_gfc12, uci\_air\_quality\_1H, walmart, solar\_with\_weather\_15T, hermes, m5\_1D, proenfo\_gfc14, entsoe\_15T, epf\_be, rossmann\_1D, entsoe\_30T, rossmann\_1W, entsoe\_1H, epf\_pjm, epf\_fr, \textbf{Geometric mean (30 tasks)}},
      every tick label/.append style={font=\tiny},
      y tick label style={color=tfcbody},
      title={(a) KNOWN-FUTURE COVARIATES},
      title style={font=\scriptsize\bfseries, color=tfcink, anchor=south, at={(0.5,1)}, yshift=1pt},
      every axis plot/.append style={line width=0.5pt},
      clip=false,
      legend to name=covliftlegend, legend columns=-1,
      legend entries={without covariates, with covariates},
    ]
    \addlegendimage{only marks, mark=o, mark size=1.4pt, tfccovwithout}
    \addlegendimage{only marks, mark=*, mark size=1.4pt, tfccovwith}
    \fill[tfcband] ({rel axis cs:0,0} |- {axis cs:0,29.5}) rectangle ({rel axis cs:1,0} |- {axis cs:0,30.5});
    \addplot[forget plot, only marks, mark=none,
              error bars/.cd, x dir=both, x explicit, error mark=none,
              error bar style={draw=tfchair, line width=0.8pt}]
      table[col sep=comma,x=log_with, y=y, x error plus=gap_plus, x error minus=gap_minus,
            restrict expr to domain={\thisrow{panel}}{0:0}]{figures/covariate-lift.csv};
    \addplot[forget plot, only marks, mark=o, mark size=1.4pt, tfccovwithout]
      table[col sep=comma,x=log_without, y=y, restrict expr to domain={\thisrow{panel}}{0:0}]{figures/covariate-lift.csv};
    \addplot[forget plot, only marks, mark=*, mark size=1.4pt, tfccovwith]
      table[col sep=comma,x=log_with, y=y, restrict expr to domain={\thisrow{panel}}{0:0}]{figures/covariate-lift.csv};
  \end{axis}
  \begin{axis}[
      name=covlift1,
      at={(covlift0.south west)}, anchor=north west, yshift=-1.2cm,
      tfcaxis, ymajorgrids=false,
      scale only axis, width=9.5cm, height=3.51cm,
      xmin=-0.422, xmax=1.208, xtick={-0.3010,0.0000,0.3010,0.4771,0.6990,1.0000}, xticklabels={0.5,1,2,3,5,10},
      ymin=-0.6, ymax=12.6, ytick={0,1,2,3,4,5,6,7,8,9,10,11,12},
      yticklabels={uk\_covid\_nation\_1D/cumulative, fred\_md\_2025/cee, uk\_covid\_nation\_1W/cumulative, uk\_covid\_nation\_1W/new, fred\_qd\_2025/cee, uk\_covid\_nation\_1D/new, favorita\_transactions\_1W, favorita\_transactions\_1M, kdd\_cup\_2022\_1D, gvar, kdd\_cup\_2022\_10T, kdd\_cup\_2022\_30T, \textbf{Geometric mean (12 tasks)}},
      every tick label/.append style={font=\tiny},
      y tick label style={color=tfcbody},
      title={(b) PAST COVARIATES ONLY},
      title style={font=\scriptsize\bfseries, color=tfcink, anchor=south, at={(0.5,1)}, yshift=1pt},
      every axis plot/.append style={line width=0.5pt},
      clip=false,
    ]
    \fill[tfcband] ({rel axis cs:0,0} |- {axis cs:0,11.5}) rectangle ({rel axis cs:1,0} |- {axis cs:0,12.5});
    \addplot[forget plot, only marks, mark=none,
              error bars/.cd, x dir=both, x explicit, error mark=none,
              error bar style={draw=tfchair, line width=0.8pt}]
      table[col sep=comma,x=log_with, y=y, x error plus=gap_plus, x error minus=gap_minus,
            restrict expr to domain={\thisrow{panel}}{1:1}]{figures/covariate-lift.csv};
    \addplot[forget plot, only marks, mark=o, mark size=1.4pt, tfccovwithout]
      table[col sep=comma,x=log_without, y=y, restrict expr to domain={\thisrow{panel}}{1:1}]{figures/covariate-lift.csv};
    \addplot[forget plot, only marks, mark=*, mark size=1.4pt, tfccovwith]
      table[col sep=comma,x=log_with, y=y, restrict expr to domain={\thisrow{panel}}{1:1}]{figures/covariate-lift.csv};
  \end{axis}
  \node[anchor=north, align=center, font=\scriptsize, color=tfcbody, yshift=-12pt]
    at (covlift1.south) {Scaled quantile loss, log axis\\[1pt]\textcolor{tfcmuted}{$\leftarrow$\ lower is better}};
  \node[anchor=north, yshift=-34pt] at (covlift1.south) {\ref{covliftlegend}};
\end{tikzpicture}
  \caption{Covariate lift per fev-bench task. Scaled quantile loss of the
    same checkpoint without covariates (hollow) and with them (green), on a
    log axis so that the length of the line joining them is the relative
    change. Within each covariate subset, tasks are sorted by their loss
    without covariates, the lowest at the top, so the hollow marks read as
    a difficulty ordering. The shaded row is the geometric mean over the
    tasks of the subset.}
  \label{fig:covariate-lift}
\end{figure}
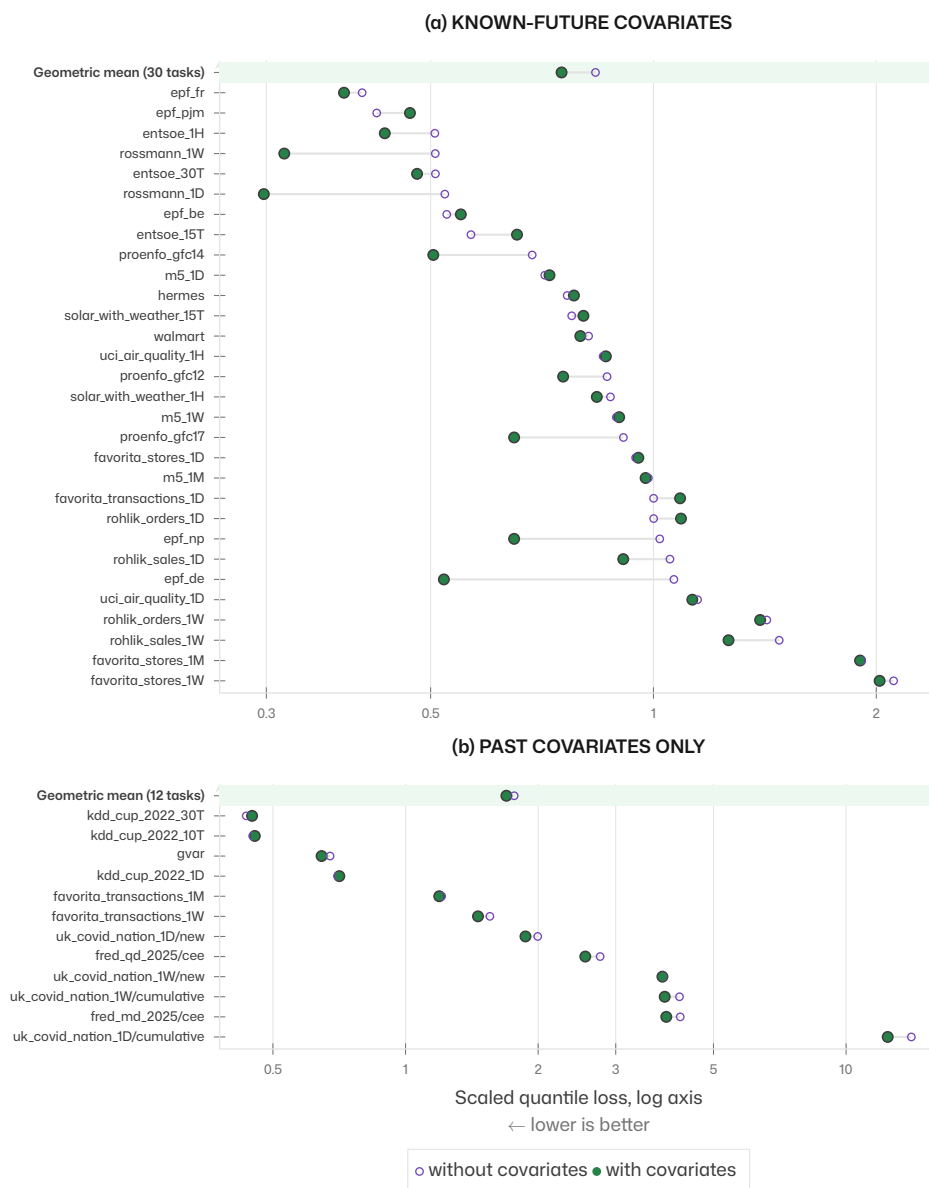

Table~\ref{tab:covariate-lift} shows that \tzeroalpha can translate operational context into substantial forecasting gains without retraining. Passing known-future covariates raises skill from 36.7 to 43.0 across 30 tasks, a gain of 6.3 percentage points, with improvements on 19 tasks. Passing
past covariates raises skill from 32.9 to 35.6, a gain of 2.7 points, with improvements on 9 tasks.
On German day-ahead
electricity prices, supplying the grid operator's day-ahead
forecasts of load and solar and wind generation reduces
quantile loss by 51\%. On daily Rossmann drugstore sales,
supplying planned promotions, holidays and opening days
reduces it by 43\%. Both comparisons use the same \tzeroalpha checkpoint with and without covariates.
The gains are not universal. Eleven of the 30
known-future tasks score worse with covariates, with the largest deterioration
reaching 15\% on a 15-minute electricity series with weather covariates.
On M5, eight
covariates covering product prices, calendar events, and food-stamp days change
quantile loss by less than 2\% in either direction at weekly and
monthly granularities (Figure~\ref{fig:covariate-lift}). The results demonstrate the value of conditioning on informative context while identifying selective use of noisy or uninformative covariates as a remaining challenge.

\begin{table}[!ht]
  \centering
  \caption{Covariate lift on fev-bench: skill of the same checkpoint with
    covariates dropped and with covariates passed, by covariate subset.
    Confidence intervals are 95\% bootstrap intervals over tasks.}
  \label{tab:covariate-lift}
  \begin{tabular}{lrrrrr}
    \toprule
    Subset & Tasks & Skill without & Skill with & Lift (95\% CI) & Tasks improved \\
    \midrule
    Known-future covariates & 30 & 36.7 & 43.0 & $+6.3$ $[+2.2, +10.9]$ & 19 of 30 \\
    Past covariates only & 12 & 32.9 & 35.6 & $+2.7$ $[+1.0, +4.2]$ & 9 of 12 \\
    \bottomrule
  \end{tabular}
\end{table}

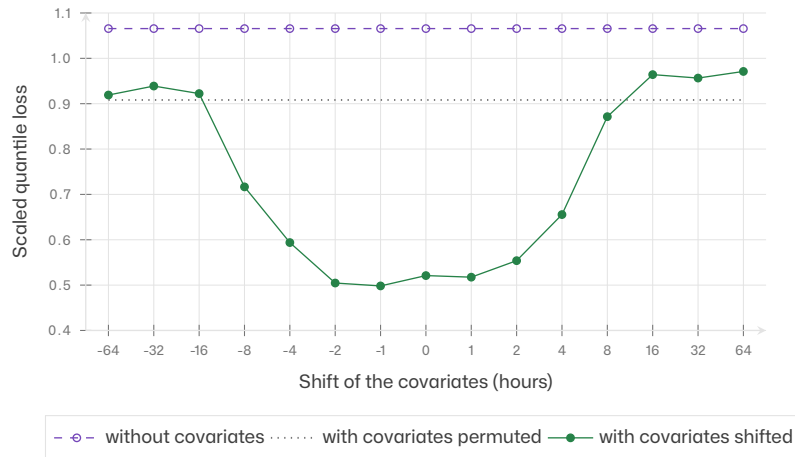
\begin{figure}[!ht]
  \centering
%
\colorlet{tfccovwith}{tfcgreen}
\colorlet{tfccovwithout}{tfcproduct}
\begin{tikzpicture}
  \begin{axis}[
      name=covsweep,
      tfcaxis,
      scale only axis, width=9.0cm, height=4.2cm,
      xmin=-0.5, xmax=14.5, xtick={0,1,2,3,4,5,6,7,8,9,10,11,12,13,14}, xticklabels={-64,-32,-16,-8,-4,-2,-1,0,1,2,4,8,16,32,64},
      ymin=0.4, ymax=1.1, ytick={0.4,0.5,0.6,0.7,0.8,0.9,1.0,1.1},
      yticklabel style={/pgf/number format/fixed, /pgf/number format/fixed zerofill, /pgf/number format/precision=1},
      every tick label/.append style={font=\tiny},
      xlabel={Shift of the covariates (hours)},
      ylabel={Scaled quantile loss},
      label style={font=\scriptsize, color=tfcbody},
      every axis plot/.append style={line width=0.5pt},
      legend to name=covsweeplegend, legend columns=-1,
      legend entries={without covariates, with covariates permuted, with covariates shifted},
      clip=false,
    ]
    \addplot[dashed, tfccovwithout, mark=o, mark size=1.4pt, mark options={solid}]
      table[col sep=comma,x=x, y=sql_without]{figures/covariate-sweep.csv};
    \addplot[dotted, tfcmuted, line width=0.7pt, mark=none]
      table[col sep=comma,x=x, y=sql_permuted]{figures/covariate-sweep.csv};
    \addplot[tfccovwith, mark=*, mark size=1.4pt]
      table[col sep=comma,x=x, y=sql_shifted]{figures/covariate-sweep.csv};
  \end{axis}
  \node[anchor=north, yshift=-2pt] at (covsweep.below south) {\ref{covsweeplegend}};
\end{tikzpicture}
  \caption{Alignment sweep on the German day-ahead electricity-price task:
    scaled quantile loss as the two forecast covariates, load and
    solar-plus-wind generation, are shifted in time by the lag on the
    x axis (green), randomly permuted in time (dotted), or dropped
    (hollow). Zero is the published alignment.}
  \label{fig:covariate-sweep}
\end{figure}

To measure further how sharply the
model reads covariate alignment, we take the German price task and degrade
its two forecast covariates, shifting them in time by up to $\pm 64$ hours,
permuting them, or dropping them (Figure~\ref{fig:covariate-sweep}).
Permutation is performed by reordering the hours of a covariate. We draw a random
permutation of the time index over the whole span the model reads, context
and horizon alike, and apply it to that covariate values, independently for
each of the two. The alignment with the price is destroyed, and so is the coupling
between load and generation. The price and its history are untouched.
Loss degrades smoothly with misalignment, roughly doubling from
perfect alignment to a 16-hour shift, and even permuted covariates beat
dropping them, since their marginal distribution still carries level
information. The model therefore keys on the point-by-point
alignment with the target, at single-step resolution, rather than on their
presence.

\FloatBarrier

\subsubsection{TIME}

TIME \citep{qiao2026time} keeps the scoring of GIFT-Eval but replaces its data. MASE
and CRPS are computed and normalized by seasonal naive as on GIFT-Eval, then
aggregated by geometric mean over 98 tasks. A task pairs one of 50 datasets,
each sampled at one frequency, with a horizon set per frequency and
application. The 50 datasets are new and independent from GIFT-Eval. They span
over the period between 2020 and 2025. There are no time series with covariates
in the benchmark, but 74 of the 98 tasks involve multiple targets.

On the 98 tasks \tzeroalpha attains a normalized MASE of 0.685 and a CRPS of
0.572 (Table~\ref{tab:timebench}). \tzerobeta attains a MASE of 0.665 and a
CRPS of 0.555, placing it 5th in Table~\ref{tab:timebench} by CRPS
and 6th by MASE, 3.5\% of CRPS short of the lead.

\begin{table}[!ht]
  \centering
  \caption{TIME aggregated CRPS and MASE, normalized by seasonal naive per
    task and combined by geometric mean over the 98 tasks (lower is better).
    Rows are the zero-shot models that Tables~\ref{tab:gifteval}
    and~\ref{tab:fevbench} also report, sorted by CRPS. The best value of
    each column is in bold. \tzeroalpha is shaded green and \tzerobeta, its
    successor, blue.}
  \label{tab:timebench}
\begin{tabular}{lrr}
\toprule
\tfcheadrow{Model} & \tfcheadrow{CRPS $\downarrow$} & \tfcheadrow{MASE $\downarrow$} \\
\midrule
TimesFM-3.0 & \textbf{0.5363} & \textbf{0.6398} \\
Toto-2.0-2.5B & 0.5394 & 0.6419 \\
Toto-2.0-313m & 0.5423 & 0.6444 \\
Toto-2.0-1B & 0.5446 & 0.6448 \\
\rowcolor{tfcbandblue} \tzerobeta & 0.5551 & 0.6647 \\
Chronos-2 & 0.5563 & 0.6620 \\
Toto-2.0-22m & 0.5616 & 0.6703 \\
TimesFM-2.5 & 0.5674 & 0.6686 \\
\rowcolor{tfcband} \tzeroalpha & 0.5721 & 0.6848 \\
TiRex & 0.5731 & 0.6831 \\
Toto-1.0 & 0.5839 & 0.6980 \\
Moirai-2.0 & 0.5885 & 0.7030 \\
Sundial & 0.6634 & 0.7575 \\
Seasonal naive & 1.0000 & 1.0000 \\
\bottomrule
\end{tabular}

\end{table}

TIME also scores models on groups of variates that share a temporal pattern, among the following:
trend strength, trend linearity, seasonal strength, seasonal correlation,
residual autocorrelation, spectral entropy and stationarity. Each continuous
feature is cut at its median over the benchmark variates, and the score of a model
on either side of the cut is the geometric mean of its normalized MASE
over those variates.

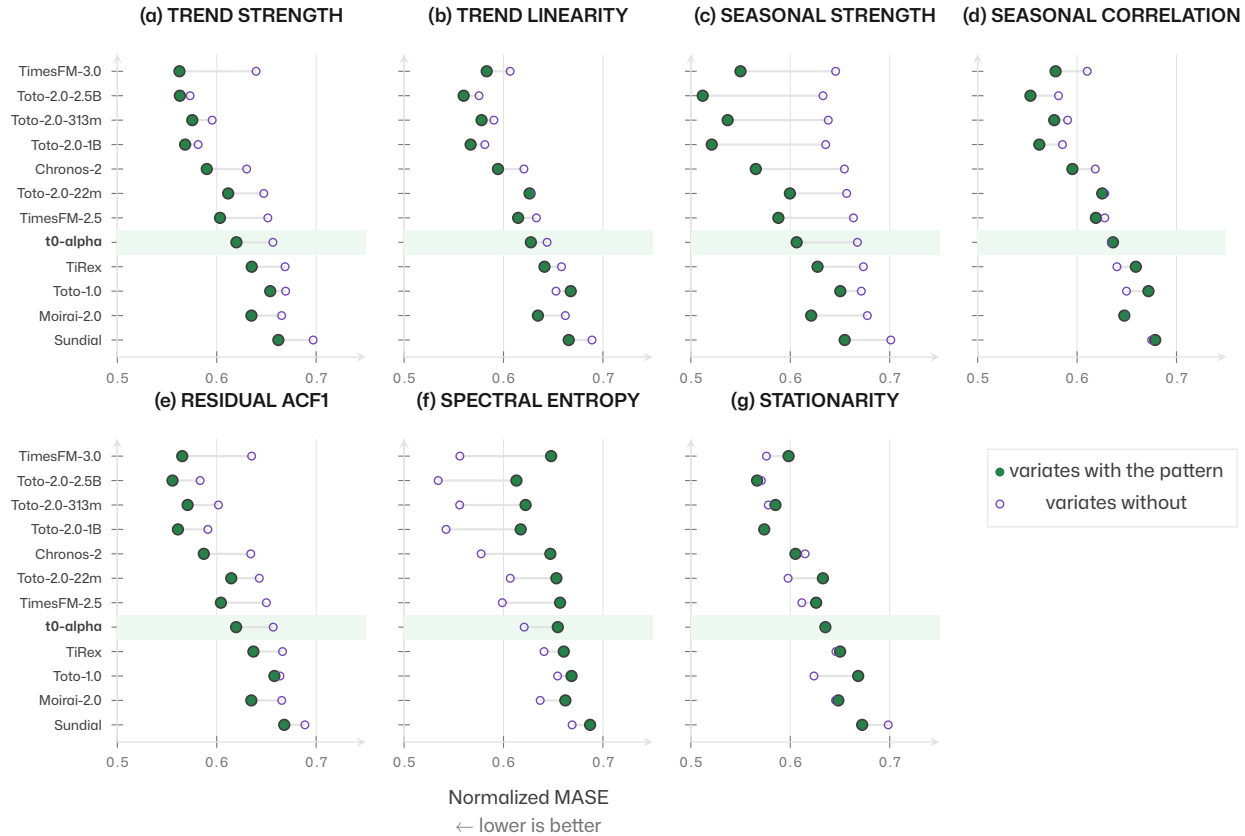
\begin{figure}[!ht]
  \centering
%
%
\colorlet{tfcpatternwith}{tfcgreen}
\colorlet{tfcpatternwithout}{tfcproduct}
\resizebox{\linewidth}{!}{%
\begin{tikzpicture}
  \begin{groupplot}[
      group style={group size=4 by 2, horizontal sep=14pt, vertical sep=30pt,
                   yticklabels at=edge left},
      tfcaxis, ymajorgrids=false,
      scale only axis, width=3.2cm, height=3.9cm,
      xmin=0.5, xmax=0.75, xtick={0.5,0.6,0.7},
      xticklabel style={/pgf/number format/fixed, /pgf/number format/fixed zerofill,
                        /pgf/number format/precision=1},
      ymin=-0.7, ymax=11.7, ytick={0,1,2,3,4,5,6,7,8,9,10,11},
      yticklabels={Sundial,Moirai-2.0,Toto-1.0,TiRex,\textbf{t0-alpha},TimesFM-2.5,Toto-2.0-22m,Chronos-2,Toto-2.0-1B,Toto-2.0-313m,Toto-2.0-2.5B,TimesFM-3.0},
      every tick label/.append style={font=\tiny},
      y tick label style={color=tfcbody},
      title style={font=\scriptsize\bfseries, color=tfcink, anchor=south, at={(0.5,1)}, yshift=1pt},
      every axis plot/.append style={line width=0.5pt},
      clip=false,
    ]
    \nextgroupplot[title={(a) TREND STRENGTH}, legend to name=timepatternslegend, legend columns=1, legend entries={variates with the pattern, variates without}]
      \addlegendimage{only marks, mark=*, mark size=1.4pt, tfcpatternwith}
      \addlegendimage{only marks, mark=o, mark size=1.4pt, tfcpatternwithout}
      \fill[tfcband] ({rel axis cs:0,0} |- {axis cs:0,3.5}) rectangle ({rel axis cs:1,0} |- {axis cs:0,4.5});
      \addplot[forget plot, only marks, mark=none,
                error bars/.cd, x dir=both, x explicit, error mark=none,
                error bar style={draw=tfchair, line width=0.8pt}]
        table[col sep=comma,x=with_trend_strength, y=y, x error plus=gap_plus_trend_strength, x error minus=gap_minus_trend_strength]{figures/timebench-patterns.csv};
      \addplot[forget plot, only marks, mark=o, mark size=1.4pt, tfcpatternwithout]
        table[col sep=comma,x=without_trend_strength, y=y]{figures/timebench-patterns.csv};
      \addplot[forget plot, only marks, mark=*, mark size=1.4pt, tfcpatternwith]
        table[col sep=comma,x=with_trend_strength, y=y]{figures/timebench-patterns.csv};
    \nextgroupplot[title={(b) TREND LINEARITY}]
      \fill[tfcband] ({rel axis cs:0,0} |- {axis cs:0,3.5}) rectangle ({rel axis cs:1,0} |- {axis cs:0,4.5});
      \addplot[forget plot, only marks, mark=none,
                error bars/.cd, x dir=both, x explicit, error mark=none,
                error bar style={draw=tfchair, line width=0.8pt}]
        table[col sep=comma,x=with_linearity, y=y, x error plus=gap_plus_linearity, x error minus=gap_minus_linearity]{figures/timebench-patterns.csv};
      \addplot[forget plot, only marks, mark=o, mark size=1.4pt, tfcpatternwithout]
        table[col sep=comma,x=without_linearity, y=y]{figures/timebench-patterns.csv};
      \addplot[forget plot, only marks, mark=*, mark size=1.4pt, tfcpatternwith]
        table[col sep=comma,x=with_linearity, y=y]{figures/timebench-patterns.csv};
    \nextgroupplot[title={(c) SEASONAL STRENGTH}]
      \fill[tfcband] ({rel axis cs:0,0} |- {axis cs:0,3.5}) rectangle ({rel axis cs:1,0} |- {axis cs:0,4.5});
      \addplot[forget plot, only marks, mark=none,
                error bars/.cd, x dir=both, x explicit, error mark=none,
                error bar style={draw=tfchair, line width=0.8pt}]
        table[col sep=comma,x=with_seasonal_strength, y=y, x error plus=gap_plus_seasonal_strength, x error minus=gap_minus_seasonal_strength]{figures/timebench-patterns.csv};
      \addplot[forget plot, only marks, mark=o, mark size=1.4pt, tfcpatternwithout]
        table[col sep=comma,x=without_seasonal_strength, y=y]{figures/timebench-patterns.csv};
      \addplot[forget plot, only marks, mark=*, mark size=1.4pt, tfcpatternwith]
        table[col sep=comma,x=with_seasonal_strength, y=y]{figures/timebench-patterns.csv};
    \nextgroupplot[title={(d) SEASONAL CORRELATION}]
      \fill[tfcband] ({rel axis cs:0,0} |- {axis cs:0,3.5}) rectangle ({rel axis cs:1,0} |- {axis cs:0,4.5});
      \addplot[forget plot, only marks, mark=none,
                error bars/.cd, x dir=both, x explicit, error mark=none,
                error bar style={draw=tfchair, line width=0.8pt}]
        table[col sep=comma,x=with_seasonal_corr, y=y, x error plus=gap_plus_seasonal_corr, x error minus=gap_minus_seasonal_corr]{figures/timebench-patterns.csv};
      \addplot[forget plot, only marks, mark=o, mark size=1.4pt, tfcpatternwithout]
        table[col sep=comma,x=without_seasonal_corr, y=y]{figures/timebench-patterns.csv};
      \addplot[forget plot, only marks, mark=*, mark size=1.4pt, tfcpatternwith]
        table[col sep=comma,x=with_seasonal_corr, y=y]{figures/timebench-patterns.csv};
    \nextgroupplot[title={(e) RESIDUAL ACF1}]
      \fill[tfcband] ({rel axis cs:0,0} |- {axis cs:0,3.5}) rectangle ({rel axis cs:1,0} |- {axis cs:0,4.5});
      \addplot[forget plot, only marks, mark=none,
                error bars/.cd, x dir=both, x explicit, error mark=none,
                error bar style={draw=tfchair, line width=0.8pt}]
        table[col sep=comma,x=with_e_acf1, y=y, x error plus=gap_plus_e_acf1, x error minus=gap_minus_e_acf1]{figures/timebench-patterns.csv};
      \addplot[forget plot, only marks, mark=o, mark size=1.4pt, tfcpatternwithout]
        table[col sep=comma,x=without_e_acf1, y=y]{figures/timebench-patterns.csv};
      \addplot[forget plot, only marks, mark=*, mark size=1.4pt, tfcpatternwith]
        table[col sep=comma,x=with_e_acf1, y=y]{figures/timebench-patterns.csv};
    \nextgroupplot[title={(f) SPECTRAL ENTROPY}]
      \fill[tfcband] ({rel axis cs:0,0} |- {axis cs:0,3.5}) rectangle ({rel axis cs:1,0} |- {axis cs:0,4.5});
      \addplot[forget plot, only marks, mark=none,
                error bars/.cd, x dir=both, x explicit, error mark=none,
                error bar style={draw=tfchair, line width=0.8pt}]
        table[col sep=comma,x=with_x_entropy, y=y, x error plus=gap_plus_x_entropy, x error minus=gap_minus_x_entropy]{figures/timebench-patterns.csv};
      \addplot[forget plot, only marks, mark=o, mark size=1.4pt, tfcpatternwithout]
        table[col sep=comma,x=without_x_entropy, y=y]{figures/timebench-patterns.csv};
      \addplot[forget plot, only marks, mark=*, mark size=1.4pt, tfcpatternwith]
        table[col sep=comma,x=with_x_entropy, y=y]{figures/timebench-patterns.csv};
    \nextgroupplot[title={(g) STATIONARITY}]
      \fill[tfcband] ({rel axis cs:0,0} |- {axis cs:0,3.5}) rectangle ({rel axis cs:1,0} |- {axis cs:0,4.5});
      \addplot[forget plot, only marks, mark=none,
                error bars/.cd, x dir=both, x explicit, error mark=none,
                error bar style={draw=tfchair, line width=0.8pt}]
        table[col sep=comma,x=with_stationarity, y=y, x error plus=gap_plus_stationarity, x error minus=gap_minus_stationarity]{figures/timebench-patterns.csv};
      \addplot[forget plot, only marks, mark=o, mark size=1.4pt, tfcpatternwithout]
        table[col sep=comma,x=without_stationarity, y=y]{figures/timebench-patterns.csv};
      \addplot[forget plot, only marks, mark=*, mark size=1.4pt, tfcpatternwith]
        table[col sep=comma,x=with_stationarity, y=y]{figures/timebench-patterns.csv};
    \nextgroupplot[group/empty plot]
  \end{groupplot}
  \node[anchor=north, align=center, font=\scriptsize, color=tfcbody, yshift=-14pt]
    at ($(group c1r2.south west)!0.5!(group c3r2.south east)$)
    {Normalized MASE\\[1pt]\textcolor{tfcmuted}{$\leftarrow$\ lower is better}};
  \node[anchor=north west] at (group c4r2.north west) {\ref{timepatternslegend}};
\end{tikzpicture}%
}
  \caption{TIME pattern-level results for the models of
    Table~\ref{tab:timebench}, in the row order of that table: their overall
    normalized CRPS on the full benchmark, most accurate at the top.
    The x axis, shared by every panel, is the MASE normalized by seasonal
    naive on the variates that carry the pattern (filled) and on those that do
    not (hollow). Lower is better. \textbf{(a)}--\textbf{(f)} cut a continuous
    feature at its median over the variates of the benchmark; \textbf{(g)} is the
    stationarity flag of an augmented Dickey-Fuller (ADF) test.}
  \label{fig:timebench-patterns}
\end{figure}

Like every model in Figure~\ref{fig:timebench-patterns}, \tzeroalpha does
better on strongly trended and strongly seasonal variates, and stationarity
does not move its score.

\subsection{Model capability analysis}
\label{sec:capabilities}

\subsubsection{Calibration}
\label{sec:calibration}

The public benchmarks above score accuracy. This section tests whether the model
quantiles are calibrated. A quantile forecast at level $q$ is calibrated
if the realized value falls below it a fraction $q$ of the time
\citep{gneiting2007probabilistic}. Calibration is what lets a forecast be
used as a probability across decisions. For instance, a buyer who covers the 0.9 quantile
of demand expects to run short one time in ten, whatever the series. Calibration is
not sufficient on its own. A model that ignores the context and emits the
unconditional distribution of the series is calibrated in this sense but
carries no informative forecast on the horizon. Eventually calibration has to be read alongside
accuracy, the sharpness of the distribution
\citep{gneiting2007probabilistic}. We measure calibration through coverage,
the empirical fraction of outcomes that fall below a quantile or inside the
band between two quantiles. A band can reach its nominal coverage while both
its quantiles are off in the same direction, so we report coverage per level
as well as per band.

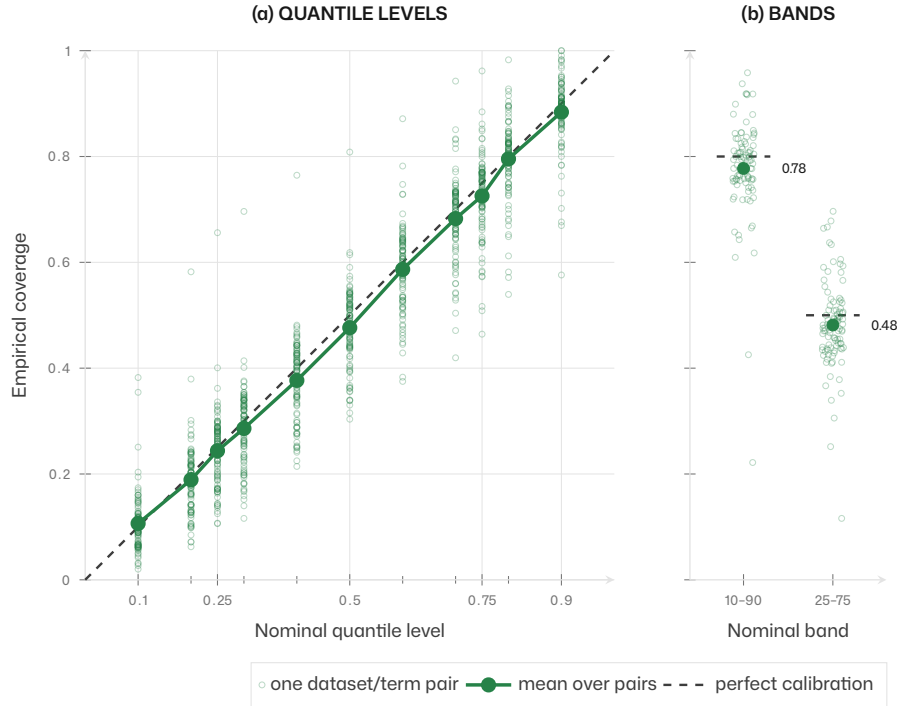
\begin{figure}[tb]
  \centering
%
\colorlet{tfccovpair}{tfcgreen}
\colorlet{tfccovmean}{tfcgreen}
\begin{tikzpicture}
  \begin{axis}[
      name=covlevels,
      tfcaxis, grid=none,
      scale only axis, height=7.0cm,
      ymin=0, ymax=1, ytick={0,0.2,0.4,0.6,0.8,1},
      yticklabel style={/pgf/number format/fixed, /pgf/number format/precision=1},
      every tick label/.append style={font=\tiny},
      label style={font=\scriptsize, color=tfcbody},
      title style={font=\scriptsize\bfseries, color=tfcink, anchor=south, at={(0.5,1)}, yshift=1pt},
      clip=false,
      width=7.0cm, grid=major,
      xmin=0, xmax=1,
      xtick={0.1,0.25,0.5,0.75,0.9},
      minor xtick={0.2,0.3,0.4,0.6,0.7,0.8},
      xticklabel style={/pgf/number format/fixed, /pgf/number format/precision=2},
      xlabel={Nominal quantile level},
      ylabel={Empirical coverage},
      title={(a) QUANTILE LEVELS},
      legend to name=coveragelegend, legend columns=-1,
      legend entries={one dataset/term pair, mean over pairs, perfect calibration},
    ]
    \addlegendimage{only marks, mark=o, mark size=1.1pt, tfccovpair, opacity=0.3}
    \addlegendimage{mark=*, mark size=2.2pt, tfccovmean, line width=1.3pt}
    \addlegendimage{tfcbody, dashed, line width=0.9pt}
    \addplot[forget plot, tfcbody, dashed, line width=0.9pt] coordinates {(0,0) (1,1)};
    \addplot[forget plot, only marks, mark=o, mark size=1.1pt, tfccovpair, opacity=0.3]
      table[col sep=comma,x=x, y=y, restrict expr to domain={\thisrow{kind_id}}{0:0}]{figures/calibration-coverage.csv};
    \addplot[forget plot, tfccovmean, mark=*, mark size=2.2pt, line width=1.3pt]
      table[col sep=comma,x=x, y=y, restrict expr to domain={\thisrow{kind_id}}{1:1}]{figures/calibration-coverage.csv};
  \end{axis}
  \begin{axis}[
      name=covbands,
      at={(covlevels.south east)}, anchor=south west, xshift=1.0cm,
      tfcaxis, grid=none,
      scale only axis, height=7.0cm,
      ymin=0, ymax=1, ytick={0,0.2,0.4,0.6,0.8,1},
      yticklabel style={/pgf/number format/fixed, /pgf/number format/precision=1},
      every tick label/.append style={font=\tiny},
      label style={font=\scriptsize, color=tfcbody},
      title style={font=\scriptsize\bfseries, color=tfcink, anchor=south, at={(0.5,1)}, yshift=1pt},
      clip=false,
      width=2.6cm,
      xmin=-0.6, xmax=1.6, xtick={0,1}, xticklabels={10--90, 25--75},
      yticklabel=\empty,
      xlabel={Nominal band},
      title={(b) BANDS},
    ]
    \addplot[forget plot, tfcbody, dashed, line width=0.9pt] coordinates {(-0.3,0.8) (0.3,0.8)};
    \addplot[forget plot, tfcbody, dashed, line width=0.9pt] coordinates {(0.7,0.5) (1.3,0.5)};
    \addplot[forget plot, only marks, mark=o, mark size=1.1pt, tfccovpair, opacity=0.3]
      table[col sep=comma,x=x, y=y, restrict expr to domain={\thisrow{kind_id}}{2:2}]{figures/calibration-coverage.csv};
    \addplot[forget plot, only marks, mark=*, mark size=2.2pt, tfccovmean]
      table[col sep=comma,x=x, y=y, restrict expr to domain={\thisrow{kind_id}}{3:3}]{figures/calibration-coverage.csv};
    \node[anchor=west, font=\tiny, color=tfcink] at (axis cs:0.32,0.7774) {0.78};
    \node[anchor=west, font=\tiny, color=tfcink] at (axis cs:1.32,0.4815) {0.48};
  \end{axis}
  \node[anchor=north, yshift=-2pt] at ($(covlevels.below south)!0.5!(covbands.below south)$) {\ref{coveragelegend}};
\end{tikzpicture}
  \caption{Calibration of \tzeroalpha on GIFT-Eval. Nominal against empirical coverage (a) at
    the eleven quantile levels and (b) for the two nominal bands. Faint
    hollow marks are single dataset/term pairs, filled marks the mean over pairs,
    dashed lines perfect calibration. The labeled levels 0.1, 0.25, 0.5,
    0.75 and 0.9 are native. The six between them are interpolated.}
  \label{fig:calibration}
\end{figure}

To assess calibration, we generate forecasts over all
97 dataset/term tasks without using any covariates. For every
forecast step we record whether the observation falls below each of eleven
quantile levels: 0.1, 0.2, 0.25, 0.3, 0.4, 0.5, 0.6, 0.7, 0.75, 0.8 and 0.9. It corresponds to
the five native levels the model produces plus the six the package interpolates between them. We also check
whether it falls inside the nominal 10--90 and 25--75 bands. We
aggregate the results in two ways. First, we pool over all forecast steps, which weights a pair by
the number of steps it contributes. Second, we aggregate as an unweighted mean over datasets,
which gives a small dataset the same weight as a large one.
Figure~\ref{fig:calibration} shows the coverage of every pair and its mean.

Empirical coverage sits below nominal for both bands
(Figure~\ref{fig:calibration}). Pooled over all forecast steps, the 10--90
band covers 0.784 of outcomes against a nominal 0.80 and the 25--75 band
0.478 against 0.50. The intervals are therefore too narrow by 0.016 and
0.022. Averaged over datasets they cover 0.777 and 0.482. The pooled median
is exact, outcomes fall below it 0.500 of the time. Per level, pooled
coverage stays within 0.014 of nominal at all eleven levels. The six
interpolated levels are no worse than the native ones. Their largest pooled
deviation is 0.011, the largest native one 0.013, so interpolation does not
degrade coverage. Averaged over datasets, every level from 0.2 upwards sits
0.004 to 0.025 below nominal, while pooling brings them back to nominal.
The datasets with few forecast steps, which pooling down-weights, are the
ones whose quantiles run low.

\subsubsection{Impact of quantile extrapolation on calibration}
\label{sec:tails}

The model emits five quantile levels, and the released package turns them
into any level a user requests. A requested level that matches a native one
is returned as is. A level between two native ones is linearly interpolated
between them. A level outside $[0.1, 0.9]$ is clamped to the nearest native
level, so every tail request inherits the coverage of the 0.1 or the 0.9
quantile. Pooled over GIFT-Eval, a requested 0.01 quantile covers 0.109 of
outcomes and a requested 0.99 quantile 0.893
(Table~\ref{tab:tail-calibration}), an order of magnitude from nominal on
the left tail.

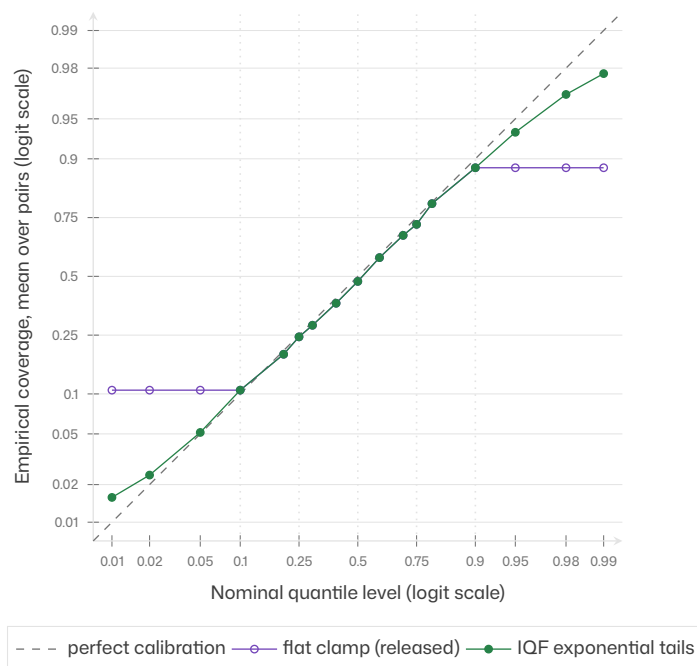
\begin{figure}[tb]
  \centering

\begin{tikzpicture}
  \begin{axis}[
      name=tailcal,
      tfcaxis, xmajorgrids=false,
      scale only axis, width=7.0cm, height=7.0cm,
      xmin=-4.945, xmax=4.945, ymin=-4.945, ymax=4.945,
      xtick={-4.5951,-3.8918,-2.9444,-2.1972,-1.0986,0.0000,1.0986,2.1972,2.9444,3.8918,4.5951}, xticklabels={0.01,0.02,0.05,0.1,0.25,0.5,0.75,0.9,0.95,0.98,0.99},
      ytick={-4.5951,-3.8918,-2.9444,-2.1972,-1.0986,0.0000,1.0986,2.1972,2.9444,3.8918,4.5951}, yticklabels={0.01,0.02,0.05,0.1,0.25,0.5,0.75,0.9,0.95,0.98,0.99},
      every tick label/.append style={font=\tiny},
      xlabel={Nominal quantile level (logit scale)},
      ylabel={Empirical coverage, mean over pairs (logit scale)},
      label style={font=\scriptsize, color=tfcbody},
      every axis plot/.append style={line width=0.5pt},
      legend to name=tailcallegend, legend columns=3,
      clip=false,
    ]
    \draw[tfchair, dotted, line width=0.6pt] (axis cs:-2.1972,-4.945) -- (axis cs:-2.1972,4.945);
    \draw[tfchair, dotted, line width=0.6pt] (axis cs:-1.0986,-4.945) -- (axis cs:-1.0986,4.945);
    \draw[tfchair, dotted, line width=0.6pt] (axis cs:0.0000,-4.945) -- (axis cs:0.0000,4.945);
    \draw[tfchair, dotted, line width=0.6pt] (axis cs:1.0986,-4.945) -- (axis cs:1.0986,4.945);
    \draw[tfchair, dotted, line width=0.6pt] (axis cs:2.1972,-4.945) -- (axis cs:2.1972,4.945);
    \addplot[tfcmuted, dashed, line width=0.5pt] coordinates {(-4.945,-4.945) (4.945,4.945)};
    \addlegendentry{perfect calibration}
    \addplot[tfcproduct, mark=o, mark size=1.4pt] table[col sep=comma,x=logit_level, y=package_mean_logit]{figures/tail-calibration.csv};
    \addlegendentry{flat clamp (released)}
    \addplot[tfcgreen, mark=*, mark size=1.4pt] table[col sep=comma,x=logit_level, y=iqf_mean_logit]{figures/tail-calibration.csv};
    \addlegendentry{IQF exponential tails}
  \end{axis}
  \node[anchor=north, yshift=-2pt] at (tailcal.below south) {\ref{tailcallegend}};
\end{tikzpicture}
  \caption{Reliability of \tzeroalpha on GIFT-Eval. Nominal
    against empirical coverage, averaged over pairs, at the 17 query levels
    from 0.01 to 0.99 under the released flat clamp (hollow) and the
    post-hoc IQF tails (filled). Both axes are on a logit scale. The dashed diagonal is perfect calibration and the
    dotted verticals mark the five native levels.}
  \label{fig:tail-calibration}
\end{figure}

We test an alternative that needs no training: the exponential tails of the
incremental quantile function (IQF) of \citet{park2022iqf}, applied post hoc
to the five native levels. Interior levels keep the released interpolation
bitwise, so every other number in this report is unaffected, and each tail
is pinned through the two outermost native levels on its side, where the
original method learns a tail parameter. Figure~\ref{fig:tail-calibration}
compares the two strategies over the 97 pairs. With IQF tails the 0.05 and
0.95 levels become close to calibrated, the coverage error at 0.01 drops
sixteen-fold, and no extrapolated quantile crossed a native level across
the benchmark. It improves accuracy. Normalized quantile loss at the extreme
levels roughly halves, 0.064 against 0.110 at 0.01 and 0.088 against 0.164
at 0.99. Residual gaps remain, 0.016 coverage at the 0.01 level and 0.984
at 0.99, and the native levels are untouched by construction. Pinning the
tails through two fixed levels is the simplest form of the method, and a
learned tail parameter would likely close most of the remaining gap.

\begin{table}[htb]
  \centering
  \caption{Empirical coverage of requested tail levels on GIFT-Eval
    (97 pairs, pooled over forecast points), under the released
    flat-clamp extrapolation and post-hoc IQF exponential tails. All
    interior levels are identical between mechanisms.}
  \label{tab:tail-calibration}
  \begin{tabular}{lrrrrrr}
    \toprule
    Nominal level & 0.01 & 0.02 & 0.05 & 0.95 & 0.98 & 0.99 \\
    \midrule
    Flat clamp (released) & 0.109 & 0.109 & 0.109 & 0.893 & 0.893 & 0.893 \\
    IQF tails & 0.016 & 0.025 & 0.054 & 0.948 & 0.975 & 0.984 \\
    \bottomrule
  \end{tabular}
\end{table}

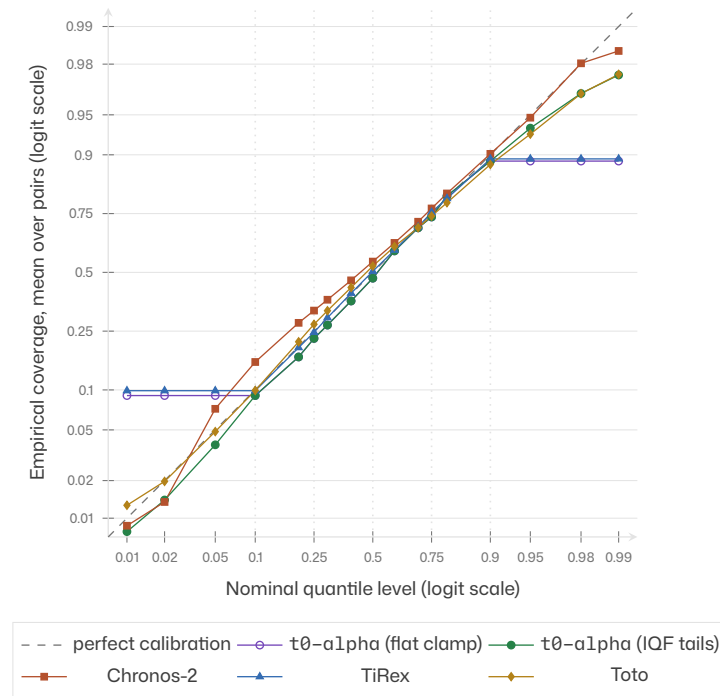
\begin{figure}[tb]
  \centering

\begin{tikzpicture}
  \begin{axis}[
      name=tailsota,
      tfcaxis, xmajorgrids=false,
      scale only axis, width=7.0cm, height=7.0cm,
      xmin=-4.945, xmax=4.945, ymin=-4.945, ymax=4.945,
      xtick={-4.5951,-3.8918,-2.9444,-2.1972,-1.0986,0.0000,1.0986,2.1972,2.9444,3.8918,4.5951}, xticklabels={0.01,0.02,0.05,0.1,0.25,0.5,0.75,0.9,0.95,0.98,0.99},
      ytick={-4.5951,-3.8918,-2.9444,-2.1972,-1.0986,0.0000,1.0986,2.1972,2.9444,3.8918,4.5951}, yticklabels={0.01,0.02,0.05,0.1,0.25,0.5,0.75,0.9,0.95,0.98,0.99},
      every tick label/.append style={font=\tiny},
      xlabel={Nominal quantile level (logit scale)},
      ylabel={Empirical coverage, mean over pairs (logit scale)},
      label style={font=\scriptsize, color=tfcbody},
      every axis plot/.append style={line width=0.5pt},
      legend to name=tailsotalegend, legend columns=3,
      clip=false,
    ]
    \draw[tfchair, dotted, line width=0.6pt] (axis cs:-2.1972,-4.945) -- (axis cs:-2.1972,4.945);
    \draw[tfchair, dotted, line width=0.6pt] (axis cs:-1.0986,-4.945) -- (axis cs:-1.0986,4.945);
    \draw[tfchair, dotted, line width=0.6pt] (axis cs:0.0000,-4.945) -- (axis cs:0.0000,4.945);
    \draw[tfchair, dotted, line width=0.6pt] (axis cs:1.0986,-4.945) -- (axis cs:1.0986,4.945);
    \draw[tfchair, dotted, line width=0.6pt] (axis cs:2.1972,-4.945) -- (axis cs:2.1972,4.945);
    \addplot[tfcmuted, dashed, line width=0.5pt] coordinates {(-4.945,-4.945) (4.945,4.945)};
    \addlegendentry{perfect calibration}
    \addplot[tfcproduct, mark=o, mark size=1.4pt] table[col sep=comma,x=logit_level, y=t0_clamp_mean_logit]{figures/tail-calibration-sota.csv};
    \addlegendentry{\tzeroalpha{} (flat clamp)}
    \addplot[tfcgreen, mark=*, mark size=1.4pt] table[col sep=comma,x=logit_level, y=t0_iqf_mean_logit]{figures/tail-calibration-sota.csv};
    \addlegendentry{\tzeroalpha{} (IQF tails)}
    \addplot[tfcresearch, mark=square*, mark size=1.2pt] table[col sep=comma,x=logit_level, y=chronos2_mean_logit]{figures/tail-calibration-sota.csv};
    \addlegendentry{Chronos-2}
    \addplot[tfcengineering, mark=triangle*, mark size=1.5pt] table[col sep=comma,x=logit_level, y=tirex_mean_logit]{figures/tail-calibration-sota.csv};
    \addlegendentry{TiRex}
    \addplot[tfcevent, mark=diamond*, mark size=1.5pt] table[col sep=comma,x=logit_level, y=toto_mean_logit]{figures/tail-calibration-sota.csv};
    \addlegendentry{Toto}
  \end{axis}
  \node[anchor=north, yshift=-2pt] at (tailsota.below south) {\ref{tailsotalegend}};
\end{tikzpicture}
  \caption{Reliability across public checkpoints on the 12-pair subset at
    the same 17 query levels, mean over pairs, on logit axes. The dashed
    diagonal is perfect calibration and the dotted verticals mark
    the native levels of \tzeroalpha. TiRex and the released \tzeroalpha
    package share the flat clamp and sit on horizontal segments beyond
    0.1 and 0.9; Chronos-2, Toto and \tzeroalpha with IQF tails follow
    the diagonal.}
  \label{fig:tail-sota}
\end{figure}

How a model answers a tail request is an architectural choice, and it
splits the field. Chronos-2 trains its head on 21 levels including 0.01
and 0.99, so extreme levels are learned outputs \citep{ansari2025chronos2}.
Toto draws 256 samples from a parametric output distribution and reads
empirical quantiles off them \citep{cohen2024toto}. TiRex emits the nine
deciles and its package clamps beyond $[0.1, 0.9]$ \citep{auer2025tirex},
as ours does. We run the three public checkpoints through the coverage
protocol on the 12-pair subset (Table~\ref{tab:gifteval-subset}), each
reproducing its published CRPS, and
restrict \tzeroalpha to the same pairs (Table~\ref{tab:tail-sota},
Figure~\ref{fig:tail-sota}), so these numbers differ slightly from the
97-pair ones above. The models that can move past their native levels,
Chronos-2, Toto and \tzeroalpha with IQF tails, follow the diagonal, while
TiRex and the released package sit on flat segments at 0.099 and 0.893. On
loss, \tzeroalpha with IQF tails is the most accurate at the 0.01 level,
0.048 against 0.071 to 0.072 for the three competitors, and ties Toto at
0.99, 0.114 against 0.112. Toto buys its tails with roughly two hours of
sampling on this subset, where the IQF tails are a closed form over five
numbers the head already emits. The interior tells the reverse story.
The 10--90 band of \tzeroalpha is the closest to its nominal 0.80 of the four
models, 0.798 against 0.794 for TiRex and 0.783 for Toto, while the intervals
of Chronos-2 run narrow, 0.744, and its 0.1 level over-covers at 0.158:
well-placed extremes on a less calibrated interior.

\begin{table}[htb]
  \centering
  \caption{Tail coverage and interior calibration across public
    checkpoints on the 12-pair subset (mean over pairs). Tail columns
    give empirical coverage at the requested level; band columns give
    exact interval coverage against nominal 0.80 and 0.50.}
  \label{tab:tail-sota}
  \begin{tabular}{lrrrrrr}
    \toprule
    Model & 0.01 & 0.05 & 0.95 & 0.99 & 10--90 & 25--75 \\
    \midrule
    \rowcolor{tfcband} \tzeroalpha (clamp) & 0.091 & 0.091 & 0.889 & 0.889 & 0.798 & 0.513 \\
    \rowcolor{tfcband} \tzeroalpha (IQF)   & 0.008 & 0.038 & 0.937 & 0.976 & 0.798 & 0.513 \\
    Chronos-2           & 0.009 & 0.072 & 0.947 & 0.984 & 0.744 & 0.439 \\
    TiRex               & 0.099 & 0.099 & 0.893 & 0.893 & 0.794 & 0.509 \\
    Toto                & 0.013 & 0.048 & 0.930 & 0.976 & 0.783 & 0.467 \\
    \bottomrule
  \end{tabular}
\end{table}

\subsection{Robustness to missing data}
\label{sec:missing-data}

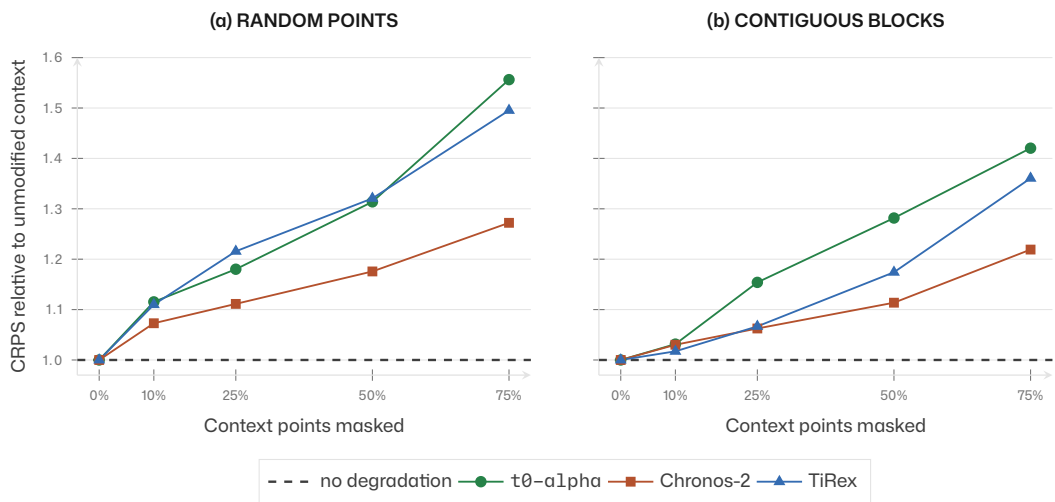
\begin{figure}[tbp]
  \centering
%
\begin{tikzpicture}
  \begin{axis}[
      name=missing0,
      tfcaxis, xmajorgrids=false,
      scale only axis, width=6.0cm, height=4.2cm,
      xmin=-4, xmax=79, xtick={0,10,25,50,75}, xticklabels={0\%,10\%,25\%,50\%,75\%},
      ymin=0.97, ymax=1.6, ytick={1.0,1.1,1.2,1.3,1.4,1.5,1.6},
      yticklabel style={/pgf/number format/fixed, /pgf/number format/fixed zerofill, /pgf/number format/precision=1},
      every tick label/.append style={font=\tiny},
      label style={font=\scriptsize, color=tfcbody},
      title style={font=\scriptsize\bfseries, color=tfcink, anchor=south, at={(0.5,1)}, yshift=1pt},
      every axis plot/.append style={line width=0.7pt},
      clip=false,
      title={(a) RANDOM POINTS},
      xlabel={Context points masked},
      ylabel={CRPS relative to unmodified context},
      legend to name=missingdatalegend, legend columns=-1,
    ]
    \addplot[tfcbody, dashed, line width=0.9pt] coordinates {(-4,1) (79,1)};
    \addlegendentry{no degradation}
    \addplot[tfcgreen, mark=*, mark size=1.8pt] table[col sep=comma,x=rate_pct, y=rel_crps,
      restrict expr to domain={\thisrow{series_id}}{0:0}]{figures/missing-data.csv};
    \addlegendentry{\tzeroalpha}
    \addplot[tfcresearch, mark=square*, mark size=1.5pt] table[col sep=comma,x=rate_pct, y=rel_crps,
      restrict expr to domain={\thisrow{series_id}}{10:10}]{figures/missing-data.csv};
    \addlegendentry{Chronos-2}
    \addplot[tfcengineering, mark=triangle*, mark size=1.9pt] table[col sep=comma,x=rate_pct, y=rel_crps,
      restrict expr to domain={\thisrow{series_id}}{20:20}]{figures/missing-data.csv};
    \addlegendentry{TiRex}
  \end{axis}
  \begin{axis}[
      name=missing1,
      at={(missing0.south east)}, anchor=south west, xshift=0.9cm,
      tfcaxis, xmajorgrids=false,
      scale only axis, width=6.0cm, height=4.2cm,
      xmin=-4, xmax=79, xtick={0,10,25,50,75}, xticklabels={0\%,10\%,25\%,50\%,75\%},
      ymin=0.97, ymax=1.6, ytick={1.0,1.1,1.2,1.3,1.4,1.5,1.6},
      yticklabel style={/pgf/number format/fixed, /pgf/number format/fixed zerofill, /pgf/number format/precision=1},
      every tick label/.append style={font=\tiny},
      label style={font=\scriptsize, color=tfcbody},
      title style={font=\scriptsize\bfseries, color=tfcink, anchor=south, at={(0.5,1)}, yshift=1pt},
      every axis plot/.append style={line width=0.7pt},
      clip=false,
      title={(b) CONTIGUOUS BLOCKS},
      xlabel={Context points masked}, yticklabel=\empty,
    ]
    \addplot[forget plot, tfcbody, dashed, line width=0.9pt] coordinates {(-4,1) (79,1)};
    \addplot[forget plot, tfcgreen, mark=*, mark size=1.8pt] table[col sep=comma,x=rate_pct, y=rel_crps,
      restrict expr to domain={\thisrow{series_id}}{1:1}]{figures/missing-data.csv};
    \addplot[forget plot, tfcresearch, mark=square*, mark size=1.5pt] table[col sep=comma,x=rate_pct, y=rel_crps,
      restrict expr to domain={\thisrow{series_id}}{11:11}]{figures/missing-data.csv};
    \addplot[forget plot, tfcengineering, mark=triangle*, mark size=1.9pt] table[col sep=comma,x=rate_pct, y=rel_crps,
      restrict expr to domain={\thisrow{series_id}}{21:21}]{figures/missing-data.csv};
  \end{axis}
  \node[anchor=north, yshift=-2pt] at ($(missing0.below south)!0.5!(missing1.below south)$) {\ref{missingdatalegend}};
\end{tikzpicture}
  \caption{CRPS relative to the unmodified context of each model against
    the share of context points masked, geometric mean over the 12-pair
    subset, for (a) randomly scattered points and (b) contiguous blocks.
    Every model sees identical injected masks. The dashed line is no degradation.}
  \label{fig:missing-data}
\end{figure}

Operational history has gaps, and many foundation models, \tzero among
them, accept them by design: missing values are treated as unobserved
rather than imputed (Section~\ref{sec:architecture}). We quantify what they
cost at inference. On the 12-pair subset (Table~\ref{tab:gifteval-subset})
we inject missingness into the
context window at rates from 10\% to 75\% under two patterns, randomly
scattered points and contiguous blocks that mimic sensor outages, both
masking the same number of points at a given rate, and measure CRPS
relative to the unmodified context of the same pair. Chronos-2 and TiRex run
through the same protocol with identical seeded masks, each scored against
its own clean baseline, and both reproduce their published per-pair CRPS
before injection, Chronos-2 exactly and TiRex within 2\% on 9 of 12 pairs.
Table~\ref{tab:missing-data} and Figure~\ref{fig:missing-data} give the
result. Deleting half the history raises the CRPS of \tzeroalpha by
1.31$\times$ for scattered points and 1.28$\times$ for blocks, and deleting
three quarters by 1.56$\times$ and 1.42$\times$. Chronos-2 degrades least
at every rate and pattern, 1.27$\times$ at 75\% scattered where
\tzeroalpha loses 1.56$\times$ and TiRex 1.50$\times$. Part of its margin
comes from its submission protocol, which forecasts the series of a dataset
jointly, so signal surviving in one series compensates masks in another.
Closing this gap is ongoing work. Against our expectation, scattered points
cost more than blocks for all three models at every rate: scattered
missingness touches every patch, while blocks leave most patches fully
observed. The hard cases are shared. covid\_deaths at 75\% scattered costs
\tzeroalpha 6.5$\times$, TiRex 5.7$\times$ and Chronos-2 4.4$\times$,
since too little history survives to anchor a forecast, while
high-frequency operational series barely move for any model.

\begin{table}[htb]
  \centering
  \caption{Relative CRPS under context missingness (geometric mean
    over the 12-pair subset; 1.00 = the unmodified
    context of each model). Chronos-2 and TiRex run under identical injected masks.}
  \label{tab:missing-data}
  \begin{tabular}{llrrrr}
    \toprule
    Pattern & Model & 10\% & 25\% & 50\% & 75\% \\
    \midrule
    \rowcolor{tfcband} Random points & \tzeroalpha & 1.12 & 1.18 & 1.31 & 1.56 \\
     & Chronos-2 & 1.07 & 1.11 & 1.18 & 1.27 \\
     & TiRex & 1.11 & 1.22 & 1.32 & 1.50 \\
    \midrule
    \rowcolor{tfcband} Contiguous blocks & \tzeroalpha & 1.03 & 1.15 & 1.28 & 1.42 \\
     & Chronos-2 & 1.03 & 1.06 & 1.11 & 1.22 \\
     & TiRex & 1.02 & 1.07 & 1.17 & 1.36 \\
    \bottomrule
  \end{tabular}
\end{table}

\subsection{Context efficiency}
\label{sec:context-efficiency}

Missing data removes points from a fixed window. A complementary question is
how short the window can be before accuracy suffers. Published context
ablations start at 1024 points \citep{hoo2025tabpfnts}. However, many
operational regimes may have even shorter context. We thus measure how the
model reacts to short contexts. On the 12-pair subset
(Table~\ref{tab:gifteval-subset}), we cap the context of
every series at its $c \in \{64, 128, \dots, 8192\}$ most recent points and
score CRPS relative to the same pair at the 8192 cap, which is the training
window length. Because most series are shorter than the larger
caps, we report the realized median context alongside the nominal cap, and a
cap where the series of a pair are already all shorter inherits the score of
the previous cap. The geometric-mean realized context at the full cap is
roughly 1{,}000 points.

\begin{figure}[tb]
  \centering
  \resizebox{0.9\linewidth}{!}{
\begin{tikzpicture}
  \begin{axis}[
      name=contexteff,
      tfcaxis,
      scale only axis, width=12.0cm, height=6.0cm,
      xmode=log, log basis x=2,
      xmin=48, xmax=10000,
      xtick={64,128,256,512,1024,2048,4096,8192},
      xticklabels={64,128,256,512,1024,2048,4096,8192},
      ymode=log,
      ymin=0.75, ymax=3.0,
      ytick={0.8,0.9,1.0,1.1,1.25,1.5,2.0,3.0},
      yticklabels={0.8x,0.9x,1x,1.1x,1.25x,1.5x,2x,3x},
      minor tick num=0,
      xlabel={Realized context length (points per window, log)},
      ylabel={Relative CRPS (vs. full 8192 context)},
      label style={font=\small, color=tfcbody},
      every tick label/.append style={font=\scriptsize},
      every axis plot/.append style={line width=0.7pt},
      legend style={draw=none, fill=none, font=\scriptsize},
      legend pos=north east,
      clip=false,
    ]
    \addplot[dashed, tfcmuted, line width=0.7pt, forget plot] coordinates {(48,1) (10000,1)};
    \addplot[only marks, mark=*, mark size=1.2pt, draw=none, fill=tfcengineering, opacity=0.35, forget plot]
      table[col sep=comma,x=ctx_median, y=skill]{figures/context-efficiency-points.csv};

    \addplot[tfcmuted, mark=triangle*, mark size=2.3pt]
      table[col sep=comma,x=ctx_subhourly, y=skill_subhourly]{figures/context-efficiency-summary-wide.csv};
    \addlegendentry{sub-hourly (n=1)}

    \addplot[tfcresearch, mark=square*, mark size=2.2pt]
      table[col sep=comma,x=ctx_hourly, y=skill_hourly]{figures/context-efficiency-summary-wide.csv};
    \addlegendentry{hourly (n=6)}

    \addplot[tfcgreen, mark=diamond*, mark size=2.4pt]
      table[col sep=comma,x=ctx_daily, y=skill_daily]{figures/context-efficiency-summary-wide.csv};
    \addlegendentry{daily+ (n=5)}

    \addplot[tfcengineering, line width=1.1pt, mark=*, mark size=2.4pt]
      table[col sep=comma,x=ctx_all, y=skill_all]{figures/context-efficiency-summary-wide.csv};
    \addlegendentry{All pairs (gmean)}
  \end{axis}
\end{tikzpicture}}
  \caption{CRPS relative to the full 8192-point context cap vs.\
    realized context length on the 12-pair subset (log $x$-axis).
    Lines are geometric means overall and per frequency band. Points
    are dataset/term pairs.}
  \label{fig:context-efficiency}
\end{figure}
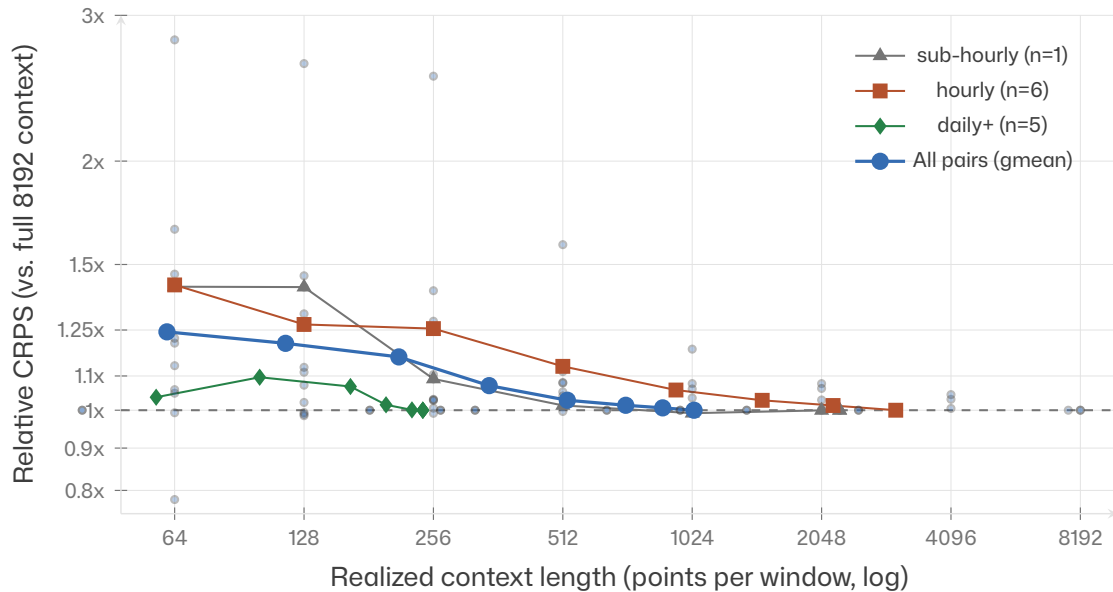

Figure~\ref{fig:context-efficiency} shows the curve. The model is within 5\%
of full-context accuracy from a 1024-point cap and within 1\% from 4096. The
cost of short context concentrates in high-frequency series: at 64 points,
hourly pairs lose 42\% while daily-and-slower pairs lose 3.7\%, because 64
daily points hold several seasonal cycles that 64 hourly points do not.
Truncation is not uniformly harmful. covid\_deaths scores 22\% better from
its last 64 points than from its full 182-point history, where
early-pandemic dynamics mislead more than they inform, and three operational
pairs improve by 1--2\% when truncated. These results show the model does not necessarily need long
histories to be competitive at daily and slower frequencies.

\subsection{Rollout strategy}
\label{sec:eval-rollout}

At inference, the model must forecast over a horizon of $H$ steps. Because it
is trained with contiguous patch masking (CPM), it can predict several patches
ahead in one pass. Following Toto-2 \citep{khwaja2026toto2}, the model falls
back on a rollout whenever $H$ is too long for one pass. In practice
\tzeroalpha predicts 1024 steps, i.e., 32 patches, in one pass and it rolls out autoregressively beyond that. During the rollout it keeps one path per quantile level instead of
reducing to the median between iterations
(Section~\ref{sec:inference}).

In what follows we compare different rollout strategies on a subset of the GIFT-Eval corpus. Specifically, we consider the sub-hourly
datasets whose series are long enough: LOOP Seattle and Bizitobs-L2C at 5
minutes, solar and Jena weather at 10 minutes, electricity, ETT1, and ETT2
at 15 minutes, with at most 32 series per dataset, 138 in total. We request a
forecast of 3072 steps given a context of 8192 steps. To follow accuracy
along the horizon, we cut it into twelve windows of 256 steps and compute,
for each dataset and each window, the CRPS as GIFT-Eval does. We compare four rollout
strategies: the released 1024-step single pass re-anchored on the 8192 true
steps before each iteration, which we use as baseline; the median-path
reduction during the rollout; the quantile-path reduction used for inference
in \tzeroalpha; and a single pass with the CPM mask extended to the full
horizon, three times the released limit. For reference, we also report an oracle forecast
one patch ahead, which predicts each 32-step patch from the 8192 true steps
before it. Curves are geometric means over the seven datasets. The first 1024
steps are the same forward pass under the baseline and the three strategies,
so their ratio is one by construction.

The baseline and the one-patch-ahead forecast stand apart from the other
three. The baseline re-anchors on the 8192 true steps preceding each of its
three iterations, and the one-patch-ahead forecast on the 8192 true steps
preceding each 32-step patch. These two refresh their context with
observations that lie inside the horizon, so neither is feasible for a
forecast made at a single origin. They show how the model would perform with
access to that future information and bound the three strategies that do not
use it: the median path, the quantile paths and the extended single pass,
which answer the 3072-step request from the context available at the origin.

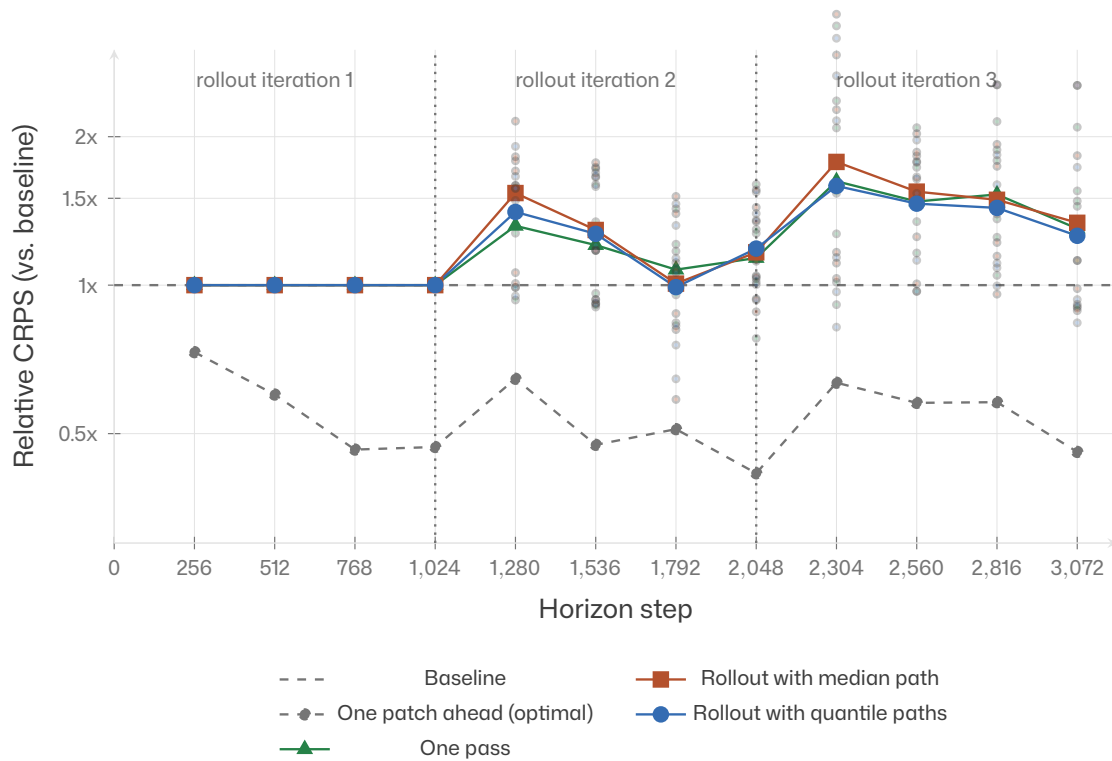
\begin{figure}[tb]
  \centering
  \resizebox{0.9\linewidth}{!}{
\begin{tikzpicture}
  \begin{axis}[
      name=rolloutreduction,
      tfcaxis,
      scale only axis, width=12.0cm, height=5.9cm,
      xmin=0, xmax=3200,
      xtick={0,256,512,768,1024,1280,1536,1792,2048,2304,2560,2816,3072},
      ymode=log,
      ymin=0.3, ymax=3.0,
      ytick={0.5,1.0,1.5,2.0},
      yticklabels={0.5x,1x,1.5x,2x},
      minor tick num=0,
      xlabel={Horizon step},
      ylabel={Relative CRPS (vs. baseline)},
      label style={font=\small, color=tfcbody},
      every tick label/.append style={font=\scriptsize},
      every axis plot/.append style={line width=0.8pt},
      legend to name=rolloutlegend,
      legend columns=2,
      legend style={draw=none, fill=none, font=\scriptsize, /tikz/every even column/.append style={column sep=12pt}},
      clip=false,
    ]
    \addplot[dashed, tfcmuted, line width=0.8pt, forget plot] coordinates {(0,1) (3200,1)};
    \addplot[dotted, tfcmuted, line width=0.8pt, forget plot] coordinates {(1024,0.3) (1024,3.0)};
    \addplot[dotted, tfcmuted, line width=0.8pt, forget plot] coordinates {(2048,0.3) (2048,3.0)};

    \node[anchor=north, font=\scriptsize, color=tfcmuted] at (axis cs:512,2.85) {rollout iteration 1};
    \node[anchor=north, font=\scriptsize, color=tfcmuted] at (axis cs:1536,2.85) {rollout iteration 2};
    \node[anchor=north, font=\scriptsize, color=tfcmuted] at (axis cs:2560,2.85) {rollout iteration 3};

    \addplot[dashed, tfcmuted, mark=*, mark size=1.7pt, forget plot]
      table[col sep=comma,x=window_end, y=next_patch]{figures/rollout-reduction-summary-wide.csv};

    \addplot[only marks, mark=*, mark size=1.1pt, draw=none, fill=tfcgreen, opacity=0.28, forget plot]
      table[col sep=comma,x=window_end, y=single_pass]{figures/rollout-reduction-points-wide.csv};
    \addplot[tfcgreen, mark=triangle*, mark size=2.5pt, forget plot]
      table[col sep=comma,x=window_end, y=single_pass]{figures/rollout-reduction-summary-wide.csv};

    \addplot[only marks, mark=*, mark size=1.1pt, draw=none, fill=tfcresearch, opacity=0.30, forget plot]
      table[col sep=comma,x=window_end, y=median]{figures/rollout-reduction-points-wide.csv};
    \addplot[tfcresearch, mark=square*, mark size=2.4pt, forget plot]
      table[col sep=comma,x=window_end, y=median]{figures/rollout-reduction-summary-wide.csv};

    \addplot[only marks, mark=*, mark size=1.1pt, draw=none, fill=tfcengineering, opacity=0.26, forget plot]
      table[col sep=comma,x=window_end, y=paths]{figures/rollout-reduction-points-wide.csv};
    \addplot[tfcengineering, mark=*, mark size=2.4pt, forget plot]
      table[col sep=comma,x=window_end, y=paths]{figures/rollout-reduction-summary-wide.csv};

    \addlegendimage{dashed, tfcmuted, line width=0.8pt}
    \addlegendentry{Baseline}
    \addlegendimage{tfcresearch, mark=square*, mark size=2.4pt}
    \addlegendentry{Rollout with median path}
    \addlegendimage{dashed, tfcmuted, mark=*, mark size=1.7pt}
    \addlegendentry{One patch ahead (optimal)}
    \addlegendimage{tfcengineering, mark=*, mark size=2.4pt}
    \addlegendentry{Rollout with quantile paths}
    \addlegendimage{tfcgreen, mark=triangle*, mark size=2.5pt}
    \addlegendentry{One pass}
  \end{axis}
  \node[anchor=north, yshift=-4pt] at (rolloutreduction.below south) {\ref{rolloutlegend}};
\end{tikzpicture}}

  \caption{Rollout strategy on long horizons. CRPS per 256-step window of the
    horizon, relative to the baseline, on 138 series from the seven sub-hourly
    GIFT-Eval datasets long enough for an 8192-step context plus a 3072-step
    horizon. \textcolor{figgray}{\emph{Baseline}}: the model predicts 1024 steps ahead in a single
    pass from the most recent context. \textcolor{figgray}{\emph{One patch ahead (optimal)}}: the
    model predicts a single patch, 32 steps ahead from the most recent context.
    \textcolor{figgreen}{\emph{One pass}}: the model predicts all the extended horizon with CPM mask covering the full 3072 steps. \textcolor{figorange}{\emph{Rollout with median path}}: the median
    folded back at each iteration. \textcolor{figblue}{\emph{Rollout with quantile paths}}: one path
    per quantile level, the released strategy. Solid lines are geometric means
    over datasets, while faint points are datasets. Vertical lines separate the
    1024-step rollout iterations. The first iteration is the same forward pass
  under the baseline, the one pass, and both rollouts.}

  \label{fig:rollout-ablation}
\end{figure}

Figure~\ref{fig:rollout-ablation}
shows
the three strategies relative to the baseline. Up-to-date context makes the
clearest difference. The optimal forecast, one patch ahead from the true
context, scores 0.54 times the baseline over the horizon. The baseline itself
re-anchors on the truth only every 1024 steps, and the rollouts, which never
do, score 2.5 to 2.6 times the optimal over the rolled-out steps. The cost of
forecasting without fresh context is large. Averaged over datasets, the
rollouts score 1.4 to 1.5 times the baseline on the third iteration and 1.6 to
1.8 times in their worst window, steps 2048--2304. The rollout with quantile
paths is almost systematically below the median path. In seven of the eight
rolled-out windows and on six of the seven datasets, by 4.8\% on average. It is
marginally better than the one pass over the full horizon, 1.31 against 1.32,
ahead in five of the eight windows and on four of the seven datasets. Both the
model and the rollout are stable. The one pass decodes three times the released
horizon without breaking down. No rollout strategy drifts away from the others
as the iterations proceed.

The lever on long horizons is therefore context, not the reducer: the
strategies differ by a few percent, re-anchoring on observed data every 1024
steps removes the 30\% the rollout costs, and re-anchoring every patch
removes a further factor of 1.9. That lever belongs to the user who can
re-issue the forecast as observations land. Whoever has to commit to the
whole 3072-step horizon at one origin pays the rollout cost.
Between the strategies, the median path is
dominated, less accurate than the one pass at 2.4 times its inference time.
The quantile paths cost 14 times the one pass and buy a 3\% gain on the third
iteration only, so the one pass is the cheap mode up to three times the
released horizon and the quantile-path rollout the accurate one at the far
end of the horizon.

\subsection{Use cases}
\subsubsection{Hourly electricity demand in Australia}
\label{sec:usecase}

This benchmark comes from skforecast \citep{skforecast}, a Python forecasting
library with a scikit-learn interface. Its user guide on foundation models
\citep{skforecast2026foundation} backtests eight zero-shot models on the
half-hourly electricity demand of Victoria, Australia. The data are taken from
the \texttt{tsibbledata} collection \citep{oharawild2022tsibbledata} and
aggregated to hourly means. The goal is to predict electricity demand given the
temperature, and a holiday indicator. Temperature and holiday are supplied over
the forecast horizon as known-future covariates, so a model receives the
realized temperature of the day it forecasts. The guide backtests over one-day
windows from October 31 to December 30, 2014, 61 windows in total. Each
window forecasts the next 24 hours from the preceding 500 observations, three
weeks of hourly history. The score is the mean absolute error, in megawatts, of
the forecast of each model over all windows. For most models the forecast
corresponds to the median. TimesFM, TabICL, and Nori however directly produce
a mean prediction. Whenever possible, we reproduced the results of the guide with its own code, on the
same backtest. Table~\ref{tab:usecase-skforecast} shows the results.

\begin{table}[htb]
  \centering
  \caption{Mean absolute error (MW) for different TSFMs on the original
  skforecast Victoria electricity demand benchmark. The year 2014 is present in
  GIFT-Eval pretraining corpus, leading to data leakage for most TSFMs.}
  \label{tab:usecase-skforecast}
  \begin{tabular}{lllr}
    \toprule
    \tfcheadrow{Model} & \tfcheadrow{Covariates} & \tfcheadrow{Point} & \tfcheadrow{MAE $\downarrow$} \\
    \midrule
    TS-ICL & \tfccheck & median & \textbf{139.93} \\
    \rowcolor{tfcband} \tzeroalpha & \tfccheck & median & 144.13 \\
    Chronos-2 (small) & \tfccheck & median & 160.05 \\
    TimesFM-2.5 (200M) & \tfccross & mean & 188.28 \\
    TabPFN-TS & \tfccheck & median & 190.97 \\
    Nori & \tfccheck & mean & 193.33 \\
    Moirai-2.0-R (small) & \tfccross & median & 196.82 \\
    TabICL & \tfccheck & mean & 210.75 \\
    \midrule
    Seasonal naive (24 h) & \tfccross & --- & 330.59 \\
    \bottomrule
  \end{tabular}
\end{table}

The guide rightly warns about leakage. The GIFT-Eval pretraining corpus
\citep{aksu2024gifteval}, on which \tzeroalpha and many other public models
train, holds the Victoria demand series twice. The
\texttt{australian\_\allowbreak electricity\_\allowbreak demand} and the
\texttt{elecdemand} datasets both contain the 2014 data of the backtest
\citep{godahewa2021australian}. To avoid this leakage, we rerun the backtest on
the year 2024. Every real series of the corpus ends before 2022, so the 2024
window is free of leakage. It is at least guaranteed that \tzeroalpha
has never seen the data during training.

We assemble the three variates from public sources, in the layout used by the
guide. Demand is the five-minute total demand of the VIC1 region
published by the Australian Energy Market Operator (AEMO)
\citep{aemo2024priceanddemand}, averaged to hourly means. Temperature is the
hourly observation of the Melbourne (Olympic Park) station, the successor of
the Melbourne Regional Office site that tsibbledata used, taken from the
Integrated Surface Database of the US National Centers for Environmental
Information \citep{noaa2001isd}. The station covers 97\% of the hours. We fill
the remaining gaps from the ERA5 reanalysis served by Open-Meteo
\citep{zippenfenig2023openmeteo}. The holiday indicator marks the 13 Victorian
public holidays of 2024, taken from the \texttt{holidays} Python package
\citep{vacanza2026holidays}. We run the script of the guide unchanged except for the
dates: 62 one-day windows from October 31 to December 31, 2024. Each model runs at two context lengths, 512 and 8192 hours, roughly
three weeks and eleven months of history. All runs use an Apple M4 Pro, with
accelerator hardware whenever it can be used by the model, otherwise on CPU
only. Runtime is the wall-clock time of one pass over the 62 forecasts of 24 hours
each, produced by the backtesting routine of skforecast. Weight loading and other overhead
costs are excluded. Nori did not complete the 8192-context pass within one
hour, and TabICL needed 25 minutes for it.

Table~\ref{tab:usecase-2024}
reports
the error at both contexts, the runtime of one pass, and the hardware. The error of
every model is 2 to 2.6 times its value on the 2014 window. Leakage alone
does not explain the gap. Indeed, the 2024 window is also harder, since a
24-hour seasonal naive forecast scores 614.6~MW on it against 330.6~MW on the
2014 window. Chronos-2 leads at both contexts. \tzeroalpha remains among the
most accurate models while being one of the fastest. \tzerobeta places second
at both contexts, and the gap to \tzeroalpha widens with the context it is
given: 2.8\% lower error at 512 hours and 10.1\% lower at 8192, where it closes
most of the distance to Chronos-2.

\begin{table}[htb]
  \centering

  \caption{Mean absolute error (MW) for different TSFMs following the
    skforecast Victoria electricity demand benchmark for the year 2024. The
    year
  2024 is absent from the GIFT-Eval pretraining corpus. The best value of each
       error column is in bold.}

  \label{tab:usecase-2024}
  \small
\begin{tabular}{llrrrrl}
\toprule
& & \multicolumn{2}{c}{\tfcheadrow{MAE $\downarrow$}} & \multicolumn{2}{c}{\tfcheadrow{Runtime (s)}} & \\
\tfcheadrow{Model} & \tfcheadrow{Cov.} & \tfcheadrow{512} & \tfcheadrow{8192} & \tfcheadrow{512} & \tfcheadrow{8192} & \tfcheadrow{Hardware} \\
\midrule
Chronos-2 & \tfccheck & \textbf{348.87} & \textbf{324.24} & 1.1 & 10.5 & GPU \\
\rowcolor{tfcbandblue} \tzerobeta & \tfccheck & 368.89 & 334.92 & 3.8 & 9.9 & GPU \\
Chronos-2 (small) & \tfccheck & 369.38 & 342.62 & 0.6 & 2.9 & GPU \\
TS-ICL & \tfccheck & 358.19 & 365.35 & 2.7 & 25.2 & CPU \\
\rowcolor{tfcband} \tzeroalpha & \tfccheck & 379.58 & 372.51 & 1.2 & 4.5 & GPU \\
TabICL & \tfccheck & 463.10 & 394.17 & 42.3 & 1513.3 & CPU \\
Moirai-2.0-R (small) & \tfccross & 456.87 & 433.45 & 0.3 & 1.2 & CPU \\
TimesFM-2.5 (200M) & \tfccross & 476.61 & 453.73 & 3.4 & 13.1 & CPU \\
Nori & \tfccheck & 388.60 & ---$^{*}$ & 54.2 & ---$^{*}$ & GPU \\
\midrule
Seasonal naive (24 h) & \tfccross & 614.57 & 614.57 & --- & --- &  \\
\bottomrule
\multicolumn{7}{l}{\footnotesize $^{*}$ Not completed within one hour on this hardware.} \\
\end{tabular}

\end{table}

\subsubsection{Forecasting and trading electricity in Texas}
\label{sec:usecase-ercot}

\emph{\textcolor{tfcmuted}{This Section~\ref{sec:usecase-ercot} has been
written by Macrocosm,\footnote{Macrocosm is developing simulation-driven
  world models, starting with the US electric grid:
  \href{https://mcsm.ai/}{\texttt{mcsm.ai}}.} which conducted an independent
evaluation of \tzero models.}}

We evaluate point and probabilistic forecasts of hourly real-time (SCED)
electricity prices in the four ERCOT (the electricity operator of Texas) load zones, from January 2024 to May 2026,
and assess their downstream trading value. We report MAE, CRPS and interval coverage. Each model forecasts the next operating day at 09:00 Central Time
on the preceding day, using only information available at that gate, including
21 covariates covering load, renewable generation, outages and weather.
As a baseline, we repeat the hourly price profile of the most recent complete operating day available at the forecast gate.

For the downstream trading evaluation, we compare five rules: three that map the predicted median spread against the day-ahead price to positions using step, linear or sigmoid functions, a Kelly rule based on the forecast distribution, and a constant-volatility rule that scales positions inversely with forecast uncertainty. For each model, we select the rule with the highest annualized daily dollar Sharpe over the evaluation window. Positions depend on the predicted spread against the day-ahead market
(DAM) price and, where applicable, uncertainty. Strategy selection is retrospective.
Trading uses a fixed 100\,MW reference size per zone, without reinvestment or
market impact. Sharpe is $\sqrt{365}$ times mean daily profit divided by its
standard deviation. An always-short control sells day-ahead every hour without
using forecasts.

\begin{table}[htb]
  \centering
  \small
  \caption{Forecast and trading results. MAE and CRPS are in \$/MWh;
    coverage is for the nominal 80\% interval. Trading results use each model's
    best rule on the same window. TimesFM-3 has one fewer forecast day.}
  \label{tab:ercot-headline}
  \begin{tabular}{lrrrrr}
\toprule
Model & MAE & CRPS & Coverage & Profit (\$M) & Sharpe \\
\midrule
TimesFM-3 & \textbf{13.78} & \textbf{10.94} & 0.82 & \textbf{16.42} & \textbf{1.47} \\
\rowcolor{tfcband} \tzeroalpha & 14.55 & 11.50 & 0.74 & 15.00 & 1.34 \\
\rowcolor{tfcbandblue} \tzerobeta & 14.46 & 11.37 & 0.75 & 14.94 & 1.29 \\
Lagged price & 23.57 & --- & --- & 14.11 & 1.27 \\
Chronos-2 & 13.90 & 11.11 & 0.75 & 14.62 & 1.25 \\
TiRex-2 & 15.22 & 12.08 & 0.76 & 11.21 & 0.97 \\
TimesFM-2.5 & 16.76 & 13.26 & 0.81 & 11.53 & 0.97 \\
\midrule
Always short & --- & --- & --- & 11.82 & 0.96 \\
\bottomrule
\end{tabular}

\end{table}

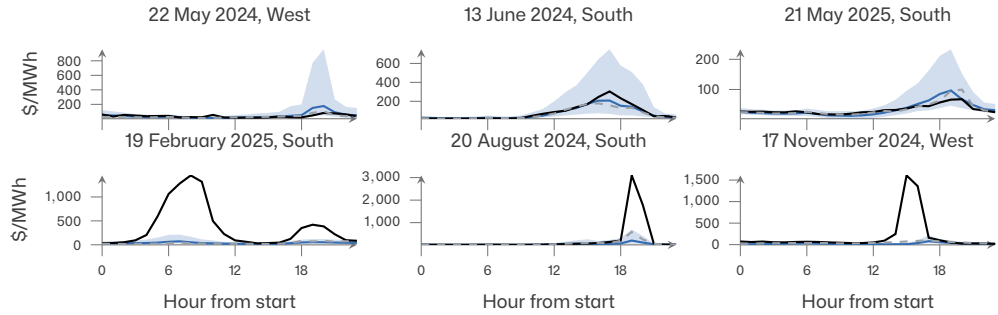
\begin{figure}[!htb]
  \centering
\begin{tikzpicture}
\begin{groupplot}[
  group style={group size=3 by 2, horizontal sep=0.85cm, vertical sep=0.75cm, x descriptions at=edge bottom},
  charliefan, width=0.30\linewidth, height=2.5cm,
  xlabel={Hour from start},
  title style={font=\scriptsize},
  label style={font=\scriptsize}, scaled y ticks=false, tick label style={font=\tiny},
]
\nextgroupplot[title={22 May 2024, West}, ylabel={\$/MWh}]
\addplot[draw=none, name path=charlielowgood] table[col sep=comma,x=index,y=tzerobetalo]{figures/macrocosm/data/tracking_days__2024-05-22__LZ_WEST.csv};
\addplot[draw=none, name path=charliehighgood] table[col sep=comma,x=index,y=tzerobetahi]{figures/macrocosm/data/tracking_days__2024-05-22__LZ_WEST.csv};
\addplot[charlietzerobeta, fill opacity=0.22] fill between[of=charlielowgood and charliehighgood];
\addplot[charlietzerobeta, thick] table[col sep=comma,x=index,y=tzerobetamid]{figures/macrocosm/data/tracking_days__2024-05-22__LZ_WEST.csv};
\addplot[black, thick] table[col sep=comma,x=index,y=realized]{figures/macrocosm/data/tracking_days__2024-05-22__LZ_WEST.csv};
\addplot[charlierule, dashed, thick] table[col sep=comma,x=index,y=dam]{figures/macrocosm/data/tracking_days__2024-05-22__LZ_WEST.csv};
\nextgroupplot[title={13 June 2024, South}]
\addplot[draw=none, name path=charlielowgood] table[col sep=comma,x=index,y=tzerobetalo]{figures/macrocosm/data/tracking_days__2024-06-13__LZ_SOUTH.csv};
\addplot[draw=none, name path=charliehighgood] table[col sep=comma,x=index,y=tzerobetahi]{figures/macrocosm/data/tracking_days__2024-06-13__LZ_SOUTH.csv};
\addplot[charlietzerobeta, fill opacity=0.22] fill between[of=charlielowgood and charliehighgood];
\addplot[charlietzerobeta, thick] table[col sep=comma,x=index,y=tzerobetamid]{figures/macrocosm/data/tracking_days__2024-06-13__LZ_SOUTH.csv};
\addplot[black, thick] table[col sep=comma,x=index,y=realized]{figures/macrocosm/data/tracking_days__2024-06-13__LZ_SOUTH.csv};
\addplot[charlierule, dashed, thick] table[col sep=comma,x=index,y=dam]{figures/macrocosm/data/tracking_days__2024-06-13__LZ_SOUTH.csv};
\nextgroupplot[title={21 May 2025, South}]
\addplot[draw=none, name path=charlielowgood] table[col sep=comma,x=index,y=tzerobetalo]{figures/macrocosm/data/tracking_days__2025-05-21__LZ_SOUTH.csv};
\addplot[draw=none, name path=charliehighgood] table[col sep=comma,x=index,y=tzerobetahi]{figures/macrocosm/data/tracking_days__2025-05-21__LZ_SOUTH.csv};
\addplot[charlietzerobeta, fill opacity=0.22] fill between[of=charlielowgood and charliehighgood];
\addplot[charlietzerobeta, thick] table[col sep=comma,x=index,y=tzerobetamid]{figures/macrocosm/data/tracking_days__2025-05-21__LZ_SOUTH.csv};
\addplot[black, thick] table[col sep=comma,x=index,y=realized]{figures/macrocosm/data/tracking_days__2025-05-21__LZ_SOUTH.csv};
\addplot[charlierule, dashed, thick] table[col sep=comma,x=index,y=dam]{figures/macrocosm/data/tracking_days__2025-05-21__LZ_SOUTH.csv};
\nextgroupplot[title={19 February 2025, South}, ylabel={\$/MWh}]
\addplot[draw=none, name path=charlielowgood] table[col sep=comma,x=index,y=tzerobetalo]{figures/macrocosm/data/spike_anatomy__2025-02-19__LZ_SOUTH.csv};
\addplot[draw=none, name path=charliehighgood] table[col sep=comma,x=index,y=tzerobetahi]{figures/macrocosm/data/spike_anatomy__2025-02-19__LZ_SOUTH.csv};
\addplot[charlietzerobeta, fill opacity=0.22] fill between[of=charlielowgood and charliehighgood];
\addplot[charlietzerobeta, thick] table[col sep=comma,x=index,y=tzerobetamid]{figures/macrocosm/data/spike_anatomy__2025-02-19__LZ_SOUTH.csv};
\addplot[black, thick] table[col sep=comma,x=index,y=realized]{figures/macrocosm/data/spike_anatomy__2025-02-19__LZ_SOUTH.csv};
\addplot[charlierule, dashed, thick] table[col sep=comma,x=index,y=dam]{figures/macrocosm/data/spike_anatomy__2025-02-19__LZ_SOUTH.csv};
\nextgroupplot[title={20 August 2024, South}]
\addplot[draw=none, name path=charlielowgood] table[col sep=comma,x=index,y=tzerobetalo]{figures/macrocosm/data/spike_anatomy__2024-08-20__LZ_SOUTH.csv};
\addplot[draw=none, name path=charliehighgood] table[col sep=comma,x=index,y=tzerobetahi]{figures/macrocosm/data/spike_anatomy__2024-08-20__LZ_SOUTH.csv};
\addplot[charlietzerobeta, fill opacity=0.22] fill between[of=charlielowgood and charliehighgood];
\addplot[charlietzerobeta, thick] table[col sep=comma,x=index,y=tzerobetamid]{figures/macrocosm/data/spike_anatomy__2024-08-20__LZ_SOUTH.csv};
\addplot[black, thick] table[col sep=comma,x=index,y=realized]{figures/macrocosm/data/spike_anatomy__2024-08-20__LZ_SOUTH.csv};
\addplot[charlierule, dashed, thick] table[col sep=comma,x=index,y=dam]{figures/macrocosm/data/spike_anatomy__2024-08-20__LZ_SOUTH.csv};
\nextgroupplot[title={17 November 2024, West}]
\addplot[draw=none, name path=charlielowgood] table[col sep=comma,x=index,y=tzerobetalo]{figures/macrocosm/data/spike_anatomy__2024-11-17__LZ_WEST.csv};
\addplot[draw=none, name path=charliehighgood] table[col sep=comma,x=index,y=tzerobetahi]{figures/macrocosm/data/spike_anatomy__2024-11-17__LZ_WEST.csv};
\addplot[charlietzerobeta, fill opacity=0.22] fill between[of=charlielowgood and charliehighgood];
\addplot[charlietzerobeta, thick] table[col sep=comma,x=index,y=tzerobetamid]{figures/macrocosm/data/spike_anatomy__2024-11-17__LZ_WEST.csv};
\addplot[black, thick] table[col sep=comma,x=index,y=realized]{figures/macrocosm/data/spike_anatomy__2024-11-17__LZ_WEST.csv};
\addplot[charlierule, dashed, thick] table[col sep=comma,x=index,y=dam]{figures/macrocosm/data/spike_anatomy__2024-11-17__LZ_WEST.csv};
\end{groupplot}
\end{tikzpicture}
  \caption{\tzerobeta{} uncertainty on six illustrative days: three peaks inside the
    10--90\% band (top) and three missed spikes (bottom). Blue: forecast median and band; black: realized price;
    dashed grey: day-ahead price. Vertical scales differ; none is clipped.}
  \label{fig:ercot-spike}
\end{figure}

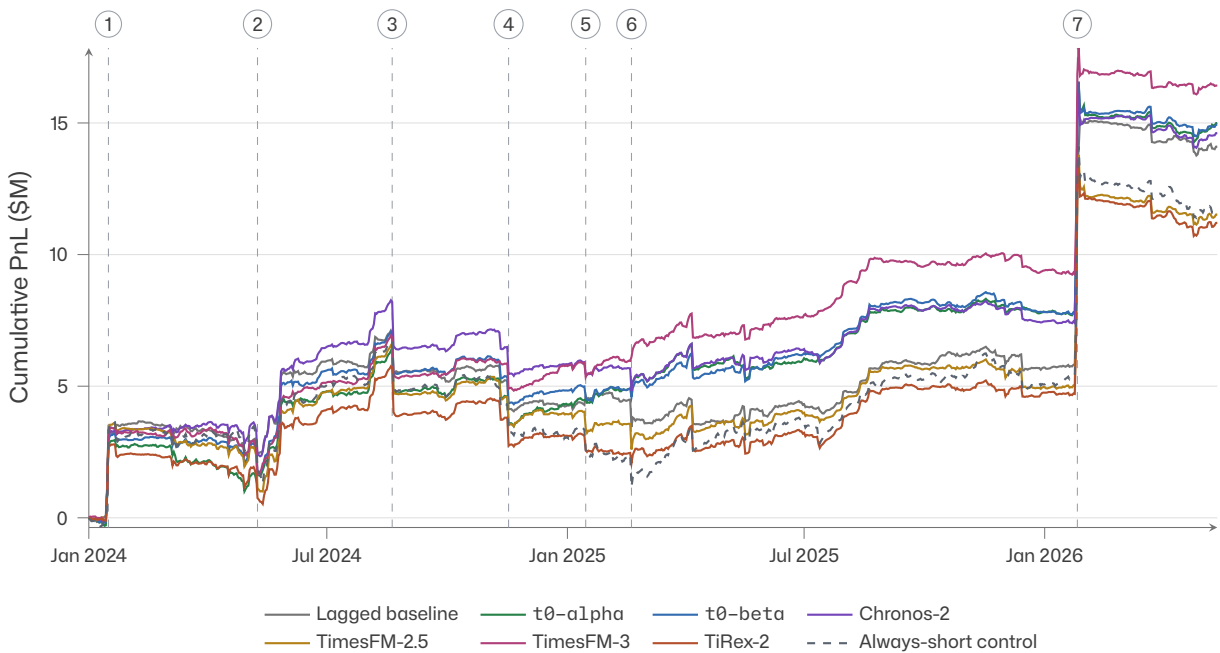
\begin{figure*}[t]
   \centering
\begingroup

\begin{tikzpicture}
\begin{axis}[
    charlieaxis,
    width=\linewidth,
    height=0.48\linewidth,
    xlabel={},
    ylabel={Cumulative PnL (\$M)},
    xmin=0,
    xmax=863,
    xtick={0,182,366,547,731},
    xticklabels={Jan 2024,Jul 2024,Jan 2025,Jul 2025,Jan 2026},
    clip=false,
    legend style={
        at={(0.5,-0.14)},
        anchor=north,
        draw=none,
        font=\scriptsize,
        /tikz/every even column/.append style={column sep=6pt}
    },
    legend columns=4,
]

  \draw[charlierule,dashed,thin]
    (axis cs:15,\pgfkeysvalueof{/pgfplots/ymin}) --
    (axis cs:15,\pgfkeysvalueof{/pgfplots/ymax});
  \draw[charlierule,dashed,thin]
    (axis cs:129,\pgfkeysvalueof{/pgfplots/ymin}) --
    (axis cs:129,\pgfkeysvalueof{/pgfplots/ymax});
  \draw[charlierule,dashed,thin]
    (axis cs:232,\pgfkeysvalueof{/pgfplots/ymin}) --
    (axis cs:232,\pgfkeysvalueof{/pgfplots/ymax});
  \draw[charlierule,dashed,thin]
    (axis cs:321,\pgfkeysvalueof{/pgfplots/ymin}) --
    (axis cs:321,\pgfkeysvalueof{/pgfplots/ymax});
  \draw[charlierule,dashed,thin]
    (axis cs:380,\pgfkeysvalueof{/pgfplots/ymin}) --
    (axis cs:380,\pgfkeysvalueof{/pgfplots/ymax});
  \draw[charlierule,dashed,thin]
    (axis cs:415,\pgfkeysvalueof{/pgfplots/ymin}) --
    (axis cs:415,\pgfkeysvalueof{/pgfplots/ymax});
  \draw[charlierule,dashed,thin]
    (axis cs:756,\pgfkeysvalueof{/pgfplots/ymin}) --
    (axis cs:756,\pgfkeysvalueof{/pgfplots/ymax});

  \tikzset{ercotevent/.style={
    circle,
    draw=charlierule,
    fill=white,
    inner sep=1.5pt,
    font=\scriptsize,
    yshift=9pt
  }}

  \node[ercotevent] at
    (axis cs:15,\pgfkeysvalueof{/pgfplots/ymax}) {1};
  \node[ercotevent] at
    (axis cs:129,\pgfkeysvalueof{/pgfplots/ymax}) {2};
  \node[ercotevent] at
    (axis cs:232,\pgfkeysvalueof{/pgfplots/ymax}) {3};
  \node[ercotevent] at
    (axis cs:321,\pgfkeysvalueof{/pgfplots/ymax}) {4};
  \node[ercotevent] at
    (axis cs:380,\pgfkeysvalueof{/pgfplots/ymax}) {5};
  \node[ercotevent] at
    (axis cs:415,\pgfkeysvalueof{/pgfplots/ymax}) {6};
  \node[ercotevent] at
    (axis cs:756,\pgfkeysvalueof{/pgfplots/ymax}) {7};
\addplot[charlienaivertdtwo,thick]
    table[col sep=comma,x=index,y=naivertdtwo]{figures/macrocosm/data/cumulative_pnl.csv};
\addlegendentry{Lagged baseline}

\addplot[charlietzeroalpha,thick]
    table[col sep=comma,x=index,y=tzeroalpha]{figures/macrocosm/data/cumulative_pnl.csv};
\addlegendentry{\tzeroalpha{}}

\addplot[charlietzerorc,thick]
    table[col sep=comma,x=index,y=tzerorc]{figures/macrocosm/data/cumulative_pnl.csv};
\addlegendentry{\tzerobeta{}}

\addplot[charliechronostwo,thick]
    table[col sep=comma,x=index,y=chronostwo]{figures/macrocosm/data/cumulative_pnl.csv};
\addlegendentry{Chronos-2}

\addplot[charlietimesfmtwopfive,thick]
    table[col sep=comma,x=index,y=timesfmtwopfive]{figures/macrocosm/data/cumulative_pnl.csv};
\addlegendentry{TimesFM-2.5}

\addplot[charlietimesfmthree,thick]
    table[col sep=comma,x=index,y=timesfmthree]{figures/macrocosm/data/cumulative_pnl.csv};
\addlegendentry{TimesFM-3}

\addplot[charlietirextwo,thick]
    table[col sep=comma,x=index,y=tirextwo]{figures/macrocosm/data/cumulative_pnl.csv};
\addlegendentry{TiRex-2}

\addplot[charliecontrol,dashed,thick]
    table[col sep=comma,x=index,y=alwaysshort]{figures/macrocosm/data/cumulative_pnl.csv};
\addlegendentry{Always-short control}

\end{axis}
\end{tikzpicture}
\endgroup
   \caption[Cumulative trading profits in ERCOT]{%
     Cumulative trading profits using each model's retrospectively
     selected rule, with a fixed 100\,MW reference size per zone,
     without reinvestment or market impact.
     The dashed curve is the always-short control.
     Numbered vertical lines mark:
     (1) 16 Jan 2024, Winter Storm Heather;
     (2) 9 May 2024, evening net-load ramp;
     (3) 20 Aug 2024, record load and evening solar ramp;
     (4) 17 Nov 2024, evening ramp;
     (5) 15 Jan 2025, midday cold;
     (6) 19 Feb 2025, cold snap and Round Rock congestion;
     (7) 26 Jan 2026, Winter Storm Fern.
   }
   \label{fig:ercot-pnl}
 \end{figure*}

All pretrained models have lower MAE than the lagged baseline (Table~\ref{tab:ercot-headline}). TimesFM-3 has the lowest MAE and CRPS. \tzerobeta{} scores better than \tzeroalpha{} on both measures, but \tzeroalpha{} has a higher trading Sharpe. The two \tzero{} models' 80\% prediction intervals contain only 74--75\% of observed prices: they cover some peaks but miss extreme spikes (Figure~\ref{fig:ercot-spike}). Trading profits may also reflect persistent differences between SCED and DAM prices: even the always-short rule makes money without using forecasts. Figure~\ref{fig:ercot-pnl} shows cumulative trading profits for each model's selected rule alongside this control. Correcting forecast bias using the previous 30 days leaves the trading ranking unchanged.

\section{Limitations}
\label{sec:limitations}

\tzeroalpha is the first release of the \tzero family. This section states
its known limits, which serve as outlines for future directions.

\paragraph{Limited quantile levels} The model emits five native quantile
levels, from $0.1$ to $0.9$. A requested level that falls between two of them
is interpolated, and a level outside the range returns the nearest native
one, so a request for the 0.99 quantile returns the 0.9 value. The grid is nonetheless fine enough for
the model to stay calibrated inside its range
(Section~\ref{sec:calibration}), and post-hoc IQF tails recover most of what
is lost outside it at no training cost (Section~\ref{sec:tails}). A denser
native grid, as Chronos-2 uses \citep{ansari2025chronos2}, is the remedy on
the training side. \tzerobeta already emits 21 native levels, from $0.01$
to $0.99$.

\paragraph{Limited support for long lead times} Recall the retail problem of Figure~\ref{fig:multivariate-example}: gloves and flip-flops are manufactured and shipped months before they reach a shelf, so an apparel buyer commits the winter order in spring, six to nine months ahead. The buyer therefore needs the demand of the coming winter forecast from a context that stops at the order date, leaving a gap between the end of the context and the first forecast step. We call this gap the \emph{lead time}\label{gl:lead-time}. Such lead times are common in production and procurement, where decisions must often be made well before their outcomes are observed. In principle, the model naturally supports such use cases: at inference time, the gap can be represented by blanking the corresponding portion of the context, a pattern the model encounters during training through contiguous patch masking. However, the maximum lead time that can be handled this way is 1024 time steps minus one patch (32 time steps for \tzeroalpha). Beyond this limit, one must resort to autoregressive rollout, as described in Section~\ref{sec:eval-rollout}.

\paragraph{Extend covariate lift studies and benchmark} We measure on fev-bench the
lift that covariates bring, but the benchmark does not separate covariate
skill from univariate accuracy (Section~\ref{sec:related}). Two questions
therefore stay open. Does the model use every covariate that carries
information? Does it ignore the ones that carry none? Answering these questions
calls for benchmarks built for the purpose.

\paragraph{Support for numerical covariates only} The model conditions on past
and known-future covariates as long as they are numeric series. A user who
holds text, images, categorical metadata, or static attributes has to encode
them as numbers first. Embedding the text description of a variate would
instead let one model learn across series what a covariate means, the route
taken by metadata-conditioned forecasters \citep{dong2024metatst,
dutta2025charm} and by the multimodal models that read text alongside the
series \citep{wang2025chattime, xie2025chatts}.

This would help address cold-start\label{gl:cold-start} predictions. A product about to
launch has no past of its own. It has a photograph, a title, a category, and a
description, and the products already on the shelf carry those attributes
together with their sales. A model that read all of it could place the new item
among the old ones and forecast from that neighborhood. No purely numeric
method can do so, and \tzeroalpha is one.

\paragraph{No joint distribution} The model emits marginal quantiles, so its
output ties neither the steps of a horizon nor the variates of a sample
together (Section~\ref{sec:problem}).

\paragraph{No controlled scaling study} This report evaluates the
102M-parameter \tzeroalpha model in depth and reports public-benchmark results
for its 256M-parameter successor, \tzerobeta. It does not study the scaling of the
\tzero{} family. It does not isolate the effect of parameter count from other
differences between the models and their evaluation protocols. Across model
families, parameter count alone does not yet determine accuracy. TimesFM-3.0
achieves a GIFT-Eval score of 0.4557 CRPS with 331M parameters, compared with
0.4759 for the 2.5B-parameter Toto-2.0 model
(Figure~\ref{fig:size-vs-crps}).
Within the Toto-2.0 family, increasing model size improves accuracy, but the
gain from 313M to 2.5B parameters is comparatively small, from 0.4814 to 0.4759
CRPS. Another model, Timer-S1, with 8.3B parameters, 0.75B of them activated
per token, scores 0.4853 \citep{liu2026timers1}. Xihe-max with 1.5B parameters
scores 0.4905 \citep{sun2025xihe}. Both models sit behind TimesFM-3.0 on
GIFT-Eval, a model less than a quarter of their size. Establishing how \tzero
benefits from additional model capacity, training data, and compute requires
controlled experiments.

\section{Conclusion}

We presented and released the first two members of the \tzero family:
\tzeroalpha, a 102M-parameter open-weights forecasting model, and
\tzerobeta, its 256M-parameter successor. Both use factorized time\slash
variate attention to give target series, past covariates, and known-future
covariates one representation and one forward pass. Pretraining combines
curated public corpora with synthetic families whose covariate-to-target
structure is known by construction.

For \tzeroalpha, we report accuracy on GIFT-Eval and fev-bench together with
the covariate lift, calibration and tail behavior, rank stability across eleven
metrics, robustness to missing history, and context efficiency
(Section~\ref{sec:evaluation}). Passing covariates to the same checkpoint
raises fev-bench skill by 6.3 points on the tasks with known-future covariates
and 2.7 points on those with past covariates only
(Section~\ref{sec:fevbench}).

\tzerobeta already ranks among the strongest zero-shot models we compare. On
the same three public benchmarks it reaches 0.4738 CRPS on GIFT-Eval, third
among those models and 1.4\% short of second place, 46.7 skill on fev-bench,
third again and 46.2 on its covariate-informed half, and 0.5551 CRPS on TIME
(Section~\ref{sec:eval-protocol}).

\tzeroalpha and \tzerobeta are the first releases of the \tzero family. Their
weights as the \texttt{tfc-t0} package and the notebook reproducing the
GIFT-Eval results are all made public. The open cases of
Section~\ref{sec:limitations} set the agenda for the
next iterations.

\addtocontents{toc}{\protect\setcounter{tocdepth}{-1}}

\bibliographystyle{tfcnat}
\bibliography{references}

@article{ansari2024chronos,
  title = {Chronos: Learning the Language of Time Series},
  author = {Ansari, Abdul Fatir and Stella, Lorenzo and Turkmen, Caner and Zhang
            , Xiyuan and Mercado, Pedro and Shen, Huibin and Shchur, Oleksandr
            and Rangapuram, Syama Sundar and Pineda Arango, Sebastian and Kapoor,
            Shubham and Zschiegner, Jasper and Maddix, Danielle C. and Wang, Hao
            and Mahoney, Michael W. and Torkkola, Kari and Gordon Wilson, Andrew
            and Bohlke-Schneider, Michael and Wang, Yuyang},
  journal = {Transactions on Machine Learning Research},
  year = {2024},
  url = {https://arxiv.org/abs/2403.07815},
}

@article{ansari2025chronos2,
  title = {Chronos-2: From Univariate to Universal Forecasting},
  author = {Ansari, Abdul Fatir and Shchur, Oleksandr and K{\"u}ken, Jaris and
            Auer, Andreas and Han, Boran and Mercado, Pedro and Rangapuram,
            Syama Sundar and Shen, Huibin and Stella, Lorenzo and Zhang, Xiyuan
            and Goswami, Mononito and Kapoor, Shubham and Maddix, Danielle C.
            and Guerron, Pablo and Hu, Tony and Yin, Junming and Erickson, Nick
            and Mutalik Desai, Prateek and Wang, Hao and Rangwala, Huzefa and
            Karypis, George and Wang, Yuyang and Bohlke-Schneider, Michael},
  journal = {arXiv preprint arXiv:2510.15821},
  year = {2025},
  url = {https://arxiv.org/abs/2510.15821},
}

@inproceedings{beck2024xlstm,
  title = {{xLSTM}: Extended Long Short-Term Memory},
  author = {Beck, Maximilian and P{\"o}ppel, Korbinian and Spanring, Markus and
            Auer, Andreas and Prudnikova, Oleksandra and Kopp, Michael and
            Klambauer, G{\"u}nter and Brandstetter, Johannes and Hochreiter, Sepp
            },
  booktitle = {Advances in Neural Information Processing Systems (NeurIPS)},
  year = {2024},
  url = {https://arxiv.org/abs/2405.04517},
}

@inproceedings{auer2025tirex,
  title = {TiRex: Zero-Shot Forecasting Across Long and Short Horizons with
           Enhanced In-Context Learning},
  author = {Auer, Andreas and Podest, Patrick and Klotz, Daniel and B{\"o}ck,
            Sebastian and Klambauer, G{\"u}nter and Hochreiter, Sepp},
  booktitle = {Advances in Neural Information Processing Systems (NeurIPS)},
  year = {2025},
  url = {https://arxiv.org/abs/2505.23719},
}

@article{lenaour2026tsicl,
  title = {{TS-ICL}: A Flexible Time-Indexed Foundation Model for Time Series
           via In-Context Learning},
  author = {Le Naour, Etienne and Nabil, Tahar and Petralia, Adrien},
  journal = {arXiv preprint arXiv:2606.05878},
  year = {2026},
  url = {https://arxiv.org/abs/2606.05878},
}

@article{apollopfn2026time,
  title={Time-aware prior fitted networks for zero-shot forecasting with exogenous variables},
  author={Potapczynski, Andres and Selvam, Ravi Kiran and Konstantinova, Tatiana and Ramasubramanian, Shankar and Wolff, Malcolm and Olivares, Kin G and Ma, Ruijun and Cao, Mengfei and Mahoney, Michael W and Wilson, Andrew Gordon and others},
  journal={arXiv preprint arXiv:2603.15802},
  year={2026},
  url = {https://arxiv.org/abs/2603.15802},
}

@article{liu2026falconx,
  title = {{Falcon-X}: A Time Series Foundation Model for Heterogeneous
           Multivariate Modeling},
  author = {Liu, Yiding and Hu, Yifan and Xia, Hongjie and Liu, Peiyuan and Chen
            , Hongzhou and Dai, Xilin and Dong, Zewei and Yang, Jiang-Ming},
  journal = {arXiv preprint arXiv:2605.27286},
  year = {2026},
  url = {https://arxiv.org/abs/2605.27286},
}

@article{podest2026tirex2,
  title = {TiRex-2: Generalizing TiRex to Multivariate Data and Streaming},
  author = {Podest, Patrick and Pichler, Marco and B{\"u}rger, Elias and Z{\'o}
            lyomi, Levente and Voggenberger, Bernhard and Berghammer, Wilhelm and
            Klotz, Daniel and B{\"o}ck, Sebastian and Klambauer, G{\"u}nter and
            Hochreiter, Sepp},
  journal = {arXiv preprint arXiv:2607.01204},
  year = {2026},
  url = {https://arxiv.org/abs/2607.01204},
}

@article{cohen2024toto,
  title = {Toto: Time Series Optimized Transformer for Observability},
  author = {Cohen, Ben and Khwaja, Emaad and Wang, Kan and Masson, Charles and
            Ram{\'e}, Elise and Doubli, Youssef and Abou-Amal, Othmane},
  journal = {arXiv preprint arXiv:2407.07874},
  year = {2024},
  url = {https://arxiv.org/abs/2407.07874},
}

@article{khwaja2026toto2,
  title = {Toto 2.0: Time Series Forecasting Enters the Scaling Era},
  author = {Khwaja, Emaad and Lettieri, Chris and Woo, Gerald and Belouadah,
            Eden and Cenac, Marc and Jarry, Guillaume and Paquin, Enguerrand and
            Zhao, Xunyi and Zhukov, Viktoriya and Abou-Amal, Othmane and Liu,
            Chenghao and Talwalkar, Ameet and Asker, David},
  journal = {arXiv preprint arXiv:2605.20119},
  year = {2026},
  url = {https://arxiv.org/abs/2605.20119},
}

@inproceedings{woo2024moirai,
  title = {Unified Training of Universal Time Series Forecasting Transformers},
  author = {Woo, Gerald and Liu, Chenghao and Kumar, Akshat and Xiong, Caiming
            and Savarese, Silvio and Sahoo, Doyen},
  booktitle = {International Conference on Machine Learning (ICML)},
  year = {2024},
  url = {https://arxiv.org/abs/2402.02592},
  note = {Introduces the LOTSA corpus},
}

@inproceedings{das2024timesfm,
  title = {A Decoder-Only Foundation Model for Time-Series Forecasting},
  author = {Das, Abhimanyu and Kong, Weihao and Sen, Rajat and Zhou, Yichen},
  booktitle = {International Conference on Machine Learning (ICML)},
  year = {2024},
  url = {https://arxiv.org/abs/2310.10688},
}

@misc{timesfm3,
  title = {TimesFM 3.0},
  author = {{Google Research}},
  howpublished = {Hugging Face},
  year = {2026},
  url = {https://huggingface.co/google/timesfm-3.0-pytorch},
  note = {Model release; no standalone paper. The release directs citations to
          arXiv:2310.10688, which describes the original TimesFM architecture and
          not this one},
}

@article{hollmann2025tabpfnv2,
  title = {Accurate Predictions on Small Data with a Tabular Foundation Model},
  author = {Hollmann, Noah and M{\"u}ller, Samuel and Purucker, Lennart and
            Krishnakumar, Arjun and K{\"o}rfer, Max and Hoo, Shi Bin and
            Schirrmeister, Robin Tibor and Hutter, Frank},
  journal = {Nature},
  volume = {637},
  pages = {319--326},
  year = {2025},
  doi = {10.1038/s41586-024-08328-6},
}

@article{hoo2025tabpfnts,
  title = {From Tables to Time: Extending {TabPFN}-v2 to Time Series
           Forecasting},
  author = {Hoo, Shi Bin and M{\"u}ller, Samuel and Salinas, David and Hutter,
            Frank},
  journal = {Transactions on Machine Learning Research},
  year = {2026},
  url = {https://arxiv.org/abs/2501.02945},
  note = {Circulated in 2025 as ``The Tabular Foundation Model TabPFN
          Outperforms Specialized Time Series Forecasting Models Based on Simple
          Features''; the TMLR version is substantially rewritten},
}

@article{sun2025xihe,
  title = {Xihe: Scalable Zero-Shot Time Series Learner via Hierarchical
           Interleaved Block Attention},
  author = {Sun, Yinbo and Fang, Yuchen and Zhu, Zhibo and Li, Jia and Liu, Yu
            and Deng, Qiwen and Zhou, Jun and Yu, Hang and Lu, Xingyu and Ma,
            Lintao},
  journal = {arXiv preprint arXiv:2510.21795},
  year = {2025},
  url = {https://arxiv.org/abs/2510.21795},
  note = {Ant Group; sizes from 9.5M (tiny) to 1.5B (max)},
}

@article{liu2026timers1,
  title = {Timer-S1: A Billion-Scale Time Series Foundation Model with Serial
           Scaling},
  author = {Liu, Yong and Su, Xingjian and Wang, Shiyu and Zhang, Haoran and
            Liu, Haixuan and Wang, Yuxuan and Ye, Zhou and Xiang, Yang and Wang,
            Jianmin and Long, Mingsheng},
  journal = {arXiv preprint arXiv:2603.04791},
  year = {2026},
  url = {https://arxiv.org/abs/2603.04791},
  note = {Tsinghua and ByteDance; 8.3B total parameters, 0.75B activated per
          token},
}

@book{box1970timeseries,
  title = {Time Series Analysis: Forecasting and Control},
  author = {Box, George E. P. and Jenkins, Gwilym M.},
  publisher = {Holden-Day},
  address = {San Francisco},
  year = {1970},
}

@book{hyndman2021fpp,
  title = {Forecasting: Principles and Practice},
  author = {Hyndman, Rob J. and Athanasopoulos, George},
  edition = {3rd},
  publisher = {OTexts},
  address = {Melbourne, Australia},
  year = {2021},
  url = {https://otexts.com/fpp3/},
}

@book{box2015tsa5,
  title = {Time Series Analysis: Forecasting and Control},
  author = {Box, George E. P. and Jenkins, Gwilym M. and Reinsel, Gregory C. and
            Ljung, Greta M.},
  edition = {5th},
  publisher = {Wiley},
  address = {Hoboken, NJ},
  year = {2015},
  note = {Seasonal (SARIMA) models},
}

@book{pankratz1991dynreg,
  title = {Forecasting with Dynamic Regression Models},
  author = {Pankratz, Alan},
  series = {Wiley Series in Probability and Statistics},
  publisher = {Wiley},
  address = {New York},
  year = {1991},
  doi = {10.1002/9781118150528},
}

@book{lutkepohl2005mts,
  title = {New Introduction to Multiple Time Series Analysis},
  author = {L{\"u}tkepohl, Helmut},
  publisher = {Springer},
  address = {Berlin},
  year = {2005},
  doi = {10.1007/978-3-540-27752-1},
}

@book{hyndman2008ets,
  title = {Forecasting with Exponential Smoothing: The State Space Approach},
  author = {Hyndman, Rob J. and Koehler, Anne B. and Ord, J. Keith and Snyder,
            Ralph D.},
  series = {Springer Series in Statistics},
  publisher = {Springer},
  address = {Berlin},
  year = {2008},
  doi = {10.1007/978-3-540-71918-2},
}

@article{friedman2001gbm,
  title = {Greedy Function Approximation: A Gradient Boosting Machine},
  author = {Friedman, Jerome H.},
  journal = {The Annals of Statistics},
  volume = {29},
  number = {5},
  pages = {1189--1232},
  year = {2001},
  doi = {10.1214/aos/1013203451},
}

@inproceedings{chen2016xgboost,
  title = {{XGBoost}: A Scalable Tree Boosting System},
  author = {Chen, Tianqi and Guestrin, Carlos},
  booktitle = {Proceedings of the 22nd ACM SIGKDD International Conference on
               Knowledge Discovery and Data Mining (KDD)},
  pages = {785--794},
  year = {2016},
  doi = {10.1145/2939672.2939785},
}

@inproceedings{ke2017lightgbm,
  title = {{LightGBM}: A Highly Efficient Gradient Boosting Decision Tree},
  author = {Ke, Guolin and Meng, Qi and Finley, Thomas and Wang, Taifeng and
            Chen, Wei and Ma, Weidong and Ye, Qiwei and Liu, Tie-Yan},
  booktitle = {Advances in Neural Information Processing Systems (NIPS)},
  pages = {3146--3154},
  year = {2017},
  url = {https://papers.nips.cc/paper_files/paper/2017/hash/6449f44a102fde848669bdd9eb6b76fa-Abstract.html},
}

@inproceedings{prokhorenkova2018catboost,
  title = {{CatBoost}: Unbiased Boosting with Categorical Features},
  author = {Prokhorenkova, Liudmila and Gusev, Gleb and Vorobev, Aleksandr and
            Dorogush, Anna Veronika and Gulin, Andrey},
  booktitle = {Advances in Neural Information Processing Systems (NeurIPS)},
  year = {2018},
  url = {https://arxiv.org/abs/1706.09516},
}

@inproceedings{nie2023patchtst,
  title = {A Time Series is Worth 64 Words: Long-Term Forecasting with
           Transformers},
  author = {Nie, Yuqi and Nguyen, Nam H. and Sinthong, Phanwadee and Kalagnanam,
            Jayant},
  booktitle = {International Conference on Learning Representations (ICLR)},
  year = {2023},
  url = {https://arxiv.org/abs/2211.14730},
}

@article{salinas2020deepar,
  title = {DeepAR: Probabilistic Forecasting with Autoregressive Recurrent
           Networks},
  author = {Salinas, David and Flunkert, Valentin and Gasthaus, Jan and
            Januschowski, Tim},
  journal = {International Journal of Forecasting},
  volume = {36},
  number = {3},
  pages = {1181--1191},
  year = {2020},
  doi = {10.1016/j.ijforecast.2019.07.001},
}

@inproceedings{oreshkin2020nbeats,
  title = {N-BEATS: Neural Basis Expansion Analysis for Interpretable Time
           Series Forecasting},
  author = {Oreshkin, Boris N. and Carpov, Dmitri and Chapados, Nicolas and
            Bengio, Yoshua},
  booktitle = {International Conference on Learning Representations (ICLR)},
  year = {2020},
  url = {https://arxiv.org/abs/1905.10437},
}

@article{aksu2024gifteval,
  title = {GIFT-Eval: A Benchmark for General Time Series Forecasting Model
           Evaluation},
  author = {Aksu, Taha and Woo, Gerald and Liu, Juncheng and Liu, Xu and Liu,
            Chenghao and Savarese, Silvio and Xiong, Caiming and Sahoo, Doyen},
  journal = {arXiv preprint arXiv:2410.10393},
  year = {2024},
  url = {https://arxiv.org/abs/2410.10393},
  note = {Also presented at the NeurIPS 2024 Workshop on Time Series in the Age
          of Large Models},
}

@article{shchur2025fevbench,
  title = {fev-bench: A Realistic Benchmark for Time Series Forecasting},
  author = {Shchur, Oleksandr and Ansari, Abdul Fatir and Turkmen, Caner and
            Stella, Lorenzo and Erickson, Nick and Guerron, Pablo and
            Bohlke-Schneider, Michael and Wang, Yuyang},
  journal = {arXiv preprint arXiv:2509.26468},
  year = {2025},
  url = {https://arxiv.org/abs/2509.26468},
}

@inproceedings{qiao2026time,
  title = {It's {TIME}: Towards the Next Generation of Time Series Forecasting
           Benchmarks},
  author = {Qiao, Zhongzheng and Pan, Sheng and Wang, Anni and Zhukova,
            Viktoriya and Liu, Yong and Jiang, Xudong and Wen, Qingsong and Long,
            Mingsheng and Jin, Ming and Liu, Chenghao},
  booktitle = {International Conference on Machine Learning (ICML)},
  year = {2026},
  url = {https://arxiv.org/abs/2602.12147},
}

@article{makridakis2022m5,
  title = {M5 Accuracy Competition: Results, Findings, and Conclusions},
  author = {Makridakis, Spyros and Spiliotis, Evangelos and Assimakopoulos,
            Vassilios},
  journal = {International Journal of Forecasting},
  volume = {38},
  number = {4},
  pages = {1346--1364},
  year = {2022},
  doi = {10.1016/j.ijforecast.2021.11.013},
}

@inproceedings{dooley2023forecastpfn,
  title = {ForecastPFN: Synthetically-Trained Zero-Shot Forecasting},
  author = {Dooley, Samuel and Khurana, Gurnoor Singh and Mohapatra, Chirag and
            Naidu, Siddartha V. and White, Colin},
  booktitle = {Advances in Neural Information Processing Systems (NeurIPS)},
  year = {2023},
  url = {https://arxiv.org/abs/2311.01933},
}

@inproceedings{xie2025cauker,
  title = {CauKer: Classification Time Series Foundation Models Can Be
           Pretrained on Synthetic Data},
  author = {Xie, Shifeng and Feofanov, Vasilii and Odonnat, Ambroise and Zan,
            Lei and Alonso, Marius and Zhang, Jianfeng and Palpanas, Themis and
            Pan, Lujia and Zhang, Keli and Redko, Ievgen},
  booktitle = {International Conference on Learning Representations (ICLR)},
  year = {2026},
  url = {https://arxiv.org/abs/2508.02879},
  note = {Oral presentation},
}

@article{moroshan2025tempopfn,
  title = {TempoPFN: Synthetic Pre-training of Linear RNNs for Zero-Shot Time
           Series Forecasting},
  author = {Moroshan, Vladyslav and Siems, Julien and Zela, Arber and
            Carstensen, Timur and Hutter, Frank},
  journal = {arXiv preprint arXiv:2510.25502},
  year = {2025},
  url = {https://arxiv.org/abs/2510.25502},
}

@inproceedings{arango2025chronosx,
  title = {ChronosX: Adapting Pretrained Time Series Models with Exogenous
           Variables},
  author = {Pineda Arango, Sebastian and Mercado, Pedro and Kapoor, Shubham and
            Ansari, Abdul Fatir and Stella, Lorenzo and Shen, Huibin and
            Senetaire, Hugo and Turkmen, Caner and Shchur, Oleksandr and Maddix,
            Danielle C. and Bohlke-Schneider, Michael and Wang, Yuyang and
            Rangapuram, Syama Sundar},
  booktitle = {International Conference on Artificial Intelligence and
               Statistics (AISTATS)},
  year = {2025},
  url = {https://arxiv.org/abs/2503.12107},
}

@article{auer2025cosmic,
  title = {Zero-Shot Time Series Forecasting with Covariates via In-Context
           Learning},
  author = {Auer, Andreas and Parthipan, Raghul and Mercado, Pedro and Ansari,
            Abdul Fatir and Stella, Lorenzo and Wang, Bernie and
            Bohlke-Schneider, Michael and Rangapuram, Syama Sundar},
  journal = {arXiv preprint arXiv:2506.03128},
  year = {2025},
  url = {https://arxiv.org/abs/2506.03128},
}

@article{qin2025cora,
  title = {CoRA: Covariate-Aware Adaptation of Time Series Foundation Models},
  author = {Qin, Guo and Chen, Zhi and Liu, Yong and Shi, Zhiyuan and Liu,
            Haixuan and Huang, Xiangdong and Wang, Jianmin and Long, Mingsheng},
  journal = {arXiv preprint arXiv:2510.12681},
  year = {2025},
  url = {https://arxiv.org/abs/2510.12681},
}

@inproceedings{das2024icf,
  title = {In-Context Fine-Tuning for Time-Series Foundation Models},
  author = {Das, Abhimanyu and Faw, Matthew and Sen, Rajat and Zhou, Yichen},
  booktitle = {International Conference on Machine Learning (ICML)},
  year = {2025},
  url = {https://arxiv.org/abs/2410.24087},
}

@article{dong2024metatst,
  title = {Metadata Matters for Time Series: Informative Forecasting with
           Transformers},
  author = {Dong, Jiaxiang and Wu, Haixu and Wang, Yuxuan and Zhang, Li and
            Wang, Jianmin and Long, Mingsheng},
  journal = {arXiv preprint arXiv:2410.03806},
  year = {2024},
  url = {https://arxiv.org/abs/2410.03806},
  note = {Encodes dataset- and variate-level descriptions as LLM metadata
          tokens alongside series tokens},
}

@article{dutta2025charm,
  title = {Time to Embed: Unlocking Foundation Models for Time Series with
           Channel Descriptions},
  author = {Dutta, Utsav and Pakazad, Sina Khoshfetrat and Ohlsson, Henrik},
  journal = {arXiv preprint arXiv:2505.14543},
  year = {2025},
  url = {https://arxiv.org/abs/2505.14543},
  note = {CHARM; channel-level textual descriptions, invariant to channel
          order},
}

@inproceedings{wang2025chattime,
  title = {ChatTime: A Unified Multimodal Time Series Foundation Model Bridging
           Numerical and Textual Data},
  author = {Wang, Chengsen and Qi, Qi and Wang, Jingyu and Sun, Haifeng and
            Zhuang, Zirui and Wu, Jinming and Zhang, Lei and Liao, Jianxin},
  booktitle = {AAAI Conference on Artificial Intelligence},
  year = {2025},
  url = {https://arxiv.org/abs/2412.11376},
}

@article{xie2025chatts,
  title = {ChatTS: Aligning Time Series with {LLM}s via Synthetic Data for
           Enhanced Understanding and Reasoning},
  author = {Xie, Zhe and Li, Zeyan and He, Xiao and Xu, Longlong and Wen, Xidao
            and Zhang, Tieying and Pei, Dan and others},
  journal = {Proceedings of the VLDB Endowment},
  volume = {18},
  number = {8},
  pages = {2385--2398},
  year = {2025},
  doi = {10.14778/3742728.3742735},
  url = {https://arxiv.org/abs/2412.03104},
}

@article{su2024roformer,
  title = {RoFormer: Enhanced Transformer with Rotary Position Embedding},
  author = {Su, Jianlin and Lu, Yu and Pan, Shengfeng and Murtadha, Ahmed and
            Wen, Bo and Liu, Yunfeng},
  journal = {Neurocomputing},
  volume = {568},
  pages = {127063},
  year = {2024},
  doi = {10.1016/j.neucom.2023.127063},
  url = {https://arxiv.org/abs/2104.09864},
}

@inproceedings{sun2023xpos,
  title = {A Length-Extrapolatable Transformer},
  author = {Sun, Yutao and Dong, Li and Patra, Barun and Ma, Shuming and Huang,
            Shaohan and Benhaim, Alon and Chaudhary, Vishrav and Song, Xia and
            Wei, Furu},
  booktitle = {Proceedings of the 61st Annual Meeting of the Association for
               Computational Linguistics (ACL)},
  pages = {14590--14604},
  year = {2023},
  doi = {10.18653/v1/2023.acl-long.816},
  url = {https://arxiv.org/abs/2212.10554},
}

@inproceedings{henry2020qknorm,
  title = {Query-Key Normalization for Transformers},
  author = {Henry, Alex and Dachapally, Prudhvi Raj and Pawar, Shubham Shantaram
            and Chen, Yuxuan},
  booktitle = {Findings of the Association for Computational Linguistics: EMNLP
               2020},
  pages = {4246--4253},
  year = {2020},
  doi = {10.18653/v1/2020.findings-emnlp.379},
  url = {https://arxiv.org/abs/2010.04245},
}

@inproceedings{dehghani2023vit22b,
  title = {Scaling Vision Transformers to 22 Billion Parameters},
  author = {Dehghani, Mostafa and Djolonga, Josip and Mustafa, Basil and
            Padlewski, Piotr and Heek, Jonathan and Gilmer, Justin and Steiner,
            Andreas and Caron, Mathilde and Geirhos, Robert and Alabdulmohsin,
            Ibrahim and others},
  booktitle = {Proceedings of the 40th International Conference on Machine
               Learning (ICML)},
  year = {2023},
  url = {https://arxiv.org/abs/2302.05442},
}

@article{shazeer2020glu,
  title = {GLU Variants Improve Transformer},
  author = {Shazeer, Noam},
  journal = {arXiv preprint arXiv:2002.05202},
  year = {2020},
  url = {https://arxiv.org/abs/2002.05202},
}

@inproceedings{zhang2019rmsnorm,
  title = {Root Mean Square Layer Normalization},
  author = {Zhang, Biao and Sennrich, Rico},
  booktitle = {Advances in Neural Information Processing Systems (NeurIPS)},
  year = {2019},
  url = {https://arxiv.org/abs/1910.07467},
}

@article{koenker1978quantiles,
  title = {Regression Quantiles},
  author = {Koenker, Roger and Bassett, Gilbert},
  journal = {Econometrica},
  volume = {46},
  number = {1},
  pages = {33--50},
  year = {1978},
  doi = {10.2307/1913643},
}

@article{wen2017mqrnn,
  title = {A Multi-Horizon Quantile Recurrent Forecaster},
  author = {Wen, Ruofeng and Torkkola, Kari and Narayanaswamy, Balakrishnan and
            Madeka, Dhruv},
  journal = {arXiv preprint arXiv:1711.11053},
  year = {2017},
  url = {https://arxiv.org/abs/1711.11053},
  note = {NIPS 2017 Time Series Workshop},
}

@article{potosnak2025forking,
  title = {Forking-Sequences: Statistically and Computationally Efficient
           Multi-Horizon Forecasting with Reduced Volatility},
  author = {Potosnak, Willa and Wolff, Malcolm and Cao, Mengfei and Ma, Ruijun
            and Konstantinova, Tatiana and Efimov, Dmitry and Mahoney, Michael W.
            and Oreshkin, Boris and Olivares, Kin G.},
  journal = {Transactions on Machine Learning Research},
  year = {2026},
  url = {https://arxiv.org/abs/2510.04487},
}

@article{eisenach2020mqtransformer,
  title = {MQTransformer: Multi-Horizon Forecasts with Context Dependent and
           Feedback-Aware Attention},
  author = {Eisenach, Carson and Patel, Yagna and Madeka, Dhruv},
  journal = {arXiv preprint arXiv:2009.14799},
  year = {2020},
  url = {https://arxiv.org/abs/2009.14799},
}

@article{chernozhukov2010rearrangement,
  title = {Quantile and Probability Curves Without Crossing},
  author = {Chernozhukov, Victor and Fern{\'a}ndez-Val, Iv{\'a}n and Galichon,
            Alfred},
  journal = {Econometrica},
  volume = {78},
  number = {3},
  pages = {1093--1125},
  year = {2010},
  doi = {10.3982/ECTA7880},
}

@article{raffel2020t5,
  title = {Exploring the Limits of Transfer Learning with a Unified Text-to-Text
           Transformer},
  author = {Raffel, Colin and Shazeer, Noam and Roberts, Adam and Lee, Katherine
            and Narang, Sharan and Matena, Michael and Zhou, Yanqi and Li, Wei
            and Liu, Peter J.},
  journal = {Journal of Machine Learning Research},
  volume = {21},
  number = {140},
  pages = {1--67},
  year = {2020},
  url = {https://arxiv.org/abs/1910.10683},
}

@article{krell2021packing,
  title = {Efficient Sequence Packing without Cross-contamination: Accelerating
           Large Language Models without Impacting Performance},
  author = {Krell, Mario Michael and Kosec, Matej and Perez, Sergio P. and
            Fitzgibbon, Andrew},
  journal = {arXiv preprint arXiv:2107.02027},
  year = {2021},
  url = {https://arxiv.org/abs/2107.02027},
}

@inproceedings{ding2024fewer,
  title = {Fewer Truncations Improve Language Modeling},
  author = {Ding, Hantian and Wang, Zijian and Paolini, Giovanni and Kumar,
            Varun and Deoras, Anoop and Roth, Dan and Soatto, Stefano},
  booktitle = {International Conference on Machine Learning (ICML)},
  year = {2024},
  url = {https://arxiv.org/abs/2404.10830},
}

@article{lodi2002twodim,
  title = {Two-Dimensional Packing Problems: A Survey},
  author = {Lodi, Andrea and Martello, Silvano and Monaci, Michele},
  journal = {European Journal of Operational Research},
  volume = {141},
  number = {2},
  pages = {241--252},
  year = {2002},
  doi = {10.1016/S0377-2217(02)00123-6},
}

@inproceedings{park2022iqf,
  title = {Learning Quantile Functions without Quantile Crossing for
           Distribution-free Time Series Forecasting},
  author = {Park, Youngsuk and Maddix, Danielle and Aubet, Fran{\c{c}}ois-Xavier
            and Kan, Kelvin and Gasthaus, Jan and Wang, Yuyang},
  booktitle = {International Conference on Artificial Intelligence and
               Statistics (AISTATS)},
  series = {Proceedings of Machine Learning Research},
  volume = {151},
  pages = {8127--8150},
  year = {2022},
  url = {https://arxiv.org/abs/2111.06581},
}

@article{laio2007verification,
  title = {Verification tools for probabilistic forecasts of continuous
           hydrological variables},
  author = {Laio, Francesco and Tamea, Stefania},
  journal = {Hydrology and Earth System Sciences},
  volume = {11},
  number = {4},
  pages = {1267--1277},
  year = {2007},
  doi = {10.5194/hess-11-1267-2007},
}

@article{gneiting2011comparing,
  title = {Comparing density forecasts using threshold- and quantile-weighted
           scoring rules},
  author = {Gneiting, Tilmann and Ranjan, Roopesh},
  journal = {Journal of Business \& Economic Statistics},
  volume = {29},
  number = {3},
  pages = {411--422},
  year = {2011},
  doi = {10.1198/jbes.2010.08110},
}

@inproceedings{bengio2009curriculum,
  title = {Curriculum learning},
  author = {Bengio, Yoshua and Louradour, J{\'e}r{\^o}me and Collobert, Ronan
            and Weston, Jason},
  booktitle = {Proceedings of the 26th Annual International Conference on
               Machine Learning},
  pages = {41--48},
  year = {2009},
  doi = {10.1145/1553374.1553380},
}

@inproceedings{loshchilov2019adamw,
  title = {Decoupled weight decay regularization},
  author = {Loshchilov, Ilya and Hutter, Frank},
  booktitle = {International Conference on Learning Representations (ICLR)},
  year = {2019},
  url = {https://arxiv.org/abs/1711.05101},
}

@misc{skforecast,
  title = {skforecast},
  author = {Amat Rodrigo, Joaquin and Escobar Ortiz, Javier},
  year = {2026},
  howpublished = {Version 0.24.0, Zenodo},
  doi = {10.5281/zenodo.8382788},
  note = {Python library for time series forecasting with a scikit-learn
          interface, BSD-3-Clause, \url{https://skforecast.org/}},
}

@misc{skforecast2026foundation,
  title = {Forecasting with foundation models},
  author = {Amat Rodrigo, Joaquin and Escobar Ortiz, Javier},
  year = {2026},
  howpublished = {skforecast user guide},
  url = {https://skforecast.org/latest/user_guides/foundation-forecasting-models},
  note = {Also published at
          \url{https://cienciadedatos.net/documentos/py79-forecasting-with-foundation-models.html};
          accessed September 7, 2026},
}

@misc{oharawild2022tsibbledata,
  title = {tsibbledata: Diverse Datasets for `tsibble'},
  author = {O'Hara-Wild, Mitchell and Hyndman, Rob and Wang, Earo and Godahewa,
            Rakshitha},
  year = {2022},
  howpublished = {R package},
  url = {https://tsibbledata.tidyverts.org/},
  note = {Dataset \texttt{vic\_elec}: half-hourly electricity demand for
          Victoria, Australia},
}

@misc{godahewa2021australian,
  title = {Australian Electricity Demand Dataset},
  author = {Godahewa, Rakshitha and Bergmeir, Christoph and Webb, Geoffrey and
            Hyndman, Rob and Montero-Manso, Pablo},
  year = {2021},
  howpublished = {Monash Time Series Forecasting Archive, Zenodo},
  doi = {10.5281/zenodo.4659727},
  note = {Half-hourly demand of five Australian states, extracted from the
          tsibbledata R package},
}

@misc{aemo2024priceanddemand,
  title = {Aggregated price and demand data, region {VIC1}, 2023--2024},
  author = {{AEMO}},
  year = {2024},
  howpublished = {Australian Energy Market Operator, monthly CSV files
                  \texttt{PRICE\_AND\_DEMAND\_<YYYYMM>\_VIC1.csv} at
                  \url{https://aemo.com.au/aemo/data/nem/priceanddemand/}},
  url = {https://aemo.com.au/energy-systems/electricity/national-electricity-market-nem/data-nem/aggregated-data},
  note = {Five-minute \texttt{TOTALDEMAND} in MW on National Electricity Market
          time. AEMO permits use of its material for any purpose with attribution
          (\url{https://aemo.com.au/privacy-and-legal-notices/copyright-permissions});
          accessed September 8, 2026},
}

@misc{noaa2001isd,
  title = {Global Surface Hourly [{Integrated Surface Database}, station
           959360-99999, {Melbourne (Olympic Park)}, 2023--2024]},
  author = {{NOAA NCEI}},
  year = {2001},
  howpublished = {NOAA National Centers for Environmental Information},
  url = {https://www.ncei.noaa.gov/data/global-hourly/},
  note = {Non-US stations fall under WMO Resolution 40: free for research,
          education and other non-commercial use; the data and derived products
          may not be redistributed. Accessed September 8, 2026},
}

@misc{zippenfenig2023openmeteo,
  title = {{Open-Meteo.com Weather API}},
  author = {Zippenfenig, Patrick},
  year = {2023},
  howpublished = {Zenodo},
  doi = {10.5281/zenodo.7970649},
  note = {Historical weather API serving the ERA5 reanalysis,
          \url{https://open-meteo.com/en/docs/historical-weather-api}. Weather
          data by Open-Meteo.com, CC BY 4.0; accessed September 8, 2026},
}

@misc{vacanza2026holidays,
  title = {holidays: {Open World Holidays Framework}},
  author = {{Vacanza Team}},
  year = {2026},
  howpublished = {Python package},
  url = {https://github.com/vacanza/holidays},
  note = {MIT license},
}

@article{gneiting2007probabilistic,
  title = {Probabilistic forecasts, calibration and sharpness},
  author = {Gneiting, Tilmann and Balabdaoui, Fadoua and Raftery, Adrian E.},
  journal = {Journal of the Royal Statistical Society: Series B (Statistical
             Methodology)},
  volume = {69},
  number = {2},
  pages = {243--268},
  year = {2007},
  doi = {10.1111/j.1467-9868.2007.00587.x},
}

@book{rochberg2004heavenly,
  title = {The Heavenly Writing: Divination, Horoscopy, and Astronomy in
           Mesopotamian Culture},
  author = {Rochberg, Francesca},
  publisher = {Cambridge University Press},
  address = {Cambridge},
  year = {2004},
  doi = {10.1017/CBO9780511617409},
}

@article{steele2000eclipse,
  title = {Eclipse Prediction in Mesopotamia},
  author = {Steele, John M.},
  journal = {Archive for History of Exact Sciences},
  volume = {54},
  number = {5},
  pages = {421--454},
  year = {2000},
  doi = {10.1007/PL00021245},
}

\appendix
\section*{Acknowledgments}

The Forecasting Company and the authors thank Macrocosm for agreeing to
evaluate the \tzero{} models independently, and for the time and effort that
took. Macrocosm designed the ERCOT forecasting and trading study of
Section~\ref{sec:usecase-ercot}, ran it on their own data and infrastructure,
and wrote the section.

\section{GIFT-Eval cross-domain subset}
\label{app:gifteval-subset}

Table~\ref{tab:gifteval-subset} lists the twelve dataset and term pairs that
the studies of Sections~\ref{sec:tails}, \ref{sec:missing-data}
and~\ref{sec:context-efficiency} run on. The twelve pairs come from twelve
distinct datasets, one pair each. They cover all seven GIFT-Eval domains,
four frequencies from 15-minute to monthly, and the three horizon terms,
at a fifth of the inference cost per configuration of the full 97-pair
benchmark. Two of
the twelve carry irregularity by design: \texttt{kdd\_cup\_2018} has native
gaps and \texttt{car\_parts} is intermittent, so the missing-data study of
Section~\ref{sec:missing-data} measures injected masks on top of
missingness a user would also meet.

\begin{table}[htb]
  \centering
  \caption{The twelve GIFT-Eval dataset and term pairs used by the studies
    of Sections~\ref{sec:tails}, \ref{sec:missing-data}
    and~\ref{sec:context-efficiency}. Dataset names, frequency codes and
    terms are those GIFT-Eval declares; domains are its seven leaderboard
    classes.}
  \label{tab:gifteval-subset}
  \begin{tabular}{llll}
    \toprule
    \tfcheadrow{Dataset} & \tfcheadrow{Frequency} & \tfcheadrow{Term} & \tfcheadrow{Domain} \\
    \midrule
    \texttt{electricity} & D & short & Energy \\
    \texttt{solar} & H & long & Energy \\
    \texttt{bizitobs\_l2c} & H & short & Web/CloudOps \\
    \texttt{bitbrains\_rnd} & H & short & Web/CloudOps \\
    \texttt{loop\_seattle} & D & short & Transport \\
    \texttt{sz\_taxi} & 15T & long & Transport \\
    \texttt{jena\_weather} & H & short & Nature \\
    \texttt{kdd\_cup\_2018} & H & medium & Nature \\
    \texttt{restaurant} & D & short & Sales \\
    \texttt{car\_parts} & M & short & Sales \\
    \texttt{m4\_hourly} & H & short & Econ/Fin \\
    \texttt{covid\_deaths} & D & short & Healthcare \\
    \bottomrule
  \end{tabular}
\end{table}

\end{document}